\documentclass[sigconf,balance=false]{acmart}
\usepackage{popets}
\usepackage[nolist]{acronym}
\usepackage{wrapfig}
\usepackage{subfig}
\usepackage{amsmath}
\usepackage{multirow}
\usepackage{amsfonts}
\usepackage[linesnumbered,ruled,vlined]{algorithm2e}

\setcopyright{popets}
\copyrightyear{YYYY}

\acmYear{YYYY}
\acmVolume{YYYY}
\acmNumber{X}
\acmDOI{XXXXXXX.XXXXXXX}
\acmISBN{}
\acmConference{Proceedings on Privacy Enhancing Technologies}
\begin{document}

\title[Vertical Federated Unlearning]{Unlearning Clients, Features and Samples in \\ Vertical Federated Learning}


\author{Ayush K. Varshney*}
\orcid{0000-0002-8073-6784}
\affiliation{%
  \institution{Umeå University}
  \city{Umeå}
  \country{Sweden}}
\email{ayushkv@cs.umu.se}
\thanks{*Corresponding author}

\author{Konstantinos Vandikas}
\orcid{0000-0001-6925-0954}
\affiliation{%
  \institution{Ericsson Research, Ericsson}
  \city{Stockholm}
  \country{Sweden}}
\email{konstantinos.vandikas@ericsson.com}

\author{Vicen\c{c} Torra}
\orcid{0000-0002-0368-8037}
\affiliation{%
  \institution{Umeå University}
  \city{Umeå}
  \country{Sweden}}
\email{vtorra@cs.umu.se}


\renewcommand{\shortauthors}{Varshney et al.}

\begin{acronym}
    \acro  {FL}   [FL]   {Federated Learning}
    \acro  {VFL}  [VFL]  {Vertical Federated Learning}
    \acro  {HFL}  [HFL]  {Horizontal Federated Learning}
    \acro  {FTL}  [FTL]  {Federated Transfer Learning}
    \acro  {MIA}  [MIA]  {Membership Inference Attack}
    \acro  {ML}   [ML]   {Machine Learning}
    \acro  {KD}   [KD]   {Knowledge Distillation}
    \acro  {GA}   [GA]   {Gradient Ascent}
\end{acronym}

\begin{abstract}
\ac{FL} has emerged as a prominent distributed learning paradigm that allows multiple users to collaboratively train a model without sharing their data thus preserving privacy. Within the scope of privacy preservation, information privacy regulations such as GDPR entitle users to request the removal (or unlearning) of their contribution from a service that is hosting the model. For this purpose, a server hosting an ML model must be able to unlearn certain information in cases such as copyright infringement or security issues that can make the model vulnerable or impact the performance of a service based on that model. While most unlearning approaches in \ac{FL} focus on \ac{HFL}, where clients share the feature space and the global model, \ac{VFL} has received less attention from the research community. \ac{VFL} involves clients (passive parties) sharing the sample space among them while not having access to the labels. In this paper, we explore unlearning in VFL from three perspectives: unlearning passive parties, unlearning features, and unlearning samples. To unlearn passive parties and features we introduce VFU-KD which is based on knowledge distillation (KD) while to unlearn samples, VFU-GA is introduced which is based on gradient ascent (GA). To provide evidence of approximate unlearning, we utilize Membership Inference Attack (MIA) to audit the effectiveness of our unlearning approach. Our experiments across six tabular datasets and two image datasets demonstrate that VFU-KD and VFU-GA achieve performance comparable to or better than both retraining from scratch and the benchmark R2S method in many cases, with improvements of $(0-2\%)$. In the remaining cases, utility scores remain comparable, with a modest utility loss ranging from $1-5\%$. Unlike existing methods, VFU-KD and VFU-GA require no communication between active and passive parties during unlearning. However, they do require the active party to store the previously communicated embeddings.
\end{abstract}

\maketitle

\section{Introduction}
\ac{ML} models have established themselves as the prominent artificial intelligence (AI) approach due to their ability to learn complex patterns from large-scale user data. The huge amount of data used to train these models is often sensitive in nature. To safe guard the information and privacy of the users, information privacy regulations such as GDPR, and CCPA have been proposed. These regulations allow users the right to be forgotten, i.e., to remove their data and its influence from the \ac{ML} model.

Removing the influence of the data or \textit{Unlearning} the data is a challenging task. A naive approach is to retrain the model in the absence of the data to remove. This approach can be time-consuming and assumes that the original data is available. A more attractive approach should remove the data and its influence without the need of retraining from scratch. The objective is to modify the model parameters of a \ac{ML} model in such a way that the modified parameters are the same as those parameters of a model that would have been retrained from scratch with an original dataset deprived of the data to be forgotten or unlearned. However, achieving such an objective can be computationally expensive. An approach proposed by Bourtoule et al. \cite{bourtoule2021machine} divides the dataset into shards, retraining only the shard containing the data to be removed upon receiving an unlearning request. While this method minimizes the scope of retraining, it may struggle to capture complex relationships across shards and becomes computationally expensive for frequent unlearning requests. To address such challenges, Ginart et al. \cite{ginart2019making} introduced the concept of approximate unlearning, where model parameters are adjusted to closely approximate those of a retrained model, reducing the need for full retraining. This concept has since been extended in several studies. For instance, Halimi et al. \cite{halimi2022federated} propose updating the model using gradient ascent on the data to be forgotten. Wu et al. \cite{wu2022federated} leverage knowledge distillation for unlearning specific samples, while Tarun et al. \cite{tarun2023fast} refine model parameters through modified fine-tuning for efficient unlearning.


The majority of existing approaches in machine unlearning operate under the assumption that the original dataset, or the specific data points to be removed, are readily available, refer \cite{zhang2023review} for more information. These methods typically rely on access to the dataset for performing the unlearning process, which involves directly manipulating or retraining the model on the modified data. This assumption simplifies the unlearning task but may not be practical or feasible in many real-world scenarios where data access is restricted due to privacy concerns, regulatory requirements, or logistical constraints. 

In a distributed paradigm like Federated Learning (FL)~\cite{mcmahan2017communication}, clients collaboratively train a ML model without sharing their data. They only communicate the model weights to a server, which in turn aggregate the weights from clients to create a global model. This continues for several rounds to learn complex relationships between client data. Based on the nature of the data distributed among clients \ac{FL} is further classified into three types, horizontal, vertical and transfer \ac{FL} \cite{prayitno2021systematic}. In \ac{HFL}, clients share the same feature space, but the samples and their distributions differ across each client. This type is suitable for a large number of heterogeneous devices in a complex network. However, its application in collaboration between companies or institutions is practically restricted due to its limited ability to handle different feature space. In \ac{VFL}, clients share the same sample space but have different feature space, with only the server having access to the labels. This setup enables institutions with distinct feature spaces to collaborate on training \ac{ML} models while ensuring the privacy of their raw data. Notably, data alignment is required in \ac{VFL} before training. To achieve this, clients utilize private set intersection (PSI) ~\cite{dong2013private} to identify the common sample space while preserving data privacy. For further details on PSI and its drawbacks, refer to Wu et al.\cite{wu2024federated}. With \ac{FTL}, the server leverages knowledge from one domain to another, facilitating collaborative training of \ac{ML} models across different domains. \ac{FTL} enables knowledge transfer and model training even when clients have different feature spaces and sample distributions. 

Unlearning in \ac{FL} has its own unique challenges. In each communication round in \ac{FL}, the client contribution is aggregated and communicated to the rest of the clients. The server does not have access to the client data. The role of the unlearner depends on the unlearning query. For example, Wang et al.~\cite{wang2023federated} highlights that when a client request to unlearn the server acts as the unlearner, and is responsible for adjusting the parameter of the global model; when a client request to unlearn a portion of their data, the client themselves are the unlearner and update their local model parameters. Jiang et al. \cite{jiang2024towards} remove the contributions from the client iteratively based on its historical data, Zhu et al. \cite{zhu2023heterogeneous} unlearn the model by averaging the model updates from the remaining clients while optimizing distillation loss between the original model and unlearned model, Li et al. \cite{li2023subspace} propose a gradient ascent based approach to unlearn a client weight, Wu et al.~\cite{wu2022federated} utilize \ac{KD} for federated unlearning and many more such methods. Majority of the unlearning literature in \ac{FL} has focused on \ac{HFL}. In \ac{HFL}, the model architecture remains the same, therefore the server can store historical updates from the clients and produce a calibrated unlearned model for approximate unlearning \cite{liu2021federaser}. However, the model architecture in \ac{VFL} may change depending on the type of unlearning required. As a result, existing unlearning approaches designed for \ac{HFL} cannot be directly applied to \ac{VFL} without significant modifications.

In VFL, each training round involves communicating embeddings from clients (passive parties) to the server (active party), making the training process communication intensive. Consequently, any unlearning approach that necessitates even a few additional training rounds between the active and passive parties is undesirable due to the communication overhead it imposes. Also, the model architecture of the active party depends on the embeddings from the passive parties i.e., removing the embeddings of a passive party reduces the size of the input layer in the active model. Here as well, the unlearner can be different based on the unlearning request. 

The literature for unlearning in \ac{VFL} is scarce, Deng et al.~\cite{deng2023vertical} propose one of the first unlearning approach in \ac{VFL}, however it is restricted  to logistic regression. Their approach stores the last communication round embeddings from each client, and assumes unlearning request can come only after the completion of training process, which is not realistic. The learning in \ac{FL} is continuous in nature and unlearning request can come during training as well. Another approach~\cite{wang2024efficient} proposes a fast retraining method for \ac{VFL} by storing bottom model checkpoints. These bottom model checkpoints are used to reinitialize the passive parties. Their approach proposes fast retraining method which maintains several passive party models which are used to reinitialize and retrain the passive models from scratch. Their retraining approach for unlearning in \ac{VFL} requires communication between active and passive parties, which is costly. Additionally, for feature unlearning and passive party unlearning—i.e., when a passive party wishes to remove certain features from its local model, and when a passive party needs to be unlearned from the global model—this often requires reducing model parameters, which in turn requires retraining from scratch in the existing literature. Notably, none of the existing approaches in the VFL unlearning literature address the challenges of feature, sample, and passive party unlearning simultaneously.

Overall, we find that for passive party unlearning and feature unlearning, approaches in VFL should be able to deal with the reduction in model size. It should not involve any communication between the active and passive parties. Considering these challenges in mind, we propose a knowledge distillation based unlearning approach for \ac{VFL} which we call Vertical Federated Unlearning with Knowledge Distillation (VFU-KD). The advantage of unlearning with \ac{KD}, is its ability to handle model compression well and it does not require communication between active and passive parties. We also extend VFU-KD to propose the first feature unlearning approach for a passive party in a VFL. For sample unlearning in VFL, unlearning with \ac{KD} is an expensive approach as sample unlearning does not require model compression. On the other hand, gradient ascent \cite{graves2021amnesiac} is a popular approach to unlearn samples in centralized machine learning. Considering this in mind, we propose a gradient ascent based unlearning for \ac{VFL} called Vertical Federated Unlearning with Gradient Ascent (VFU-GA) which maximize the loss on the forget set and fine tune the model for few epochs on the remaining data for approximate unlearning.

Auditing the unlearning algorithm is also crucial to verify that the unlearning has been done. In the literature, the majority of the approaches use the drop in accuracy of a backdoor attack \cite{bagdasaryan2020backdoor} as the sign of unlearning. However, in case of unlearning in VFL, backdoor attacks can not be used to audit unlearning as once the passive party embeddings are removed, it is not possible to place the backdoor in the active model. The data poisoning attack proposed in Deng et al.~\cite{deng2023vertical} considers the availability of labels to the passive party\footnote{\href{https://github.com/dateaaalive/vfl/blob/main/data_poison_attack.py}{https://github.com/dateaaalive/vfl/blob/main/data\_poison\_attack.py}} which violates the fundamental assumptions in the VFL setting. In our work, we propose a membership inference attack (MIA) \cite{nasr2019comprehensive}, which does not violate the VFL constraints and can be used to audit unlearning in VFL.

In summary, we make the following contributions.

\begin{enumerate}
    \item A novel vertical federated unlearning framework with knowledge distillation to unlearn a passive party.
    \item A novel feature unlearning framework for a passive party in VFL.
    \item A novel sample unlearning framework using gradient ascent in VFL.
    \item A membership inference attack to audit the unlearning in VFL.
    \item Empirical analysis on tabular as well as image datasets.
\end{enumerate}

The rest of the paper is organized as follows. Section \ref{background} provides the necessary background. Section \ref{Proposed} describes our proposed frameworks. Section \ref{Experiments} gives the experimental analysis. The paper finishes with conclusion and some future work in Section \ref{Conclusion}.

\section{Background} \label{background}

\subsection{Vertical federated learning} \label{VFL}

\begin{figure}
    \includegraphics[width=\columnwidth]{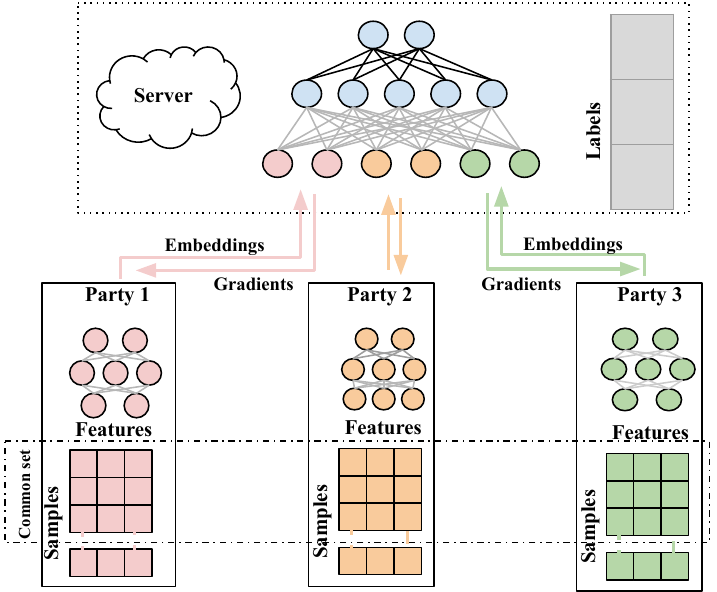}
    \caption{Vertical federated learning framework} \label{fig:VFL framework}
\end{figure}

The demand for \ac{VFL} has grown rapidly in recent years \cite{liu2021fate}. \ac{VFL} is a distributed learning paradigm that allows organizations with distinct feature spaces to collaboratively train \ac{ML} model while safeguarding the privacy of their raw data. This approach is particularly valuable in scenarios where data privacy is critical, such as in healthcare \cite{yu2024communication}, finance \cite{zheng2020vertical}, and cross-enterprise collaborations \cite{liu2024vertical}. In these domains, institutions often possess complementary datasets but are constrained from sharing them due to regulatory requirements or competitive concerns. In a typical \ac{VFL} setup, there are $N$ passive parties (i.e., clients) that own data but lack access to the labels, and a single active party, typically the server or a trusted third party, that holds the labels as shown in Fig. \ref{fig:VFL framework}. This ensures that the learning process can leverage distributed data sources while maintaining strict privacy guarantees. Organizations and institutions with limited and fragmented datasets constantly seek data partners to collaboratively train ML models to maximize data utilization \cite{li2021survey}. \ac{VFL} requires all participants to share sample space but allows for different feature spaces. This essentially suggests less number of participants in \ac{VFL} (Wei et al. \cite{wei2022vertical} suggests $N < 5$), with two-party \ac{VFL} as the most common setting. In \ac{VFL}, the first step is to perform private set intersection \cite{lu2020multi} to identify common sample IDs among participating parties while preserving privacy. Each passive party can have a local model, referred to as a passive model. Based on the common set from private set intersection, passive parties do a forward pass on their local models. The computed embeddings are communicated to the active party, which concatenates the embeddings from the passive parties and then performs a forward pass on its model, referred to as the active model. The active party computes the loss and backpropagates the gradients to the active model and to the respective embeddings. The active party then communicates the respective gradients to each passive party. Upon receiving the gradients, passive parties update their local models. It is important to highlight that this process continues for each batch and requires several communication rounds between the active and passive parties for one epoch of training data. Hence, training in \ac{VFL} is a communication-intensive algorithm \cite{wei2022vertical}.

\subsection{Knowledge distillation} \label{KD}

\ac{KD} approaches has been used in the literature for their ability to transfer knowledge from a bigger teacher model to a smaller student model \cite{hinton2015distilling}. The simple idea of KD is that student model tries to mimic the output probabilities of the teacher model. Usually, student model learns from its own loss and its divergence from the teacher's logits. For a labeled dataset with $\hat{y}$ as true labels, the loss function for the student model can be given as follows:
\begin{equation}
    \mathcal{L} = (1-\alpha)L_{pred}(\hat{y},\hat{y}_{student}) + KL\_DIV(\hat{y}_{student}, \hat{y}_{teacher})
\end{equation}
here $\alpha$ is used to manage the trade-off between distillation loss and prediction loss; $\hat{y}_{student},$ $\hat{y}_{teacher}$ are the logits of the student and teacher models respectively; and $KL\_DIV()$ is the KL divergence between them. Please refer \cite{gou2021knowledge} for a discussion of recent advancements in the area of \ac{KD}.

\subsection{Gradient ascent} \label{GA}

Gradient ascent is the counterpart of gradient descent which is the typical optimization approach used to train a machine learning model. In gradient descent, the objective is to minimize the loss function for a given set of samples, whereas in gradient ascent (GA), the goal is to maximize the loss function for those samples. This approach is particularly useful for unlearning, as it allows for the approximate removal of specific samples by adjusting the model weights in the direction that maximizes the loss on the target samples \cite{tarun2023fast}, \cite{thudi2022unrolling}. Let $\theta$ represent the model weight, $\eta$ be the learning rate and $L$ be the loss function, then \ac{GA} iteratively updates the model weights in the following manner.
\begin{equation}
    \theta_{t+1} = \theta_t + \eta\frac{\delta L}{\delta \theta_t}
\end{equation}

\subsection{Membership inference attack} \label{MIA}

\ac{MIA} enables adversaries to determine whether a particular record was part of the training set in a \ac{ML} model. \ac{MIA} leverages the memorization of ML models i.e, ML models behave differently on the data seen during training. Consequently by analyzing the model's response on various inputs, an attacker can infer the presence or absence of specific data during training. 

\section{Proposed Work} \label{Proposed}

In this section, we present the details of the proposed unlearning framework for passive party unlearning, sample unlearning and feature unlearning in \ac{VFL}. We also provide the details of \ac{MIA} used to audit unlearning in our framework. Unlearning is a crucial capability for addressing a variety of issues, including privacy compliance, security, and adaptability. Existing approaches either violate VFL constraints or are expensive in terms of communication for passive party unlearning. None of the approaches (to the best of our knowledge) focuses on feature unlearning in VFL.

Algorithm \ref{Algo VFL} shows the generic VFL framework with $K$ passive parties, and $K^{th}$ party being the active party as well. Let $\mathcal{G}()$ be a function which takes model parameters $\theta$, and minibatch $x$ as input and returns the embeddings from the model. 

Passive parties have their local models $\theta_1,\theta_2,..., \theta_K$ and $\Theta_K$ is the active model. For each batch, passive parties do a forward pass and communicate their embeddings to the server. Server in turn trains its local model and forward the gradients with respect to each embedding ($\nabla_{k}^tL = \frac{\delta L}{\delta \theta_k^t} = \frac{\delta L}{\delta \Theta_K^t} \times \frac{\delta \Theta_K^t}{\delta \theta_k^t}$, $L$ being the loss function) back to their respective passive party. Algorithm \ref{Algo VFL} requires parameters $\eta_1, \eta_2$ which needs to be calibrated to have a successful learning. However this can be eliminated if we assume the loss $L$ to be twice differential and strictly convex, then the parameter update can be written as:
\begin{equation}
    \Theta_K^{t+1} = \Theta_K^t - \mathcal{H}_{\Theta_K^t}^{-1} \frac{\delta L}{\delta \Theta_K^t} \label{hessian vfl}
\end{equation}
And for passive parties:
\begin{equation}
    \theta_k^{t+1} = \theta_k^t - \mathcal{H}_{\theta_k^t}^{-1}\nabla_{k}^tL
\end{equation}
where $\mathcal{H}$ is the hessian matrix for each model.~\cite{warnecke2021machine} suggests computing and storing comes at an additional computational cost of $O(np^2+p^3)$ and $O(p^2)$ respectively. We have considered both the approaches i.e., using $\eta_1, \eta_2$ and $\mathcal{H}^{-1}$ for the respective models. 

\begin{algorithm}
\DontPrintSemicolon 
\SetKwInput{KwData}{Input}
\SetKwInput{KwResult}{Output}
\caption{Vertical Federated Learning Algorithm}
\label{Algo VFL}
\KwData{Passive parties $\theta_1, \theta_2, ..., \theta_K$, Active party $\Theta_K$, learning rates $\eta_1, \eta_2$}
\KwResult{Trained model weights $\theta_1, \theta_2, ..., \theta_K$ and $\Theta_K$}
Randomly initialize $\theta_1, \theta_2, ..., \theta_K$ and $\Theta_K$. \\
\For{$t = 1,2,\dots, T$} {
    Randomly sample a minibatch $x \in \mathcal{D}$ \\
    \For{each party $k=1,2,...,K$ \textbf{in parallel}} {
        Party $k$ does a forward pass and computes embeddings $H_k = \mathcal{G}(\theta_k^t, x)$ \\
        Party $k$ sends $H_k$ to the Active party \\
    }
    Active party computes $H^t = concat(H_1, H_2,..., H_k)$ \\
    Active party stores $H^t$ \\
    Compute prediction $\hat{y} = \Theta_K^t(H^t)$ \\
    Compute loss $L = LossFunction(\hat{y}, y_{true})$ \\
    Active party updates $\Theta_K^{t+1} = \Theta_K^t - \eta_1\frac{\delta L}{\delta \Theta_K^t}$ \\
    Active party computes $\frac{\delta L}{H_k}$ and sends it to the respective passive parties \\
    \For{each party $k=1,2,...,K$ \textbf{in parallel}} {
        Party $k$ computes $\nabla_{k}^tL = \frac{\delta L}{\delta \theta_k^t} = \frac{\delta L}{\delta \Theta_K^t} \times \frac{\delta \Theta_K^t}{\delta \theta_k^t}$ \\
        Party $k$ updates $\theta_k^{t+1} = \theta_k^t - \eta_2\nabla_{k}^tL$
    }
}
\end{algorithm}

\begin{algorithm}
\DontPrintSemicolon 
\SetKwInput{KwData}{Input}
\SetKwInput{KwResult}{Output}
\caption{Passive Party Unlearning in VFL}
\label{Algo Client unlearning VFL}
\KwData{Target passive party  $\theta_u$, Active party $\Theta_K$, learning rate $\eta_1$, distillation rate $\alpha$, current epoch $ep$}
\KwResult{Unlearned Active party model $\Theta_K$, Updated historical embeddings}
Randomly initialize $\Theta_{student}$ with input size of $H \setminus H_u$ \\
Assign $\Theta_K$ as teacher model $\Theta_{teacher}$ \\
\For{$t = 1,2,\dots, ep$} {  
    Active party reads $H^t$ \\
    Compute teacher prediction $\hat{y}_{teacher} = \Theta_{teacher}(H^t) $ \\
    Update $H^t = H^t \setminus H_u$ \tcp*{$H_u$ is the target party $\theta_u$ embeddings}
    Active party removes the old $H^t$ and stores the updated $H^t$\\
    Compute prediction $\hat{y}_{student} = \Theta_{student}^t(H^t)$ \\
    Compute prediction loss $L_{pred} = LossFunction(\hat{y}_{student}, y_{true})$ \\
    Compute distillation loss $L_{distil} = KL\_DIV(\hat{y}_{student}, \hat{y}_{teacher})$ \\
    Overall loss $L = \alpha *L_{distil} + (1-\alpha)*L_{pred}$ \\
    Active party updates $\Theta_{student}^{t+1} = \Theta_{student}^t - \eta_1\frac{\delta L}{\delta \Theta_{student}^t}$
}
\end{algorithm}

To unlearn a passive party, the active party must remove the contribution of the target passive party from all of its historical embeddings, i.e., the active party updates $H = H \setminus H_u$ in all the training rounds, $H_u$ being the target passive party. In our work, we use the \ac{KD} approach as the unlearning mechanism as it can deal with model compression, and since the active party already has the embeddings from the previous rounds, the active party does not need to have any communication between active and passive parties.

In our approach, we first randomly initialize a new student model to eliminate any previous information, and assign the old active model as the teacher model. The active party updates its student model based on the historical embeddings from the rest of the clients. The loss for the student model is the combination of prediction loss and the distillation loss. 
\begin{equation} \label{overall_loss}
    L = \alpha *L_{distil} + (1-\alpha)*L_{pred}
\end{equation}
The parameter $\alpha$ balances the trade-off between prediction loss ($L_{pred}$) and distillation loss ($L_{distil}$). In our case, we have considered the $KL\_DIV()$ between as the distillation loss. The output probabilities of the teacher model is used to guide the training of the student model, but our approach is not restricted to it, other distillation functions such as in \cite{gou2021knowledge} can also be used. The training of student model considers stored embedding for each batch and computes overall loss as defined in eq.~\ref{overall_loss}. This process continues till the student model converges or till the given number of epochs. Algorithm \ref{Algo Client unlearning VFL} shows the formal algorithm for unlearning a passive party in VFL setting. Here as well, if the loss is twice differential and strictly convex the model update can be written as:
\begin{equation}
    \Theta_{student}^{t+1} = \Theta_{student}^t - \mathcal{H}_{\Theta_{student}}^{-1}\frac{\delta L}{\delta \Theta_{student}^t} \label{hessian client unlearning}
\end{equation}

\begin{algorithm}
\DontPrintSemicolon 
\SetKwInput{KwData}{Input}
\SetKwInput{KwResult}{Output}
\KwData{Target passive party  $\theta_u$, target features $f_u \subset f$, learning rate $\eta_2$, current epoch $ep$}
\KwResult{Unlearned passive party model $\theta_u$}
Randomly initialize $\theta_{student}$ with input size of $f \setminus f_u$ \tcp*{$f$ -> feature space of $\theta_u$}
Assign $\theta_u$ as teacher model $\theta_{teacher}$ \\
\For{$t = 1,2,\dots, ep$} {  
    Randomly sample a minibatch $x \in \mathcal{D}$\\
    Compute teacher embedding $emb_{teacher} = H^t_u = \mathcal{G}(\theta_{teacher}, x)$ \\
    Compute student embeddings $emb_{student} = \mathcal{G}(\theta_{student}^t, x\setminus x_u)$ \\
    Compute loss $L = KL\_DIV(emb_{student}, emb_{teacher})$ \\
    Passive party updates $\theta_{student}^{t+1} = \theta_{student}^t - \eta_2\frac{\delta L}{\delta \theta_{student}^t}$
}
\caption{Feature Unlearning for a Passive Party in VFL}
\label{Algo feature unlearning VFL}
\end{algorithm}

With the ever-changing privacy regulations all over the world, passive parties must have the ability to remove/unlearn the influence of controversial features e.g, sensitive features such as gender, ethnicity are unlikely to be used in training the \ac{ML} model anymore. In case of VFL, since passive parties do not have access to the true labels (only their embeddings, say $emb$), computing prediction loss for the student model is not possible without any communication with the active party. Our goal is to minimize or not have communication between active and passive parties for unlearning. Hence, the distillation loss is the overall loss $(L=KL\_DIV(emb_{student}, emb_{teacher}))$ for the student model in this case. Algorithm~\ref{Algo feature unlearning VFL} presents the formal feature unlearning algorithm for VFL. Here as well, we randomly initialize a student model with the input size of new feature space. In each training round, the student model updates its parameter with the $KL\_DIV()$ from its teacher model. And its hessian variation can be written as:
\begin{equation}
    \theta_{student}^{t+1} = \theta_{student}^t - \mathcal{H}_{\theta_{student}}^{-1}\frac{\delta L}{\delta \theta_{student}^t}
\end{equation}
Algorithm~\ref{Algo feature unlearning VFL} does not require any additional storage for embeddings at the passive party as the teacher and student models can compute the embeddings from the randomly sampled minibatch. 

The model parameters in Algorithm \ref{Algo Client unlearning VFL} and~\ref{Algo feature unlearning VFL} are vectors in high-dimensional model space. Consider that the teacher model is in $\mathbb{R}^n$ i.e., $\Theta_{teacher}^t \in \mathbb{R}^n$ and the student model is in $\mathbb{R}^m$ i.e., $\Theta_{student}^t \in \mathbb{R}^m, m<n$. The embedding $H^t\setminus H_u$ lies in the lower dimension manifold of $H^t$. Consequently, the posterior $p(\Theta_{teacher}|H)$ is formed with broader set of features, potentially leading to a more complex model. Similarly, the posterior $p(\Theta_{student}|H \setminus H_u)$ is formed with a reduced set of features, meaning $\Theta_{student}$ would lose some information compared to the $\Theta_{teacher}$ model. With each training epoch, the $KL\_DIV((\hat{y}_{student}|H \setminus H_u)||(\hat{y}_{teacher}|H))$ increases due to the loss of information in $\Theta_{student}$ until the model converges. The $KL\_DIV((\hat{y}_{student}|H \setminus H_u)||(\hat{y}_{teacher}|H)) > 0$ can be lower bounded by some $\delta$, which further can be used to measure the degree of unlearning. Similarly, for feature unlearning, the features $f\setminus f_u$ are in the lower dimension manifold of $f$, and $(KL\_DIV(emb_{student}|f \setminus f_u)||emb_{teacher} | f)$ can be lower bounded by some $\delta_f$ which can be used to measure the degree of unlearning.

\begin{algorithm}
\DontPrintSemicolon 
\SetKwInput{KwData}{Input}
\SetKwInput{KwResult}{Output}
\caption{Sample Unlearning in VFL}
\label{Algo Sample unlearning VFL}
\KwData{Target batch $id$ corresponding tosamples to unlearn, Active party $\Theta_K$, learning rate $\eta_1$, unlearning rate $\lambda$, unlearning epochs $u_{ep}$, current epoch $ep$}
\KwResult{Unlearned Active party model $\Theta_K$}
Assign $\Theta_K$ as teacher model $\Theta_{teacher}$  \\
\For{$t = ep-u_{ep},\dots, ep$} {  
    Active party reads $H$ \\
    Find $H_u = H_{id}$ \tcp*{$H_u$ is the embeddings corresponding to the target samples}
    Update $H = H \setminus H_u$ \\
    Compute prediction $\hat{y}_{retain} = \Theta_K^t(H)$ \\
    Compute prediction $\hat{y}_{target} = \Theta_K^t(H_u)$ \\
    Compute loss $L_{retain} = LossFunction(\hat{y}_{retain}, y_{true})$ \\
    Compute loss $L_{target} = LossFunction(\hat{y}_{target}, y_{true})$ \\
    Active party updates $\Theta_K^{t+1} = \Theta_K^t - \eta_1\frac{\delta L_{retain}}{\delta \Theta_K^t} + \lambda\frac{\delta L_{target}}{\delta \Theta_K^t}$
}
\end{algorithm}

Unlearning samples in \ac{VFL} does not require model compression, making knowledge distillation \ac{KD} a computationally expensive approach for this task, as it involves retraining the model. To make sample unlearning more efficient, we propose using gradient ascent, which significantly reduces training time. Our approach approximately unlearns samples by maximizing the model's loss on the target set (the samples to be unlearned) while minimizing the loss on the remaining dataset (the retain set) over a specified number of epochs.  Algorithm \ref{Algo Sample unlearning VFL} presents the formal algorithm for sample unlearning in \ac{VFL}. Here, the algorithm requires  batch ids corresponding to the samples in target set and the number of unlearning epochs $(u_{ep})$. The unlearning epochs determines how many retraining rounds will be performed. The active party computes the loss on both the target and retain sets, updating its model parameters to maximize the loss on the target set and minimize the loss on the retain set i.e., if $L_{retain}$ and $L_{target}$ is the loss on retain set and target set then the model update can be written as:
\begin{equation}
    \Theta_K^{t+1} = \Theta_K^t - \eta_1\frac{\delta L_{retain}}{\delta \Theta_K^t} + \lambda\frac{\delta L_{target}}{\delta \Theta_K^t}
\end{equation}

When the target set is absent then, the updates simplifies to:
\begin{equation} \label{fine-tune sample unlearniing}
    \Theta_K^{t+1} = \Theta_K^t - \eta_1\frac{\delta L_{retain}}{\delta \Theta_K^t}
\end{equation}
The choice of unlearning rate $\lambda$ plays a crucial role in determining the speed of convergence and stability. A poor choice of $\lambda$ may result in the model parameters getting stuck in local optimum. In comparison with fine-tuning (eq. \ref{fine-tune sample unlearniing}), the model accelerates with $\lambda \frac{\delta L_{target}}{\Theta_K^t}$ to quickly unlearn the updates from target set with Algorithm \ref{Algo Sample unlearning VFL}. The choice of $\lambda$ and $u_{ep}$ can further be used to determine the rate of unlearning.

\subsection{Auditing the unlearning} \label{sec:MIA}

\begin{figure}
    \includegraphics[width=0.75\columnwidth]{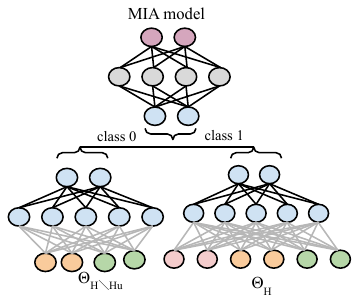}
    \caption{MIA attack model.} \label{fig:MIA unlearning}
\end{figure}

Auditing the unlearning process in \ac{VFL} is especially crucial when institutions with commercial interests are involved. It ensures that proprietary and sensitive information shared among parties are thoroughly removed upon request, thus complying with information privacy regulations such as GDPR. Effective auditing maintains the trust of all participating institutions by verifying that data removal processes are complete and accurate, thereby preventing any residual influence of deleted data on the model's predictions. 

In the literature of unlearning in \ac{FL} data poisoning \cite{deng2023vertical} and backdoor \cite{wu2022federated} attacks are the most common ways to verify unlearning. Specifically, data poisoning attacks involve injecting malicious data into the training process to manipulate the model’s behavior, thereby challenging the unlearning system’s ability to completely remove the influence of such tampered data \cite{deng2023vertical}. On the other hand, backdoor attacks involve embedding hidden triggers within the model, which cause it to behave maliciously when encountering specific inputs, thus providing a rigorous test of whether unlearning mechanisms can entirely eliminate such hidden backdoors \cite{wu2022federated}. However, we argue that auditing unlearning with data poisoning attack and backdoor attacks is not suitable in \ac{VFL} setup. For backdoor attacks, once the embeddings of the target party are unlearned from the active model, the target party can not place the backdoor (for verification) in the new active model. A similar reasoning continues for data poisoning attack along with the absence of true labels in passive parties, thus data poisoning attack is also unsuitable to audit VFL. In our case, we have considered \ac{MIA} (refer section \ref{MIA}) to audit unlearning in \ac{VFL}. 

For MIA, we have a binary classification model which is trained on the output probabilities of the active model in the presence (class $1$) and absence (class $0$) of the target party's embeddings for few epochs (see Fig.~\ref{fig:MIA unlearning}). Once the model is trained, its inference can be used to audit the unlearning of the target party, i.e., whether the target party participated in training the active model or not. Similarly, for auditing sample unlearning, the MIA model is trained with the output probabilities in the presence (class $1$) and absence (class $0$) of the samples to unlearn.

\section{Experimental analysis} \label{Experiments}

\begin{figure*}[t]
    \centering
    \subfloat[Adult]{
        \includegraphics[width=0.32\linewidth]{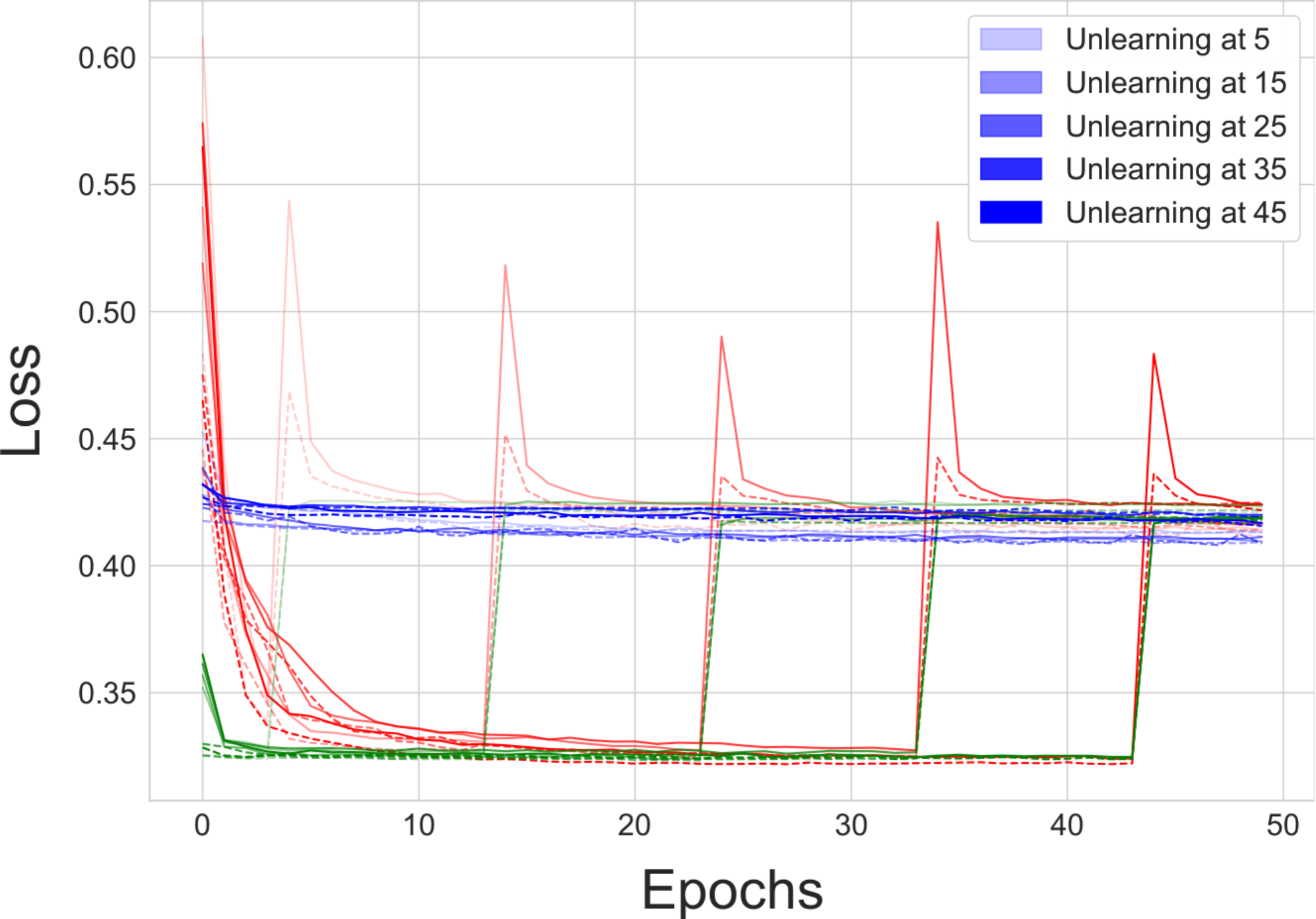}
        \label{adult_loss}
    }
    \hfill
    \subfloat[ai4i]{
        \includegraphics[width=0.32\linewidth]{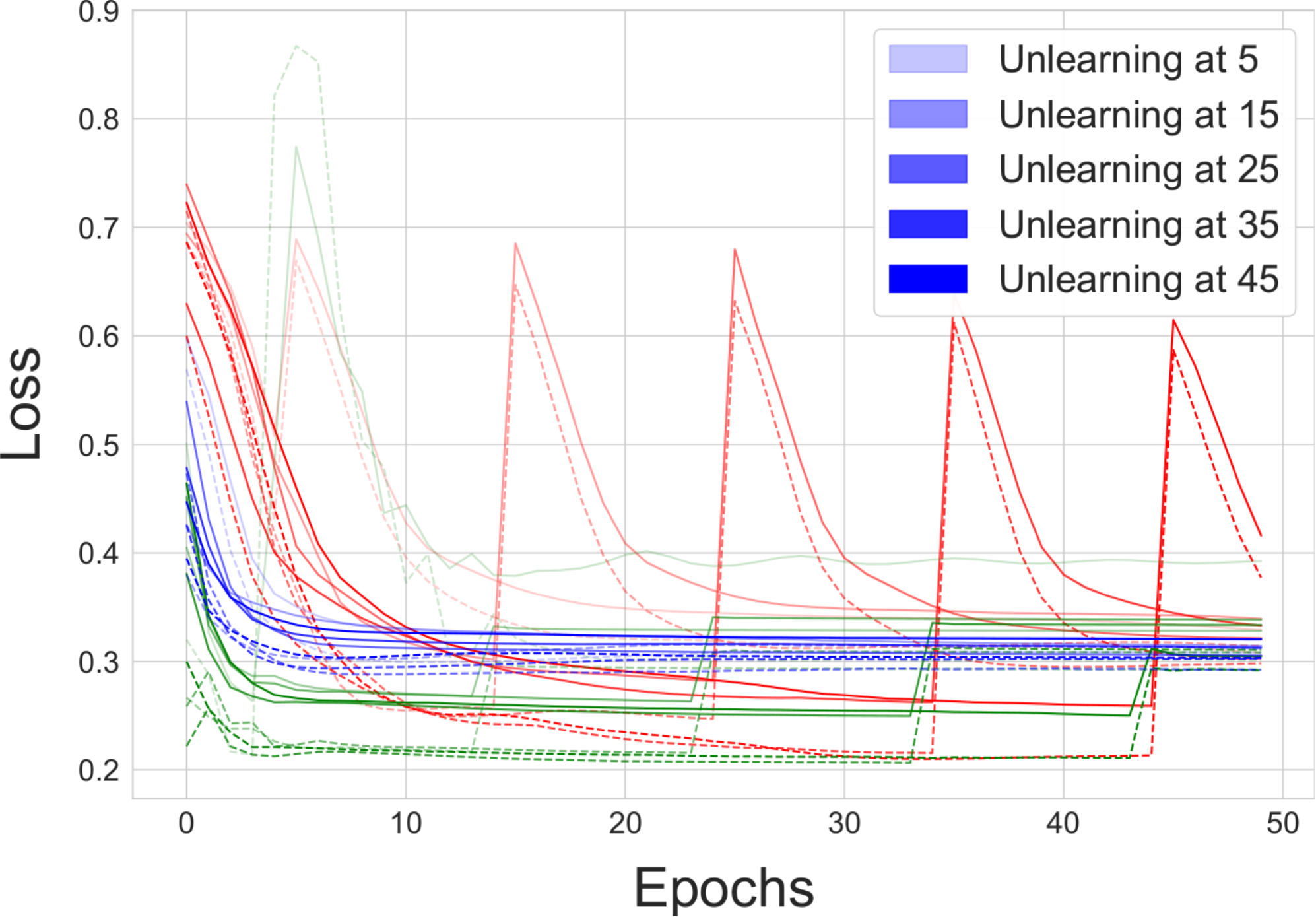}
        \label{ai4i_loss}
    }
    \hfill
    \subfloat[Hepmass]{
        \includegraphics[width=0.32\linewidth]{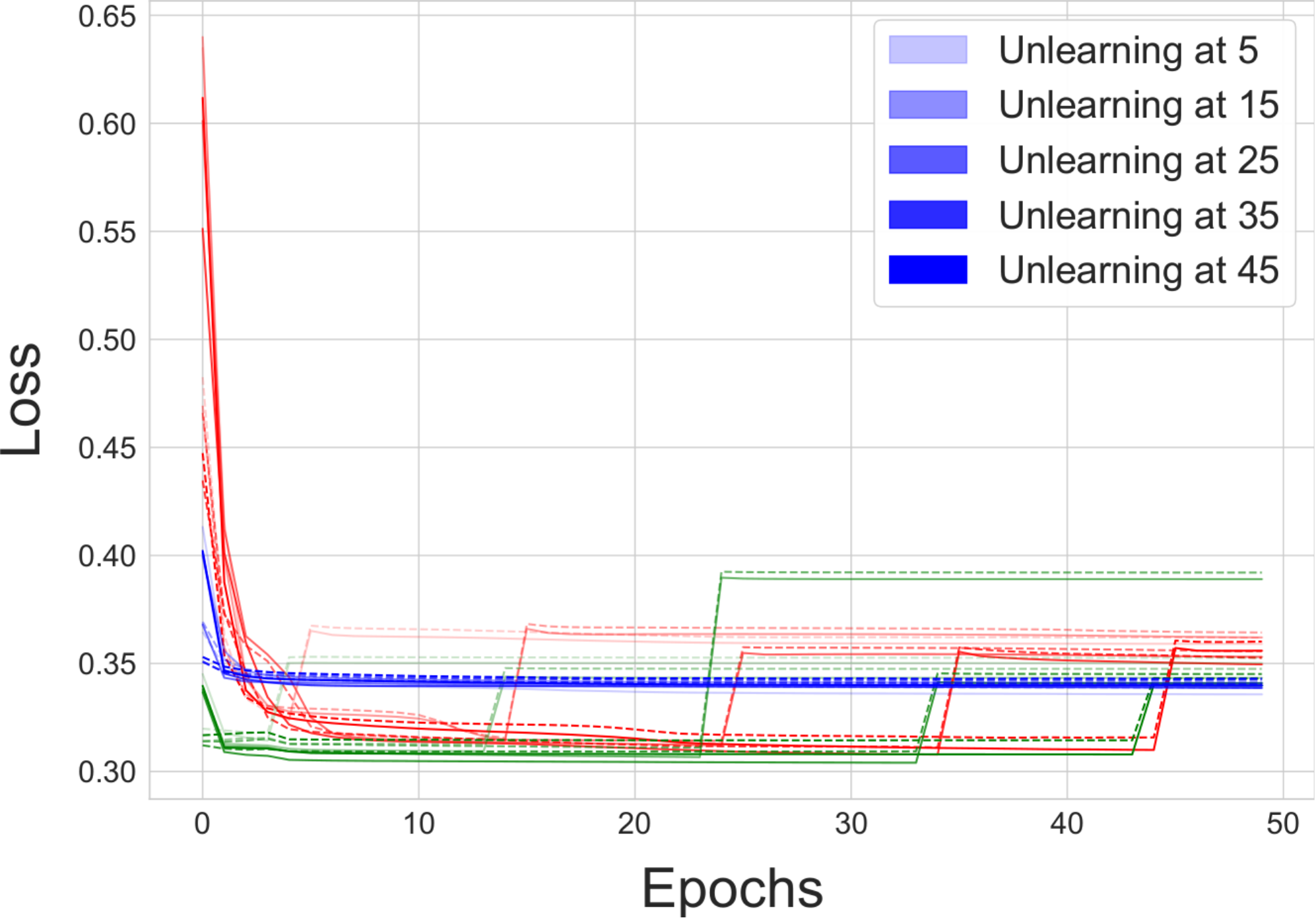}
        \label{hepmass_loss}
    }
    \hfill
    \subfloat[Poqemon]{
        \includegraphics[width=0.32\linewidth]{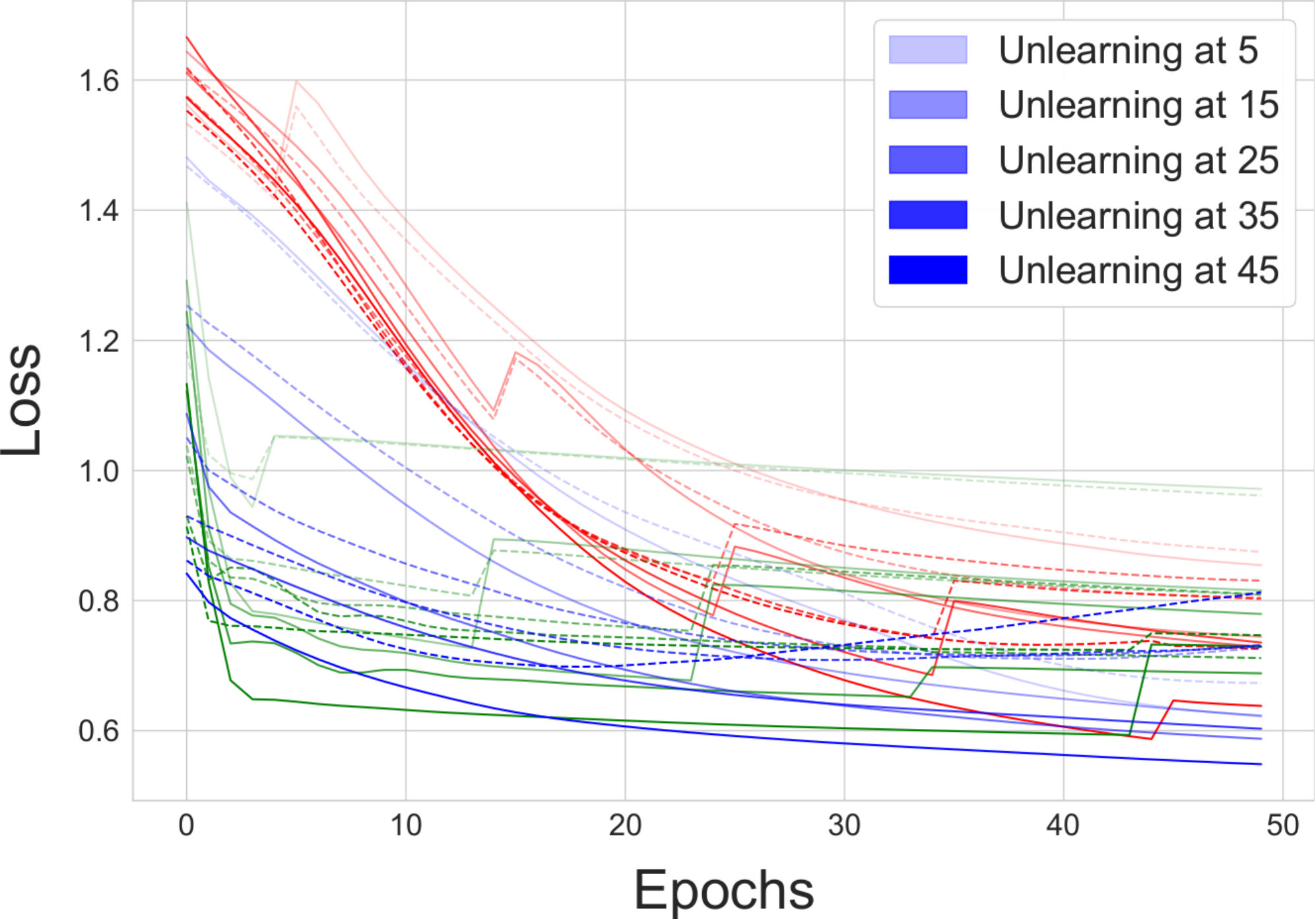}
        \label{poqemon_loss}
    }
    \hfill
    \subfloat[Susy]{
        \includegraphics[width=0.32\linewidth]{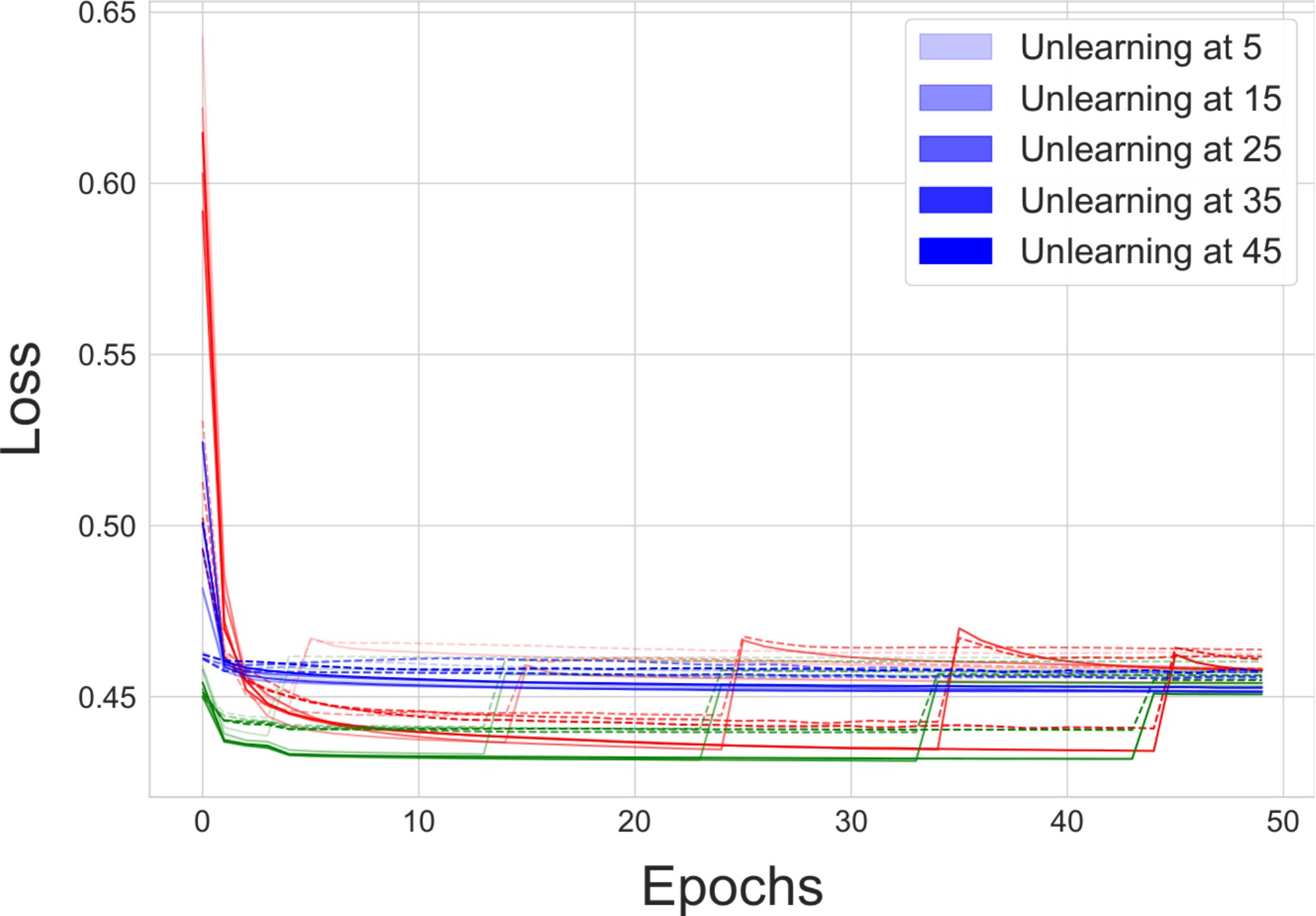}
        \label{susy_loss}
    }
    \hfill
    \subfloat[Wine]{
    \includegraphics[width=0.3\linewidth]{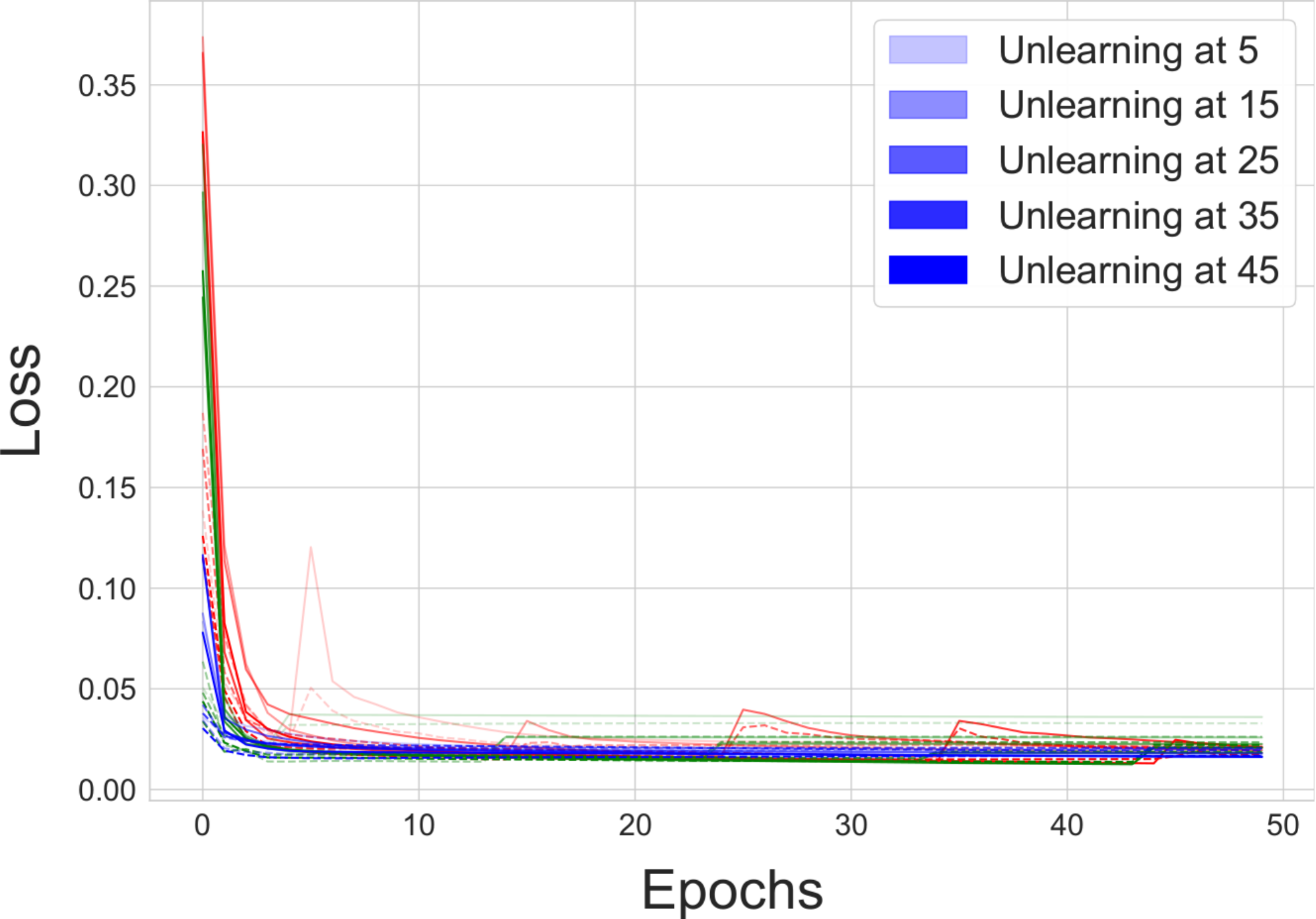}
        \label{wine_loss}
    }
    \hfill
    \subfloat{
    \includegraphics[width=0.5\linewidth]{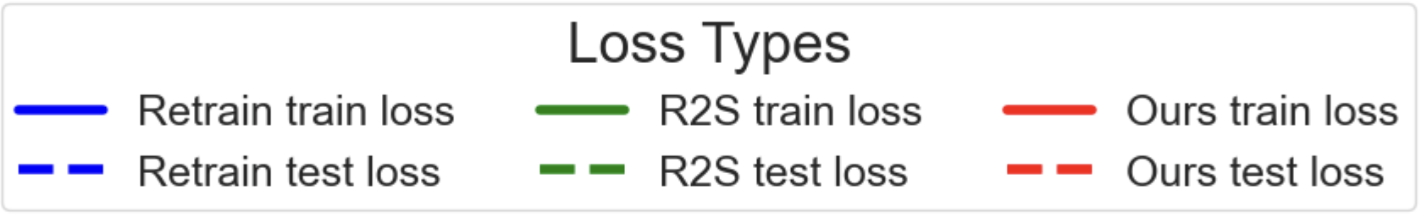}
        \label{additional legend}
    }
    \caption{The training and test loss of VFU-KD compared to the retrained model from scratch and R2S method.} 
    \label{loss_values_scores} 
\end{figure*}

\begin{table*}[h]
    \centering
    \begin{tabular}{c|ccc|ccc|ccc|ccc|ccc|ccc}
        \toprule
         \multirow{2}{*}{Epoch} & \multicolumn{3}{c|}{Adult} & \multicolumn{3}{c|}{{ai4i}} & \multicolumn{3}{c|}{{Hepmass}} & \multicolumn{3}{c|}{{Poqemon}} & \multicolumn{3}{c|}{{Susy}} & \multicolumn{3}{c}{{Wine}} \\
         & {RfS} & {R2S} & {Ours} & {RfS} & {R2S} & {Ours} & {RfS} & {R2S} & {Ours} & {RfS} & {R2S} & {Ours} & {RfS} & {R2S} & {Ours} & {RfS} & {R2S} & {Ours} \\
         \toprule 
         {5th Ep.} & {0.65} & {0.62} & {\textbf{0.65}} & {0.87} & {0.87} & {\textbf{0.88}} & {0.83} & {0.82} & {0.82} & {0.94} & {0.86} & {0.88} & {0.78} & {0.56} & {\textbf{0.78}} & {0.98} & {0.96} & {0.97} \\ 
         {15th Ep.} & {0.65} & {0.63} & {0.64} & {0.88} & {0.89} & {\textbf{0.89}} & {0.83} & {0.83} & {0.82} & {0.94} & {0.92} & {0.91} & {0.78} & {0.78} & {\textbf{0.78}} & {0.98} & {0.97} & {\textbf{0.98}} \\
         {25th Ep.} & {0.66} & {0.63} & {\textbf{0.66}} & {0.89} & {0.89} & {\textbf{0.89}} & {0.83} & {0.81} & {\textbf{0.83}} & {0.94} & {0.91} & {0.92} & {0.78} & {0.78} & {\textbf{0.78}} & {0.98} & {0.97} & {\textbf{0.98}} \\
         {35th Ep.} & {0.64} & {0.64} & {0.63} & {0.87} & {0.88} & {\textbf{0.89}} & {0.83} & {0.83} & {\textbf{0.83}} & {0.94} & {0.93} & {0.92} & {0.78} & {0.78} & {\textbf{0.78}} & {0.98} & {0.97} & {\textbf{0.98}} \\ 
         {45th Ep.} & {0.64} & {0.64} & {0.63} & {0.87} & {0.87} & {0.83} & {0.83} & {0.83} & {0.82} & {0.94} & {0.93} & {0.93} & {0.78} & {0.78} & {\textbf{0.78}} & {0.98} & {0.97} & {\textbf{0.98}} \\
         \bottomrule
    \end{tabular}
    \caption{The AUC score comparison of VFU-KD (Ours) with the retrained-from-scratch (RfS) model and the R2S method.}
    \label{tab:VFU-KD_unl_epchs_AUC}
\end{table*}

\begin{table*}[h]
    \centering
    \begin{tabular}{c|ccc|ccc|ccc|ccc|ccc|ccc}
        \toprule
         \multirow{2}{*}{{Epoch}} & \multicolumn{3}{c|}{{Adult}} & \multicolumn{3}{c|}{{ai4i}} & \multicolumn{3}{c|}{{Hepmass}} & \multicolumn{3}{c|}{{Poqemon}} & \multicolumn{3}{c|}{{Susy}} & \multicolumn{3}{c}{{Wine}} \\
         & {RfS} & {R2S} & {Ours} & {RfS} & {R2S} & {Ours} & {RfS} & {R2S} & {Ours} & {RfS} & {R2S} & {Ours} & {RfS} & {R2S} & {Ours} & {RfS} & {R2S} & {Ours} \\
         \toprule 
         {5th Ep.} & {0.77} & {0.75} & {\textbf{0.77}} & {0.88} & {0.90} & {0.87} & {0.83} & {0.82} & {0.82} & {0.75} & {0.56} & {0.54} & {0.78} & {0.43} & {\textbf{0.78}} & {0.99} & {0.98} & {0.98} \\ 
         {15th Ep.} & {0.77} & {0.76} & {\textbf{0.77}} & {0.88} & {0.89} & {0.87} & {0.83} & {0.83} & {0.82} & {0.76} & {0.67} & {0.70} & {0.78} & {0.78} & {\textbf{0.78}} & {0.99} & {0.99} & {\textbf{0.99}} \\
         {25th Ep.} & {0.77} & {0.75} & {\textbf{0.77}} & {0.88} & {0.88} & {\textbf{0.88}} & {0.83} & {0.81} & {0.82} & {0.77} & {0.60} & {0.70} & {0.78} & {0.78} & {\textbf{0.78}} & {0.99} & {0.98} & {\textbf{0.99}} \\
         {35th Ep.} & {0.76} & {0.76} & {\textbf{0.76}} & {0.88} & {0.89} & {\textbf{0.89}} & {0.83} & {0.83} & {0.82} & {0.77} & {0.72} & {0.67} & {0.79} & {0.78} & {0.78} & {0.99} & {0.99} & {\textbf{0.99}} \\ 
         {45th Ep.} & {0.77} & {0.76} & {0.76} & {0.88} & {0.88} & {0.87} & {0.83} & {0.83} & {0.82} & {0.77} & {0.74} & {0.72} & {0.79} & {0.79} & {0.78} & {0.99} & {0.99} & {\textbf{0.99}} \\
         \bottomrule
    \end{tabular}
    \caption{The F1 score comparison of VFU-KD (Ours) with the retrained-from-scratch (RfS) model and the R2S method.}
    \label{tab:VFU-KD_unl_epchs_F1}
\end{table*}

In this section, we present the experimental analysis of our proposed unlearning framework. As discussed in Section \ref{VFL}, the most common \ac{VFL} setting typically involves two parties, with a maximum of four parties. For this paper, we consider a three-party \ac{VFL} setup consisting of $clientA$, $clientB$, and $clientC$, collaborating to train a joint \ac{VFL} model. The training process spans 50 epochs, with $clientA$ having the flexibility to request unlearning at any point during the training process. Since, communication in \ac{VFL} is communication-intensive, using a larger batch size is preferred to optimize efficiency~\cite{gao2024complementary}. Accordingly, we set the batch size to 512 in our experiments. The learning rate is set to $10^{-2}$ for tabular data and $10^{-3}$ for image data for both active and passive parties. Additionally, the distillation parameter which controls the trade-off between actual loss and distillation loss is set to $0.3$. To demonstrate the effectiveness of our approach in unlearning at any stage during training, we conduct experiments at various epochs: $[5^{th}, 15^{th}, 25^{th}, 35^{th}, 45^{th}]$. After confirming the effectiveness and feasibility of our unlearning method, we fix the unlearning epoch at the $25^{th}$ epoch for further evaluation. Each experiment is then repeated three times to account for variability and capture uncertainty in the results.


In our experiments, we have considered 6 tabular datasets, namely Adult, ai4i, hepmass, susy, and wine dataset from UCI repository \cite{kelly2023uci}, poqemon dataset \cite{amour2015building} and 2 image dataset CIFAR10 \cite{krizhevsky2009cifar-10} and STL10 dataset from \cite{coates2011analysis}. For tabular datasets, passive models have a single hidden layer models with 8 hidden neurons, and active model is also a single hidden layer model with 32 neurons followed by an output layer. The features are distributed equally among passive parties. For example, wine dataset has 12 features, $clientA$ has first 4, $clientB$ has next 4 and $clientC$ has last 4 features. For image datasets, clients have resnet-18 model as the passive model, and the active model is a single hidden layer model with 512 neurons followed by an output layer. The number of neurons in output layer for active model depends on the output classes of the datasets, e.g., 2 for adult, 5 for poqemon, 10 for CIFAR10. 

It is important to emphasize that these experimental setups are not optimized for peak performance. Our primary objective is to showcase the effects of unlearning on both tabular and image datasets. The parameters were selected arbitrarily and are not finetuned for optimal results on each dataset.  Fine-tuning the experimental configurations for optimal results on each dataset is out-of-scope for this work.

\subsection{Passive party unlearning}

We compare our approach with the gold standard i.e., a retrained model from scratch and a benchmark R2S fast retraining  unlearning approach \cite{wang2024efficient}. For passive party unlearning, the R2S method is equivalent to retraining from scratch with smarter optimizer. Specifically, the R2S method switches between RAdam and SGD with momentum based on the training epoch and the predefined threshold. In our approach, we train the model using the RAdam optimizer \cite{liu2019variance}. Fig. \ref{loss_values_scores} shows the training and test loss throughout the learning process for the tabular datasets at $[5^{th}, 15^{th}, 25^{th}, 35^{th}, 45^{th}]$ epochs (Ep.). The shading indicates the epoch at which unlearning occurs: lighter shades represent earlier unlearning epochs (e.g., unlearning at the $5^{th}$ Ep.), while darker shades represent later epochs (e.g., unlearning at the $45^{th}$ Ep.). Results for retraining from scratch are depicted with blue lines, the R2S method is shown with green lines, and our approach is represented with red lines. It can be clearly seen from Fig.~\ref{loss_values_scores} that, after unlearning at each unlearning epoch, there is a spike in the loss values for all the datasets. The spike can be attributed to the distillation process inherent in the unlearning procedure. However, the spike is not significant and the loss curve is comparable to the benchmark models,  i.e., retrained model and R2S model, in all the cases. The results in Table \ref{tab:VFU-KD_unl_epchs_AUC} and Table ~\ref{tab:VFU-KD_unl_epchs_F1} indicate that, for VFU-KD, the utility score (area under the curve (AUC score) and F1 score) after unlearning at the $50^{th}$ epoch is either similar to or better than that of retraining from scratch (abbreviated as RfS, due to limited space) and the R2S method in many cases (highlighted in bold). In the remaining cases, the utility scores are comparable, with a utility loss ranging between 1-5\%. The observed improvement in loss values and utility scores may be attributed to the negative impact of $clientA$ on the training of the active model. This explanation is further supported by the observation that the retrained model achieves better scores than the model trained with all three clients. Notably, our approach is not limited to the use of the RAdam optimizer; it can be seamlessly integrated with any advanced optimizer to further enhance performance. However, for the purpose of benchmarking against state-of-the-art (SOTA) methods, we specifically compare our approach using the RAdam optimizer to a retraining strategy based on the R2S optimizer.
These benchmark include extensive communication with the clients. Table \ref{tab:additional_VFU-R2S communication cost} shows significant additional communication cost required while unlearning, specially at the later epochs. This is also under an assumption that passive parties are willing to collaborate to train a new model and this also exposes the unlearn party for further privacy attacks \cite{wang2023federated}.

\begin{figure}[t]
    \centering
    \subfloat[Adult]{
        \includegraphics[width=0.48\columnwidth]{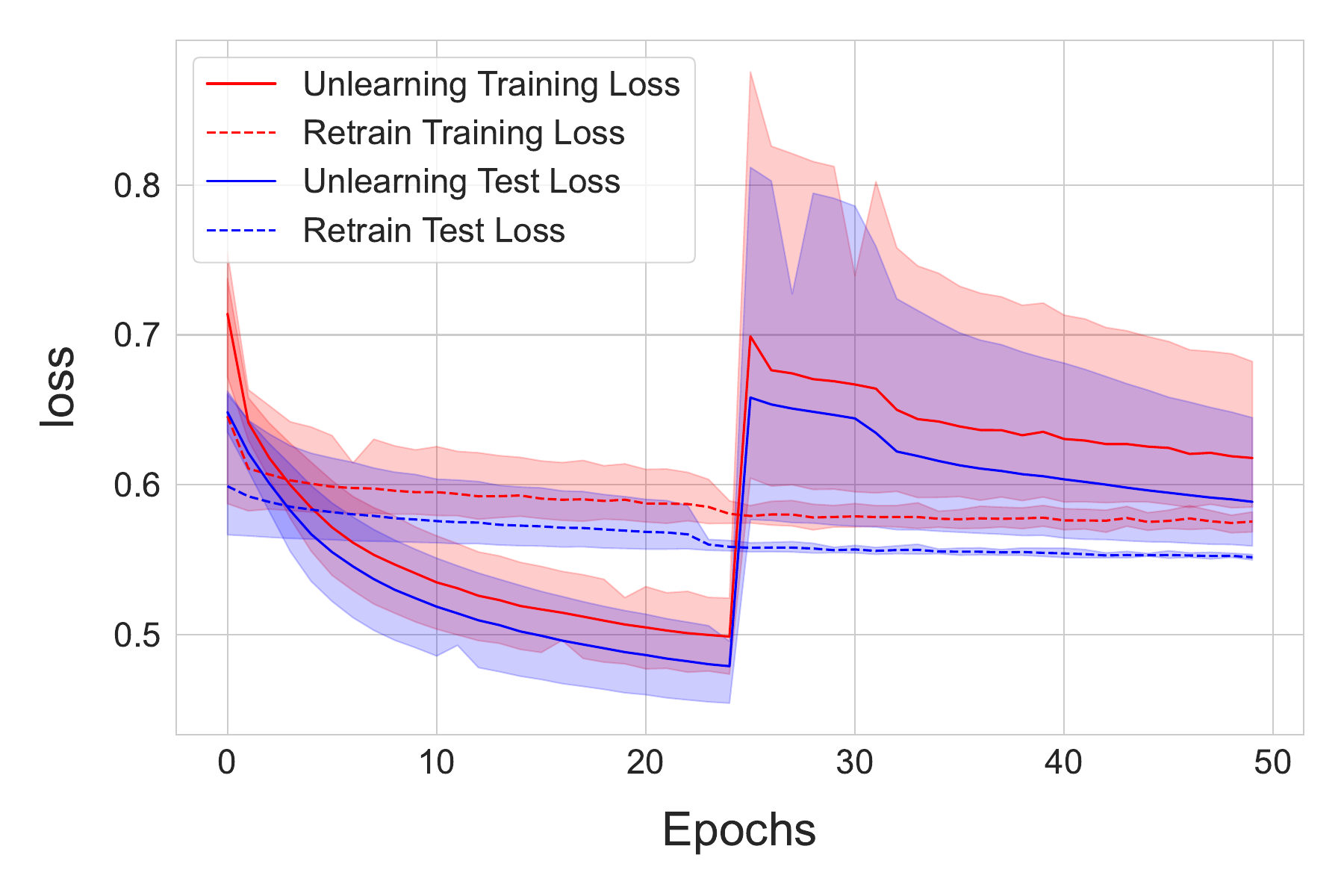}
        \label{adult_loss_hess}
    }
    \hfill
    \subfloat[ai4i]{
        \includegraphics[width=0.48\columnwidth]{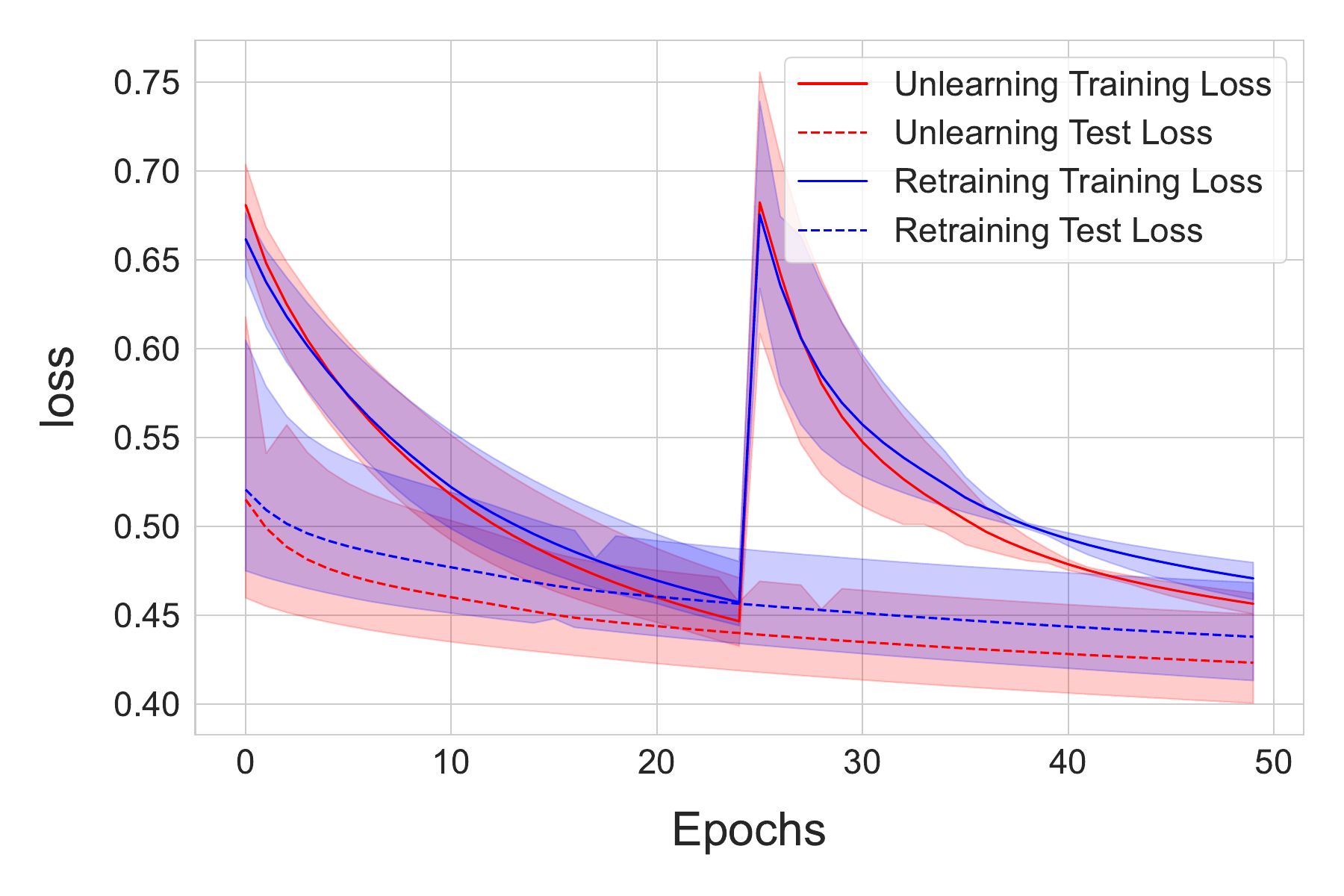}
        \label{ai4i_loss_hess}
    }
    \hfill
    \subfloat[Hepmass]{
        \includegraphics[width=0.48\columnwidth]{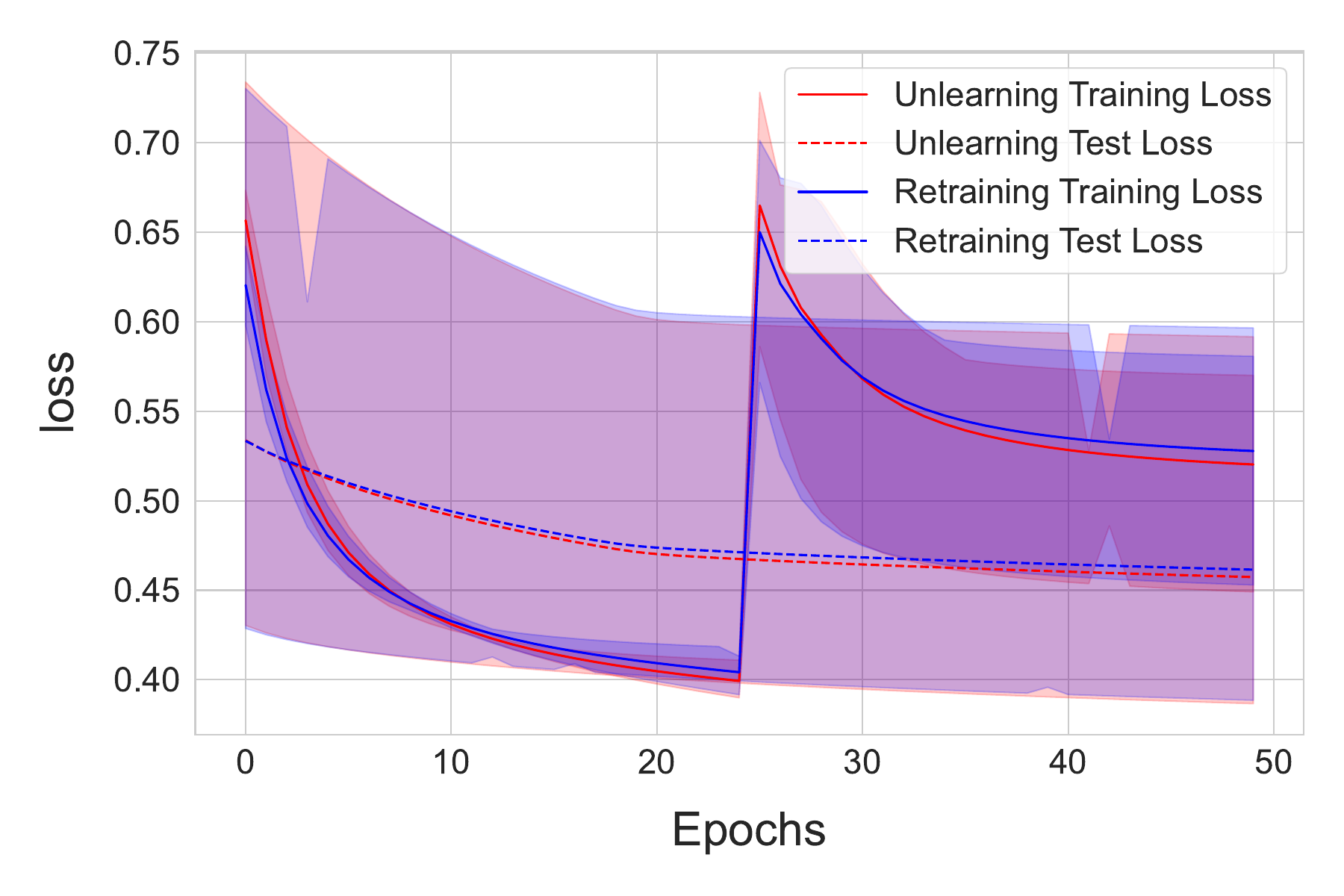}
        \label{hepmass_loss_hess}
    }
    \hfill
    \subfloat[Poqemon]{
        \includegraphics[width=0.48\columnwidth]{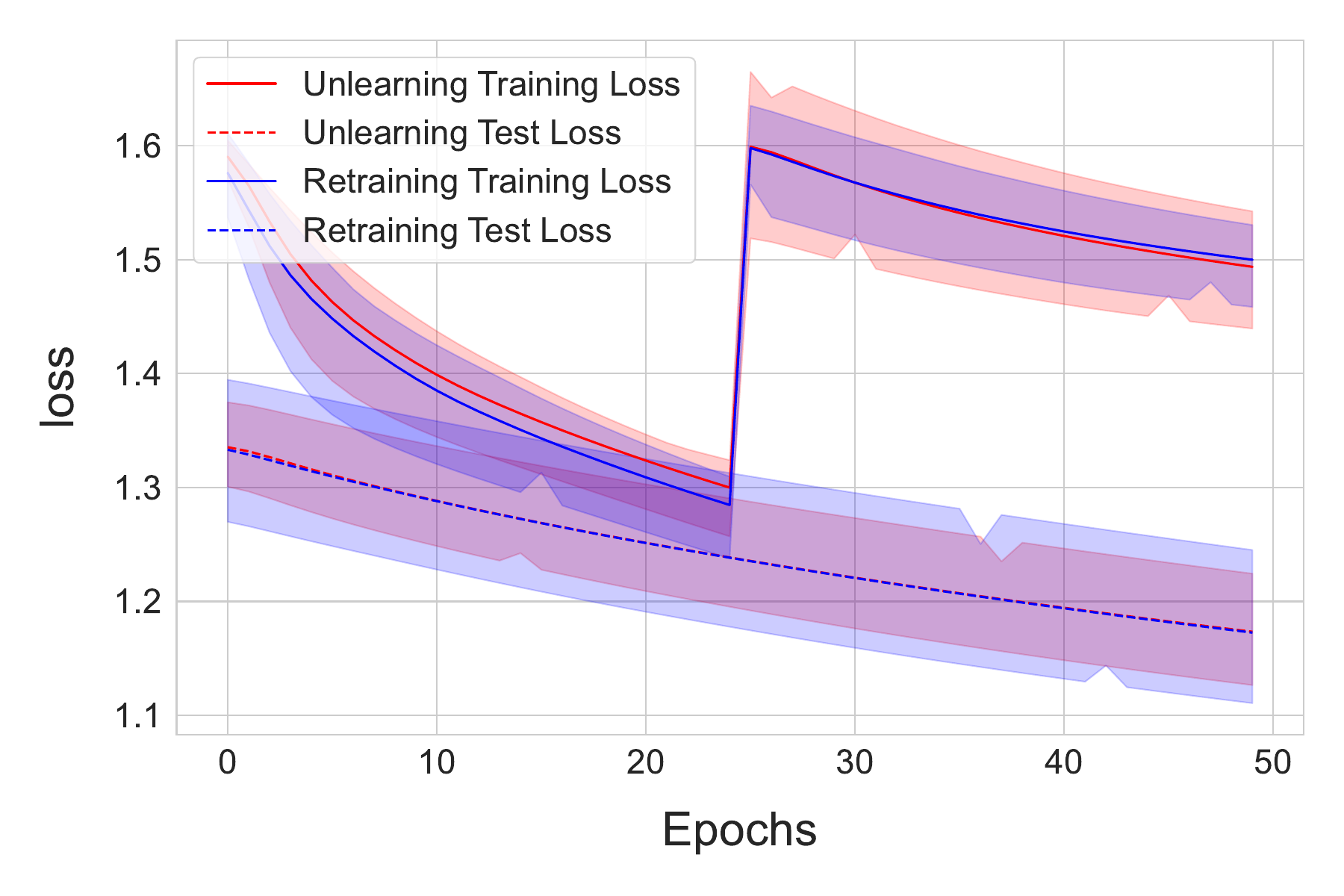}
        \label{poqemon_loss_hess}
    }
    \hfill
    \subfloat[Susy]{
        \includegraphics[width=0.48\columnwidth]{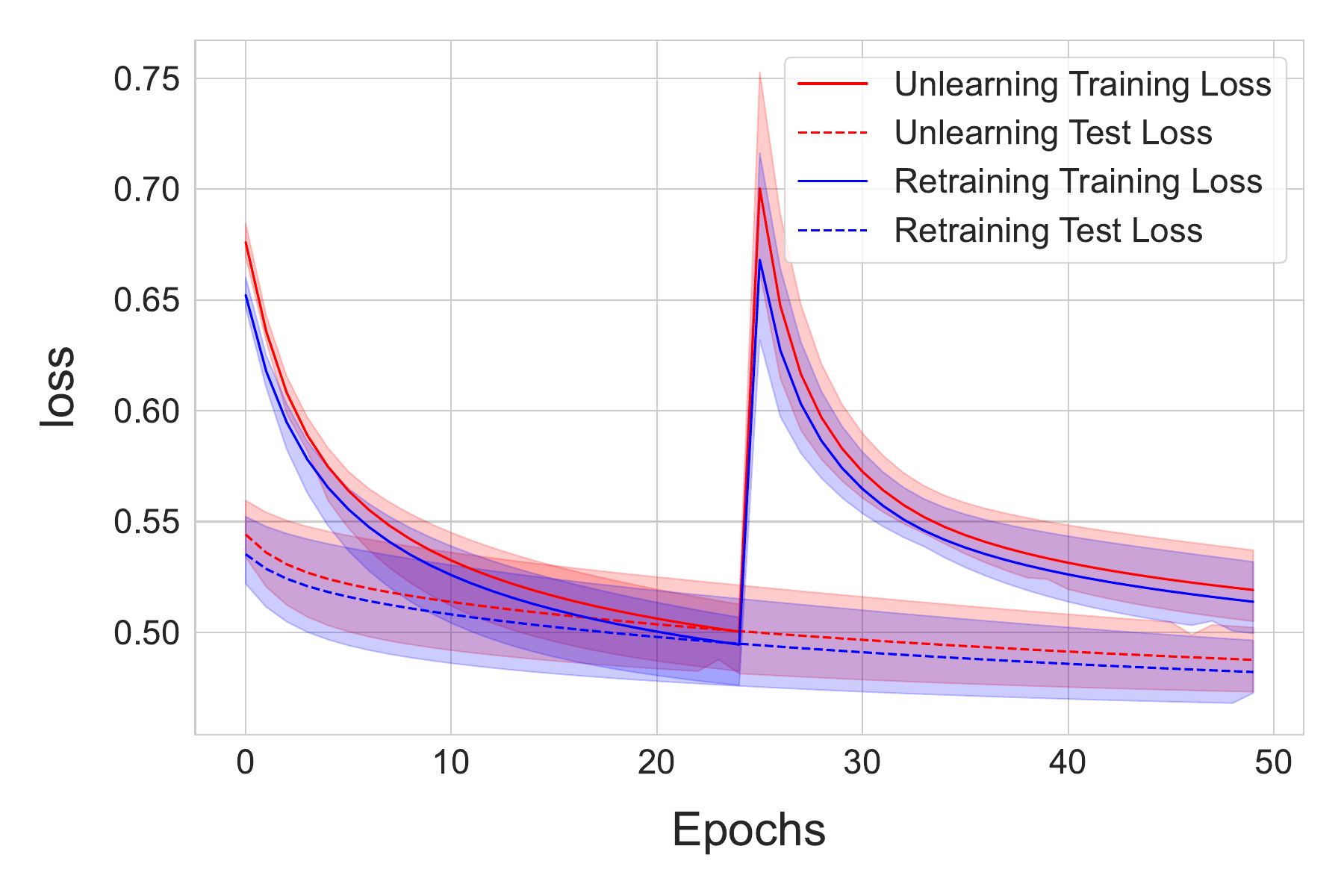}
        \label{susy_loss_hess}
    }
    \hfill
    \subfloat[Wine]{
        \includegraphics[width=0.48\columnwidth]{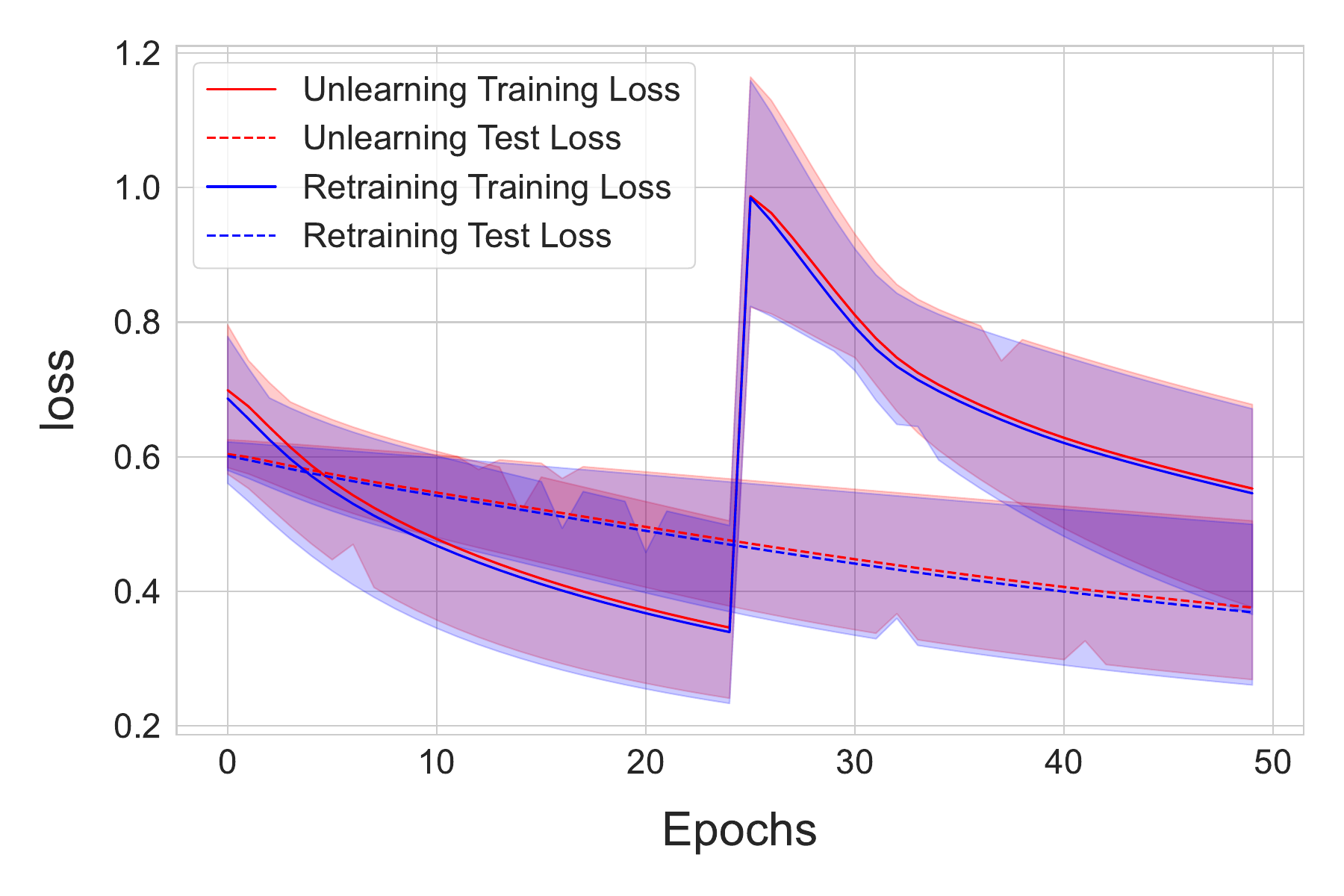}
        \label{wine_loss_hess}
    }
    \caption{The training (red) and test loss (blue) of VFU-KD (solid lines) with $\mathcal{H}^{-1}$ compared to the retrained model from scratch (dotted lines).} 
    \label{loss_values_scores_hessian} 
\end{figure}

\begin{table}[]
    \centering
    \begin{tabular}{cccccc}
        \toprule
         Dataset & 5th Ep. & 15th Ep. & 25th Ep. & 35th Ep. & 45th Ep. \\
         \midrule
         Adult & {33.4} & {98.6} & {164.3} & {229.2} & {295.0} \\
         {ai4i }& {5.4 }& {16.3} & {27.2} & {38.0} & {48.9} \\
         {Hepmass} & {12.6 }& {37.8 }& {63.0 }& {88.2 } & {11.3 }\\
         {Poqemon} & {7.3 }& {22 }& {36.7} & {51.4} & {66.1} \\
         {Susy} & {27.2} & {81.6} & {136 }& {190.4} & {244.8} \\
         {Wine} & {3.5} & {10.6} & {17.6} & {24.7} & {31.8} \\
        \bottomrule
    \end{tabular}
    \caption{The table presents the additional communication cost in gigabytes (GB) for unlearning with R2S method against VFU-KD.}
    \label{tab:additional_VFU-R2S communication cost}
\end{table}

\begin{figure}[h]
    \centering
    \subfloat[CIFAR10-loss]{
        \includegraphics[width=0.48\columnwidth]{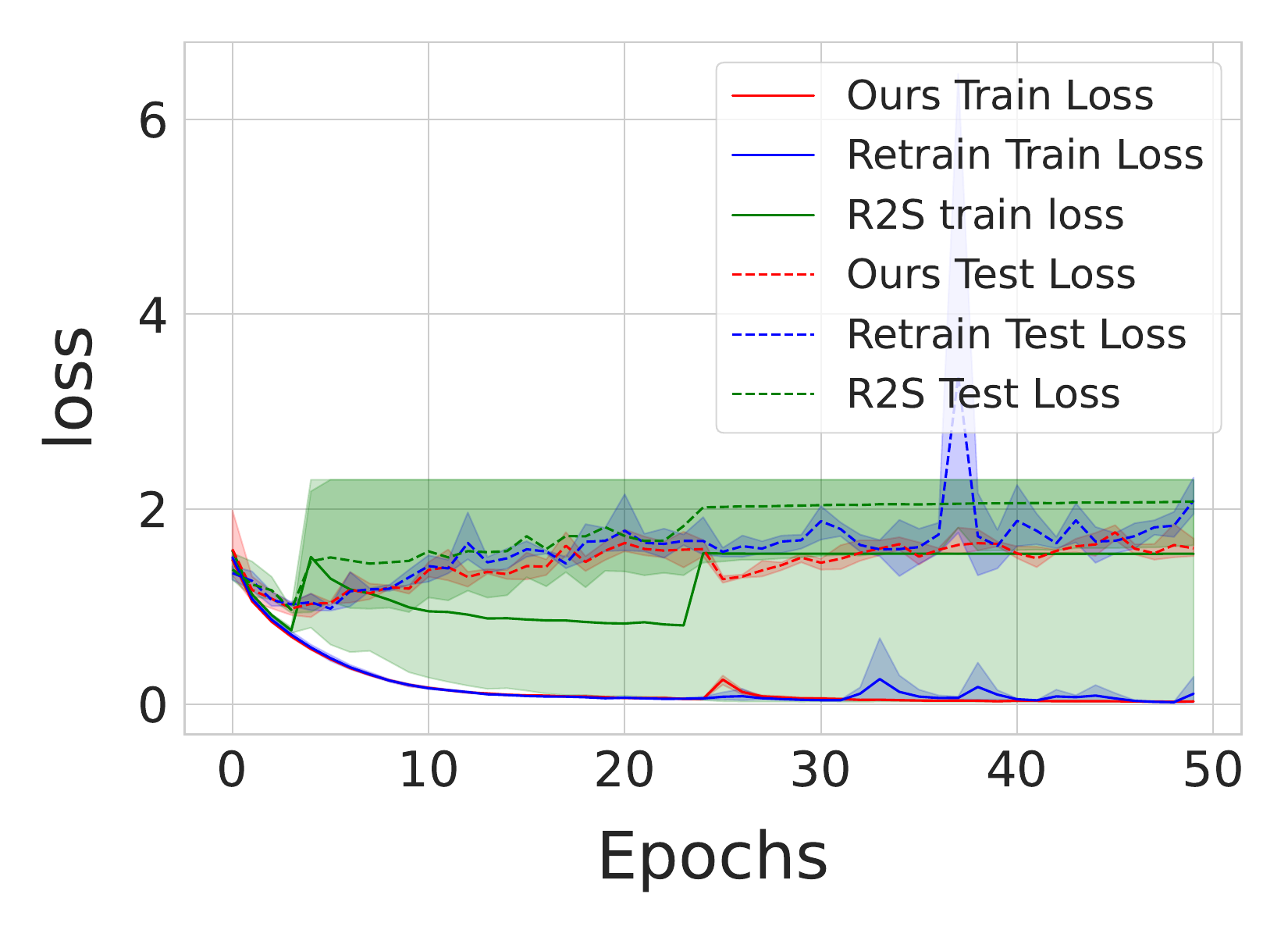}
        \label{cifar10_loss}
    }
    \hfill
    \subfloat[STL10-loss]{
        \includegraphics[width=0.48\columnwidth]{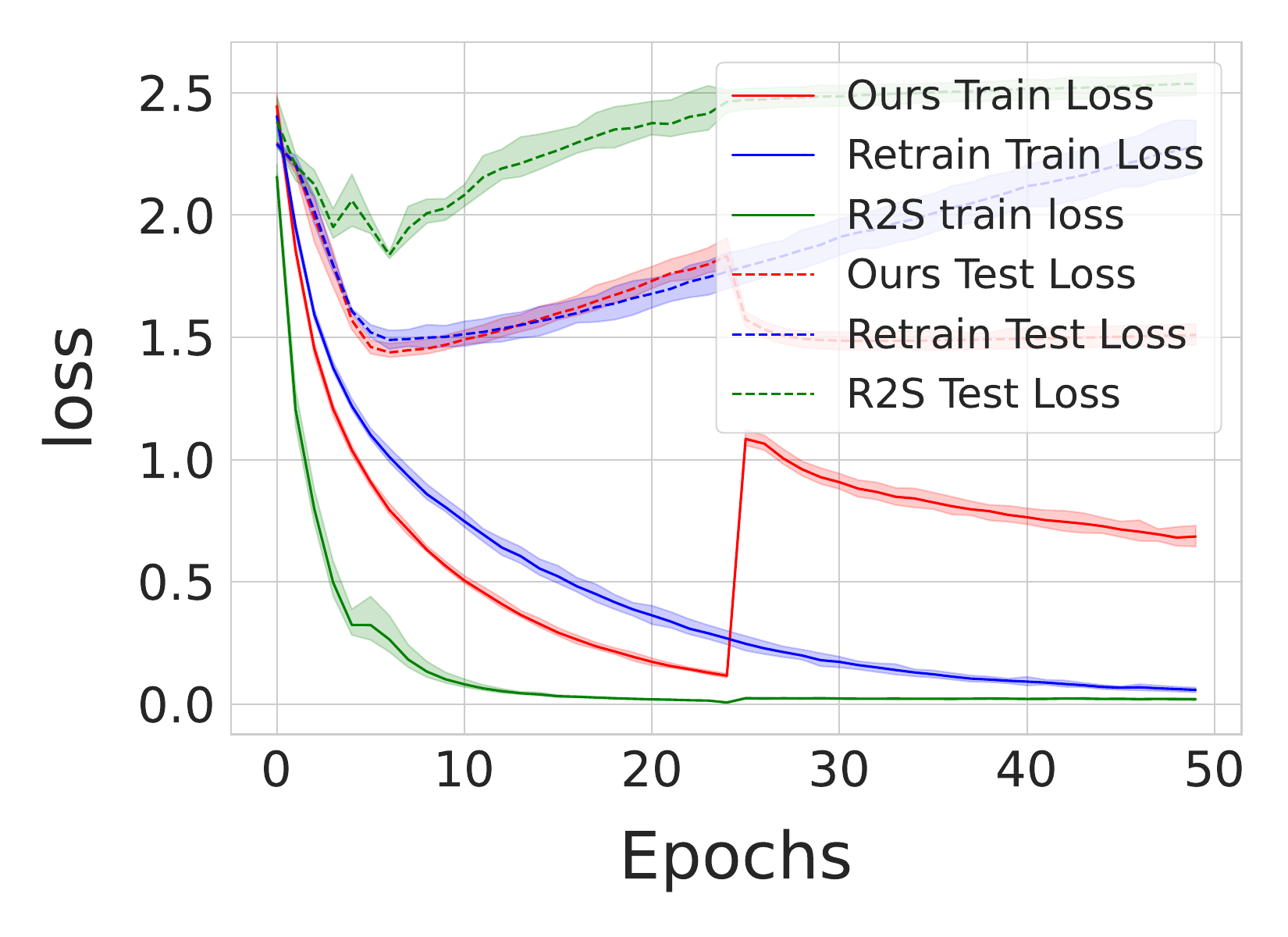}
        \label{stl10_loss}
    }
    \caption{The loss curves of VFU-KD compared to the retrained model from scratch and R2S method.}
    \label{images_results}
\end{figure}

\begin{table}[]
    \centering
    \begin{tabular}{c|cc|cc|cc}
        \toprule
         \multirow{2}{*}{{
         Dataset}} & \multicolumn{2}{c|}{{RfS}} & \multicolumn{2}{c|}{{R2S}} & \multicolumn{2}{c}{{Ours}}\\
         & {AUC} & {F1} & {AUC} & {F1} & {AUC} & {F1} \\
         \midrule
         {CIFAR10} & {0.92} & {0.70} & {0.68} & {0.29} & {\textbf{0.92}} & {\textbf{0.70}} \\
         {STL10} & {0.89} & {0.49} & {0.91} & {0.52} & {\textbf{0.89}} & {\textbf{0.49}} \\
        \bottomrule
    \end{tabular}
    \caption{The AUC and F1 score comparison of VFU-KD (Ours) with the retrained-from-scratch (RfS) model and the R2S method.}
    \label{tab:VFU-img_res}
\end{table}

\begin{figure}[!ht]
    \centering
    \subfloat[Adult]{
        \includegraphics[width=0.48\columnwidth]{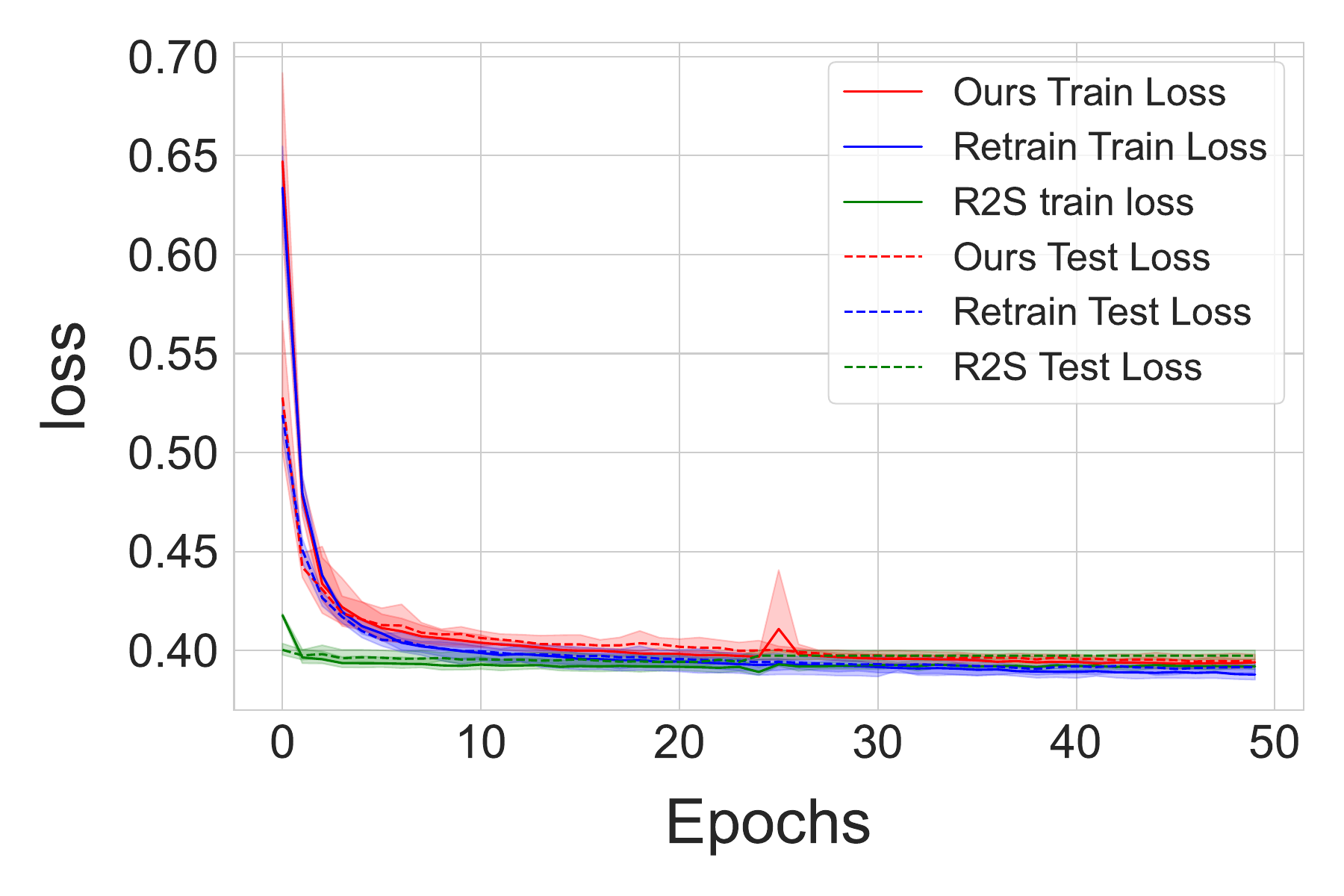}
        \label{adult_loss_MI}
    }
    \hfill
    \subfloat[ai4i]{
        \includegraphics[width=0.48\columnwidth]{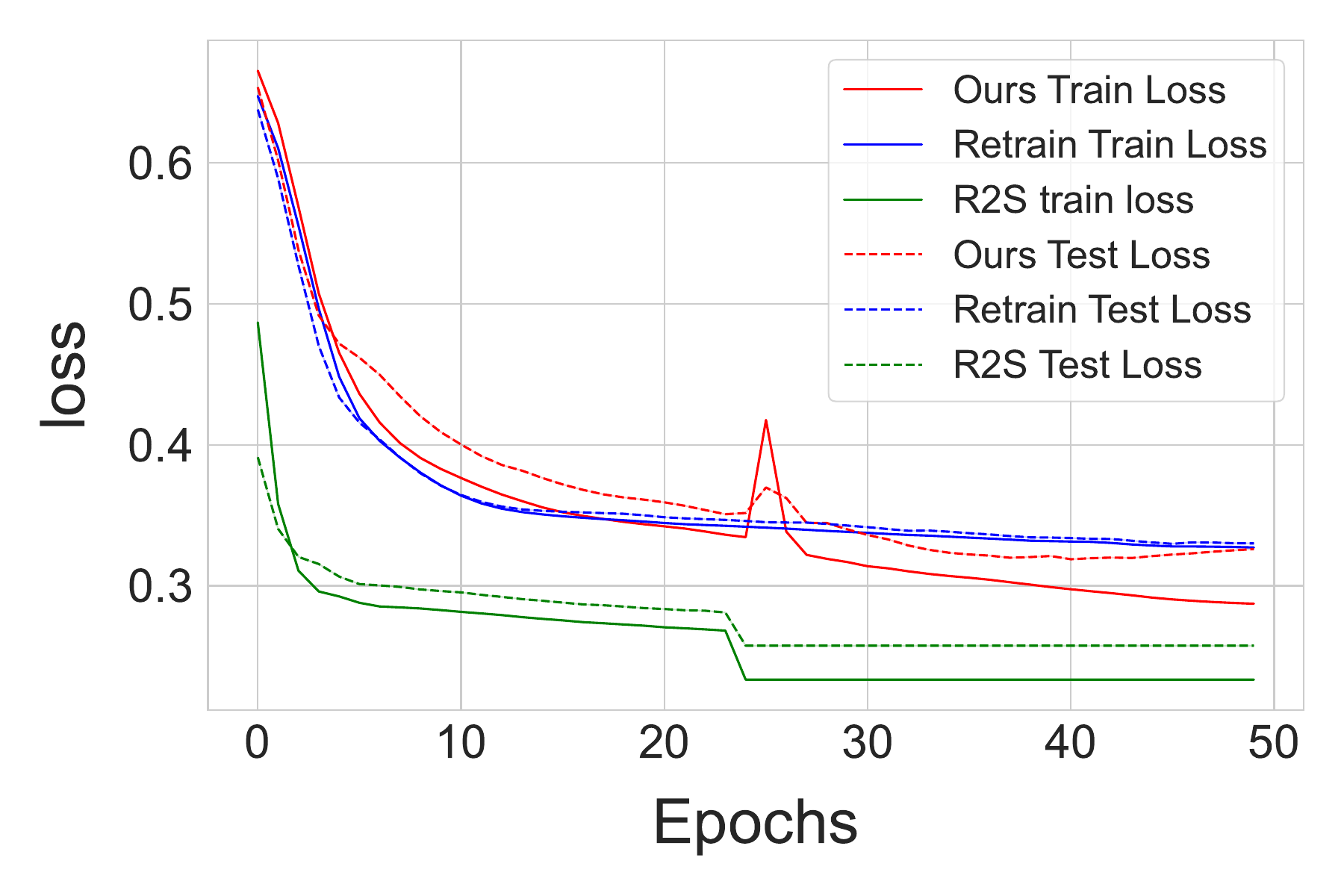}
        \label{ai4i_loss_MI}
    }
    \hfill
    \subfloat[Hepmass]{
        \includegraphics[width=0.48\columnwidth]{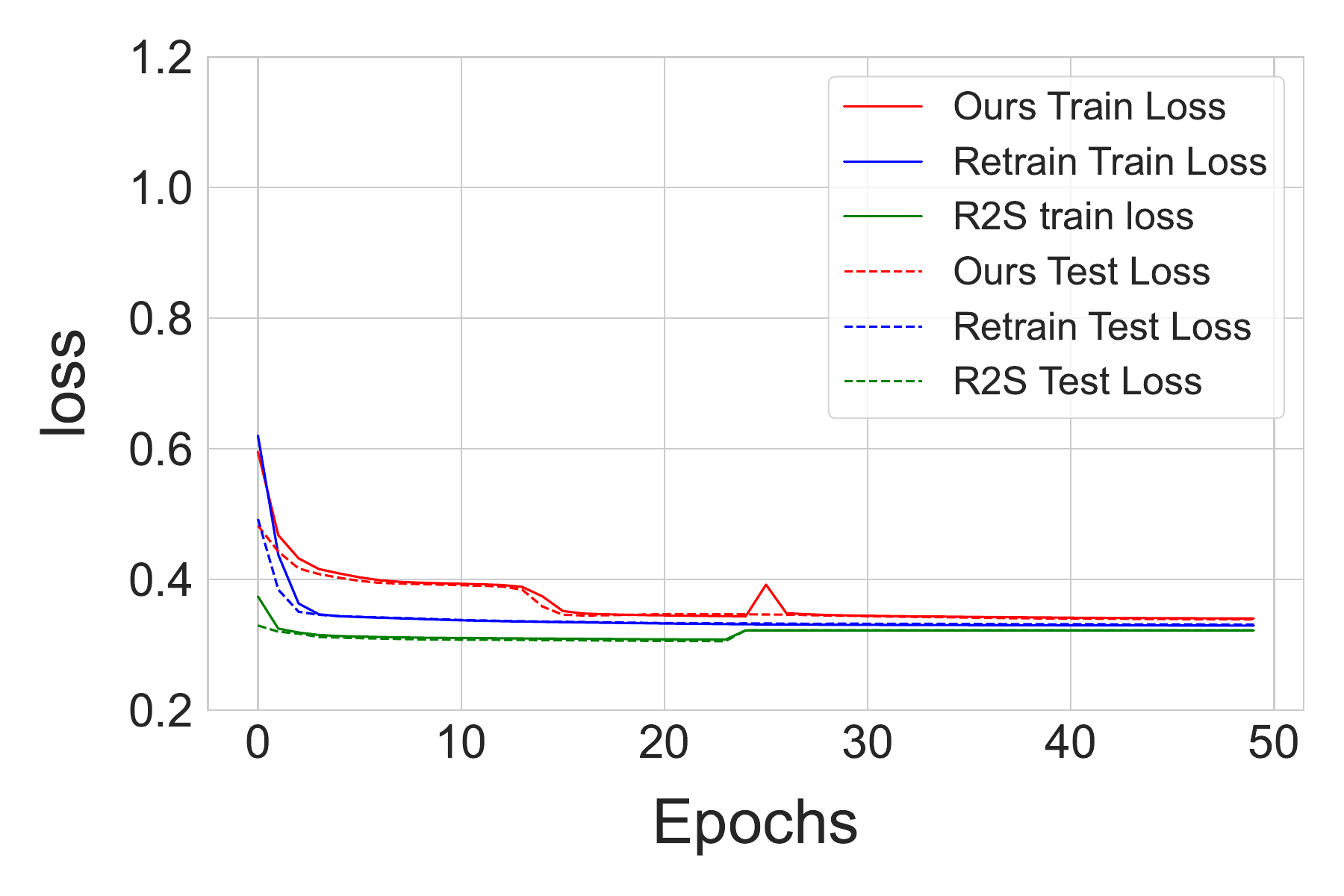}
        \label{hepmass_loss_MI}
    }
    \hfill
    \subfloat[Poqemon]{
        \includegraphics[width=0.48\columnwidth]{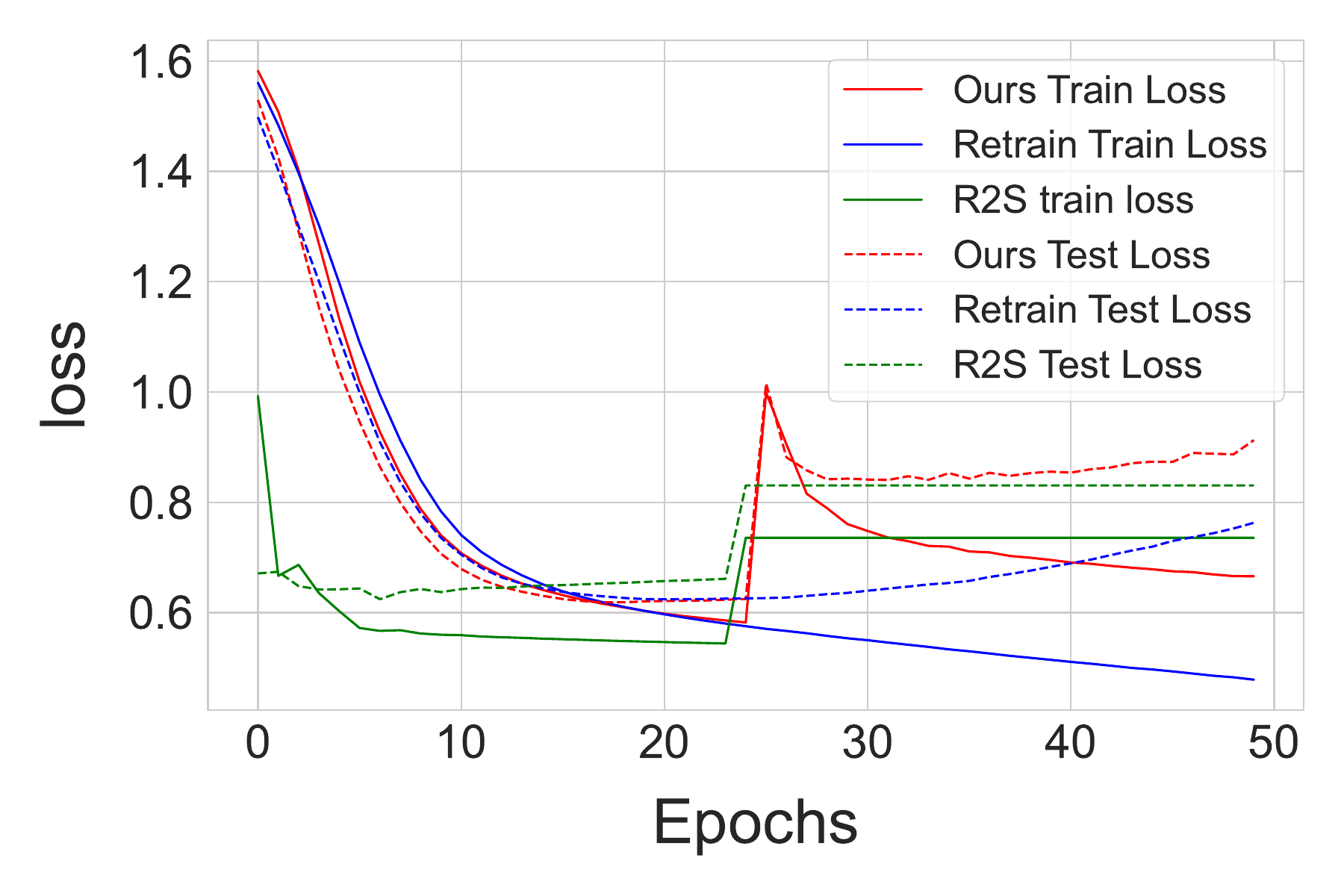}
        \label{poqemon_loss_MI}
    }
    \hfill
    \subfloat[Susy]{
        \includegraphics[width=0.48\columnwidth]{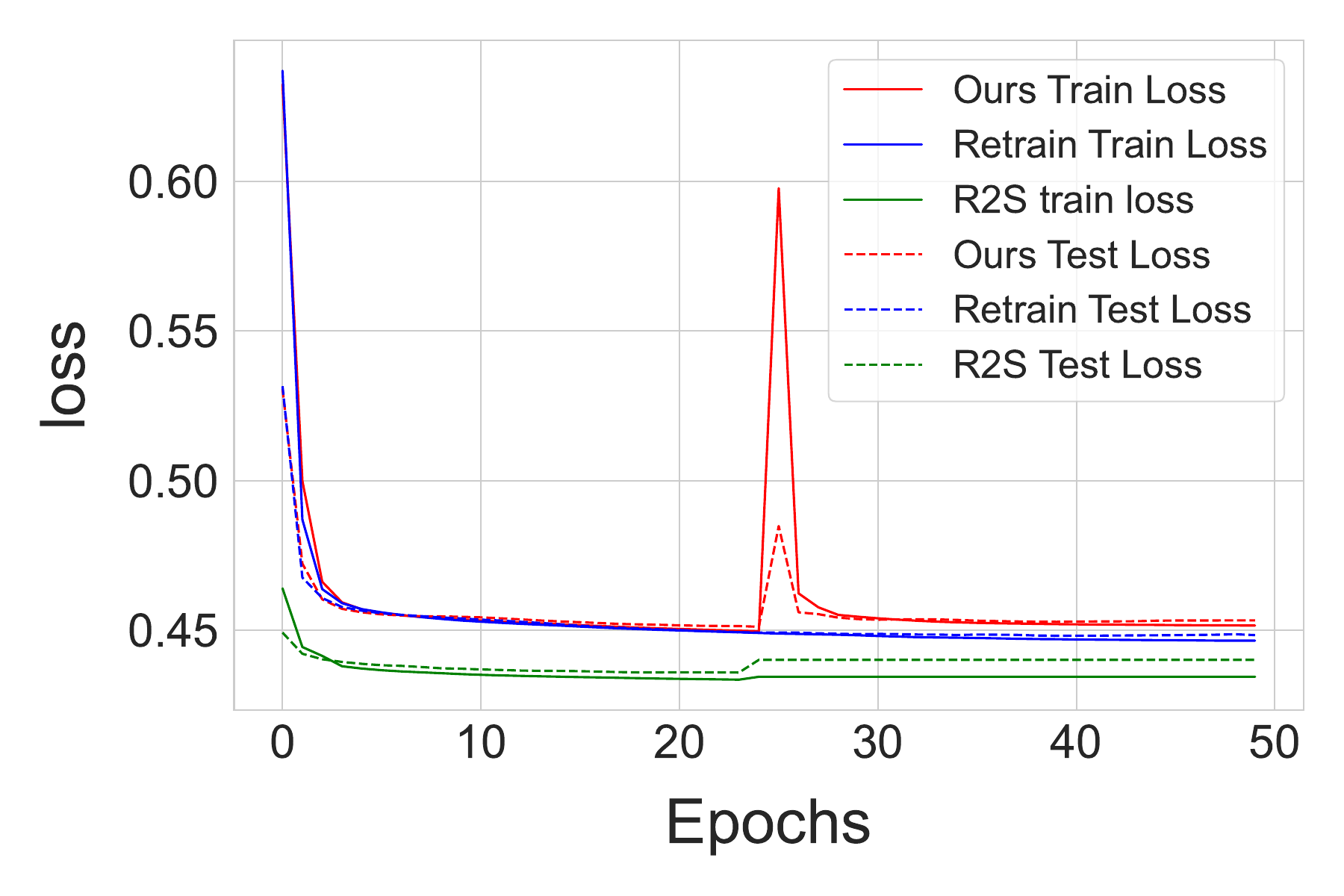}
        \label{susy_loss_MI}
    }
    \hfill
    \subfloat[Wine]{
        \includegraphics[width=0.48\columnwidth]{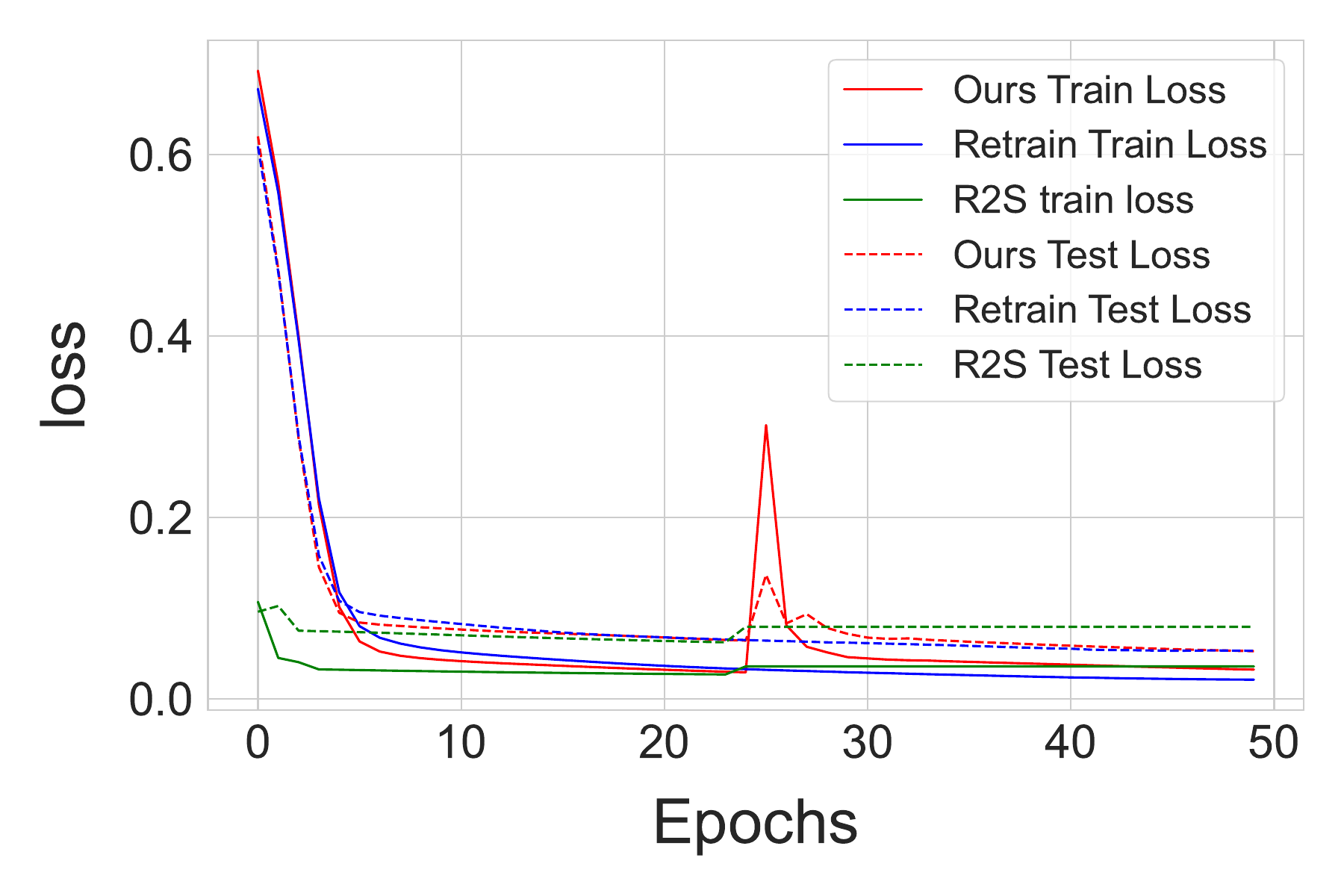}
        \label{wine_loss_MI}
    }
    \caption{The training and test loss of VFU-KD for most important feature, compared to the retrained model from scratch and R2S method.}
    \label{loss_values_mostImp_feat}
\end{figure}

\begin{table}[]
    \centering
    \begin{tabular}{c|cc|cc|cc}
        \toprule
         \multirow{2}{*}{{
         Dataset}} & \multicolumn{2}{c|}{{RfS}} & \multicolumn{2}{c|}{{R2S}} & \multicolumn{2}{c}{{Ours}}\\
         & {AUC} & {F1} & {AUC} & {F1} & {AUC} & {F1} \\
         \midrule
         {Adult} & {0.82} & {0.82} & {0.82} & {0.82} & {\textbf{0.82}} & {\textbf{0.82}} \\
         {ai4i} & {0.86} & {0.87} & {0.87} & {0.90} & {0.86} & {\textbf{0.96}} \\
         {Hepmass} & {0.84} & {0.84} & {0.85} & {0.85} & {0.84} & {0.84} \\
         {Poqemon} & {0.92} & {0.78} & {0.89} & {0.70} & {0.89} & {0.71} \\
         {Susy} & {0.79} & {0.79} & {0.79} & {0.79} & {\textbf{0.79}} & {\textbf{0.79}} \\
         {Wine} & {0.99} & {0.99} & {0.99} & {0.99} & {\textbf{0.99}} & {\textbf{0.99}} \\
        \bottomrule
    \end{tabular}
    \caption{The AUC and F1 score comparison of VFU-KD (Ours) with the retrained-from-scratch (RfS) model and the R2S method for unlearning the most important feature.}
    \label{tab:VFU-KD_MI}
\end{table}

\begin{figure}[h]
    \centering
    \subfloat[Adult]{
        \includegraphics[width=0.48\columnwidth]{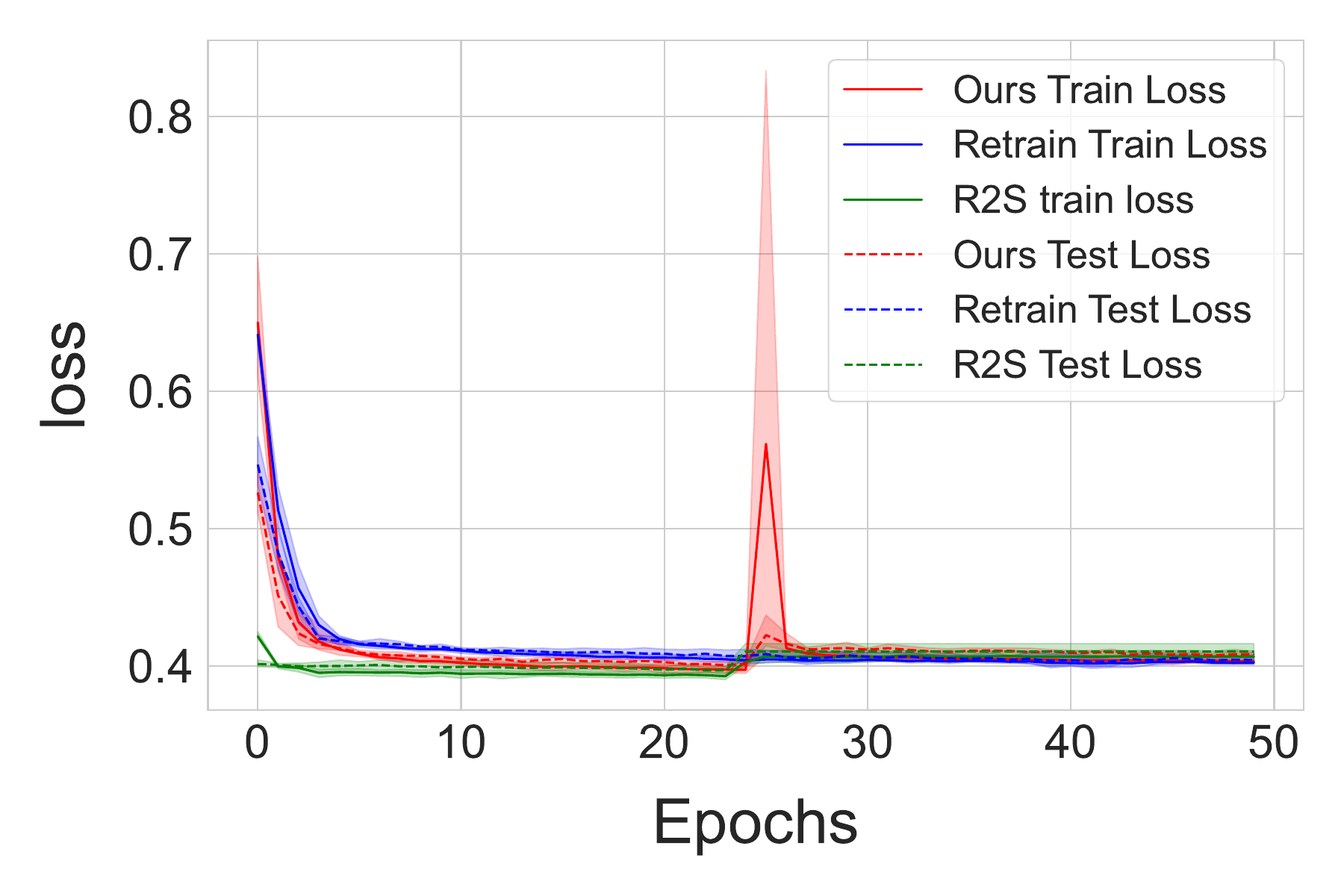}
        \label{adult_loss_LI}
    }
    \hfill
    \subfloat[ai4i]{
        \includegraphics[width=0.48\columnwidth]{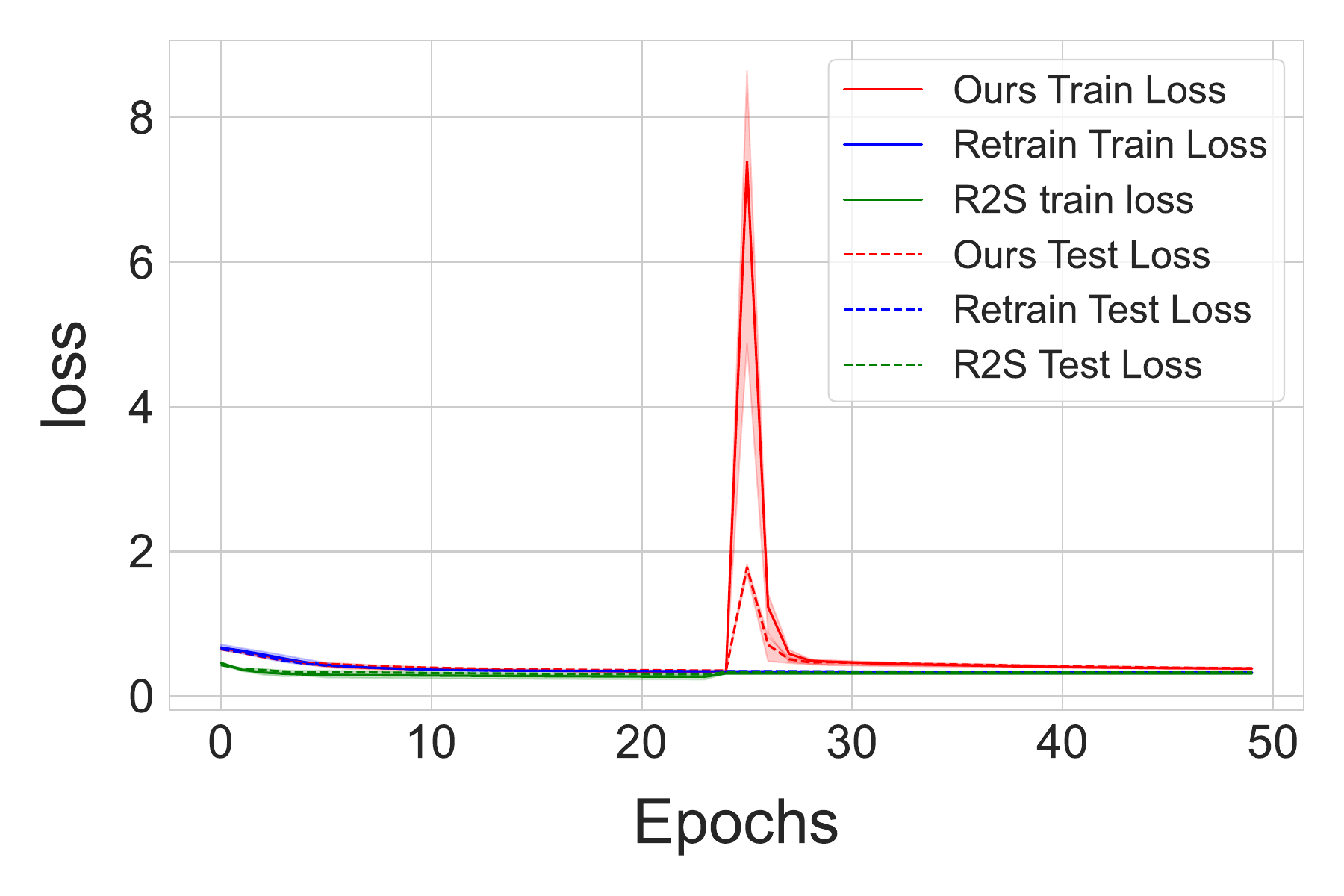}
        \label{ai4i_loss_LI}
    }
    \hfill
    \subfloat[Hepmass]{
        \includegraphics[width=0.48\columnwidth]{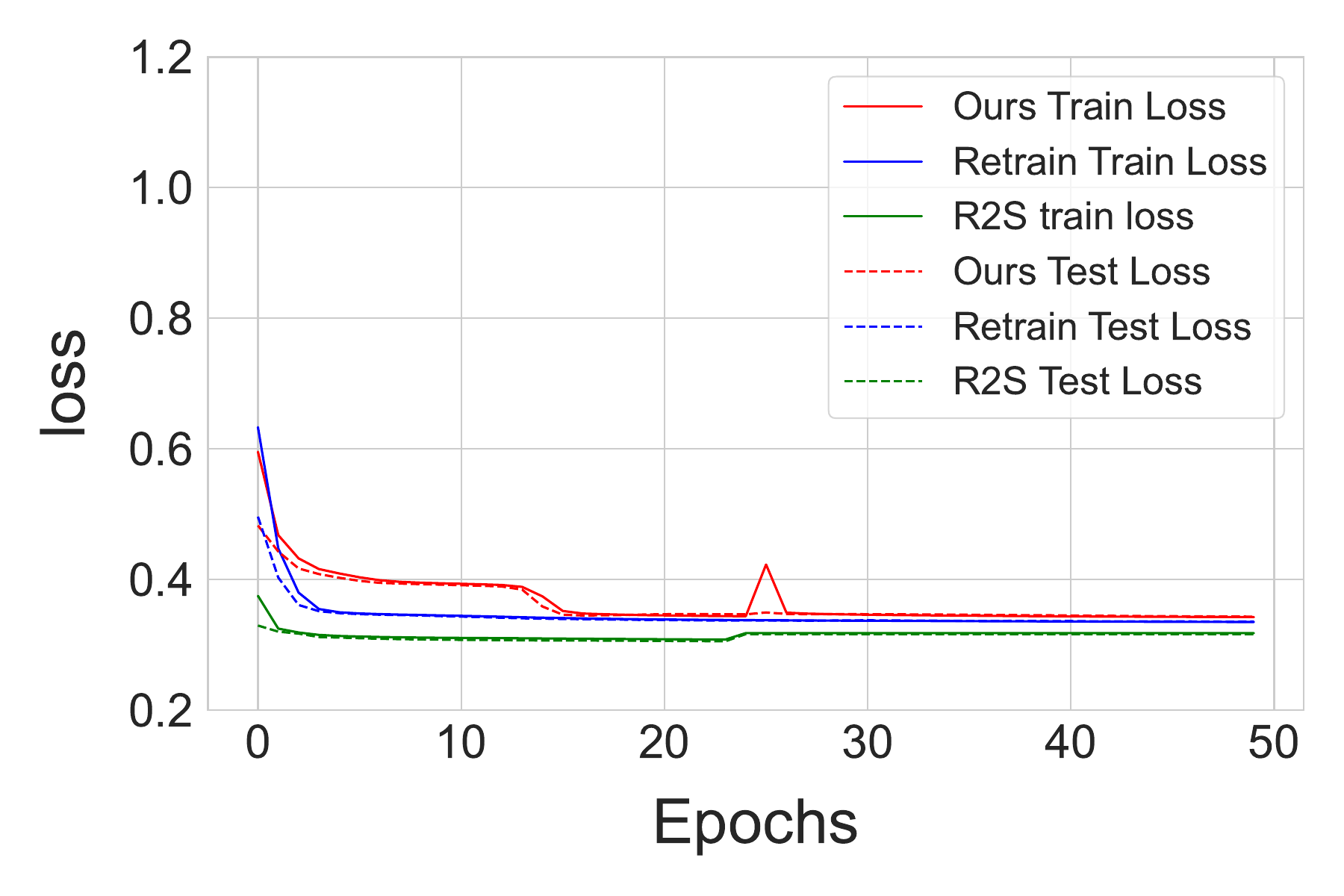}
        \label{hepmass_loss_LI}
    }
    \hfill
    \subfloat[Poqemon]{
        \includegraphics[width=0.48\columnwidth]{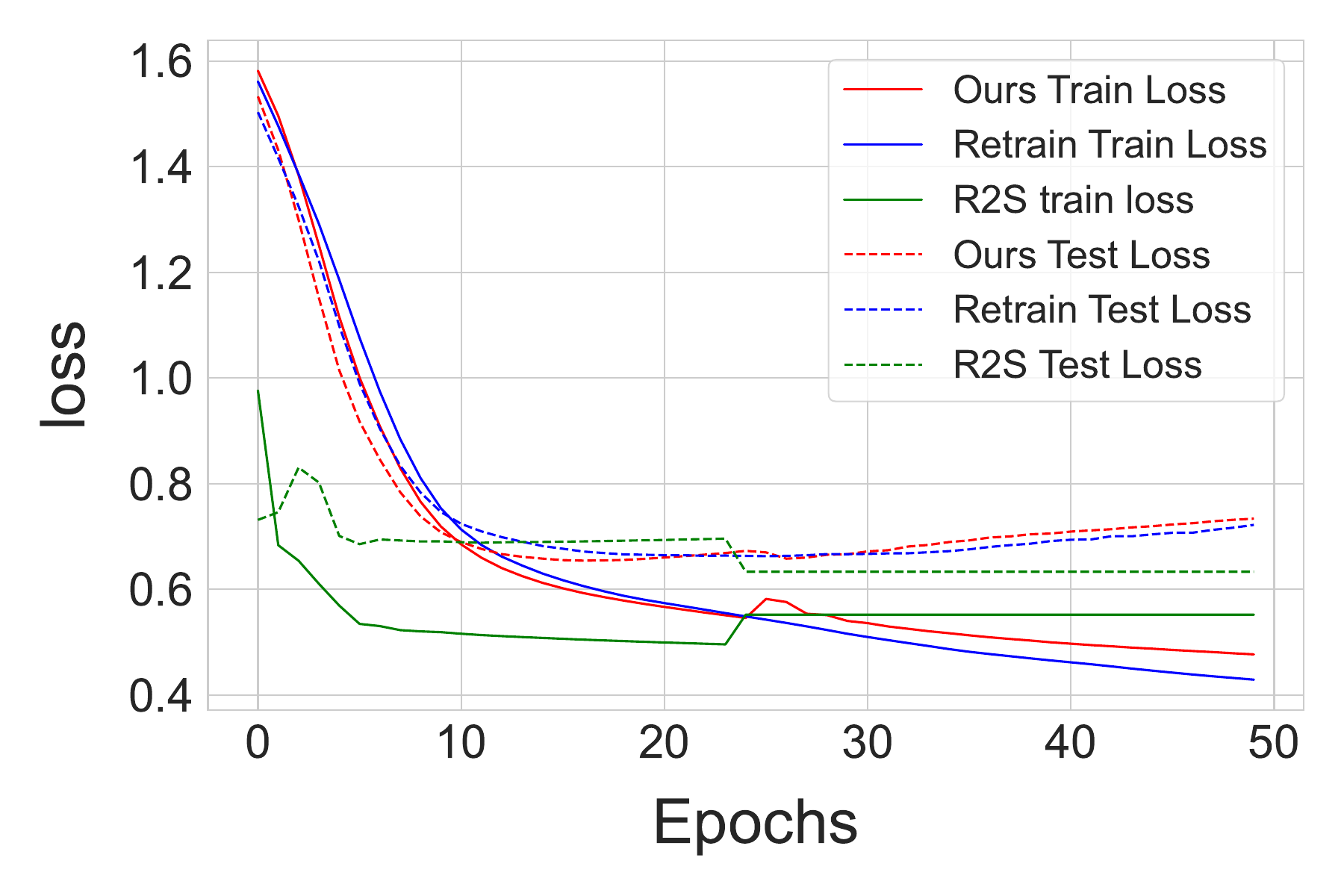}
        \label{poqemon_loss_LI}
    }
    \hfill
    \subfloat[Susy]{
        \includegraphics[width=0.48\columnwidth]{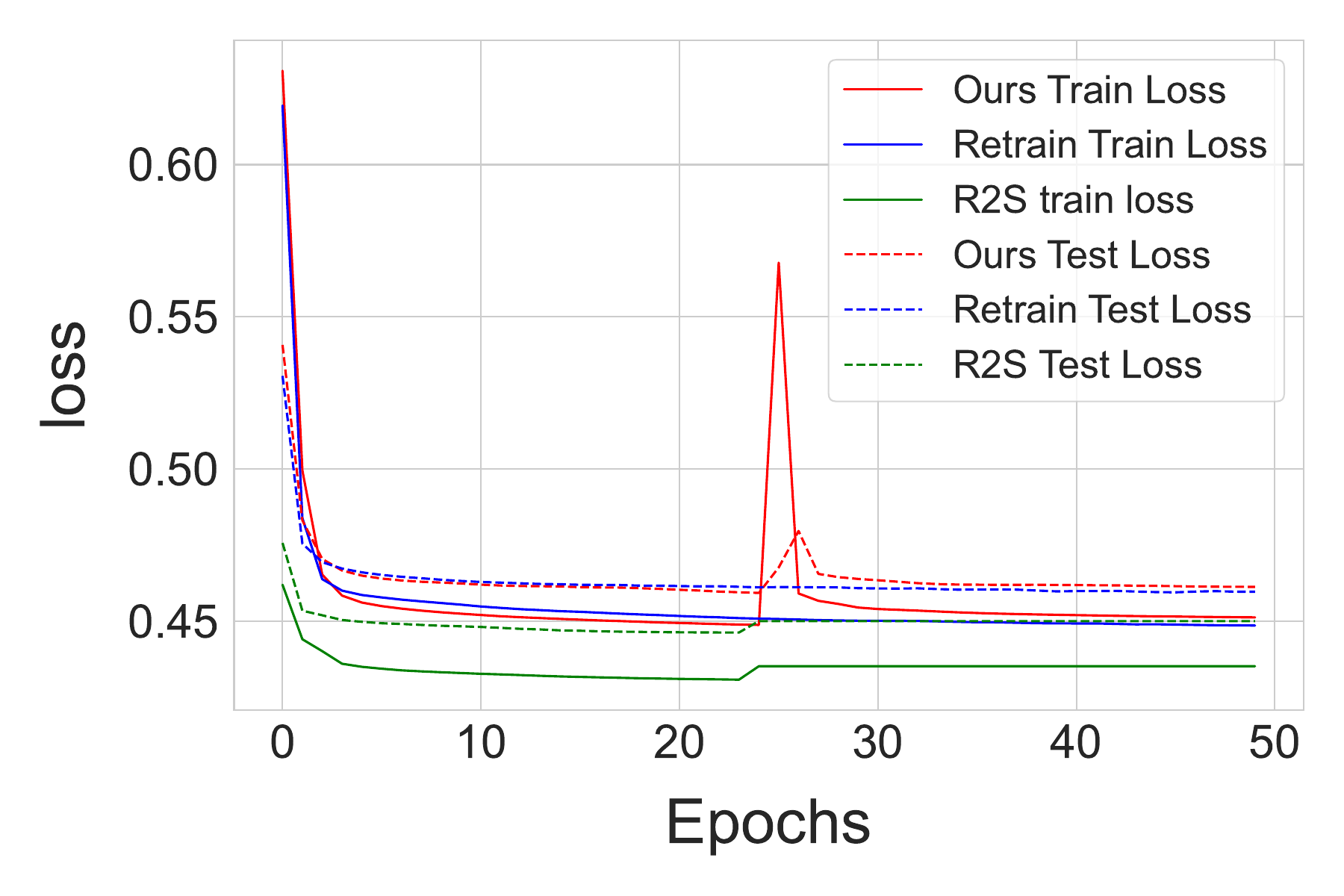}
        \label{susy_loss_LI}
    }
    \hfill
    \subfloat[Wine]{
        \includegraphics[width=0.48\columnwidth]{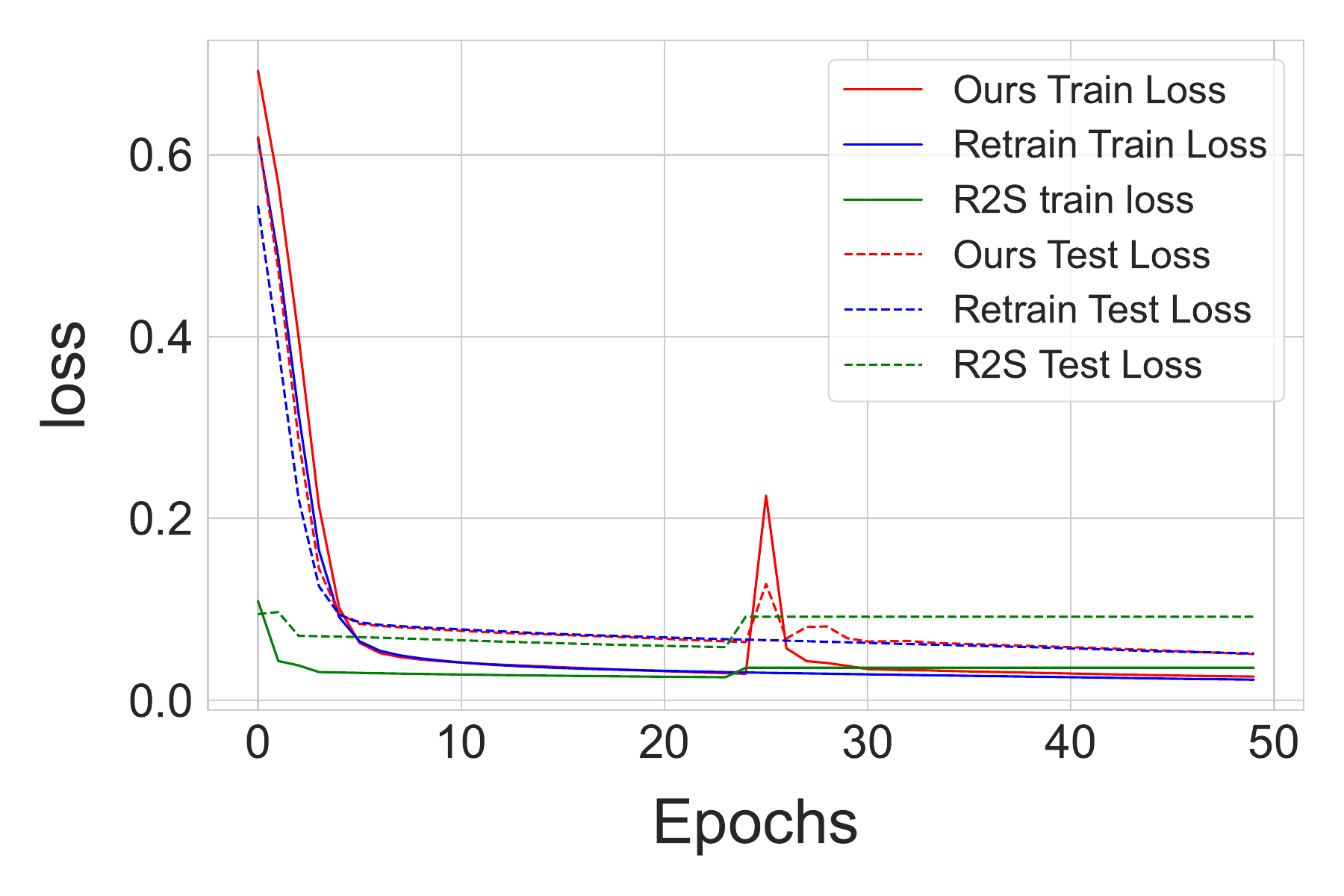}
        \label{wine_loss_LI}
    }
    \caption{The training and test loss of VFU-KD for least important feature, compared to the retrained model from scratch and R2S method.}
    \label{loss_values_leastImp_feat}
\end{figure}

\begin{table}[]
    \centering
    \begin{tabular}{c|cc|cc|cc}
        \toprule
         \multirow{2}{*}{{
         Dataset}} & \multicolumn{2}{c|}{{RfS}} & \multicolumn{2}{c|}{{R2S}} & \multicolumn{2}{c}{{Ours}}\\
         & {AUC} & {F1} & {AUC} & {F1} & {AUC} & {F1} \\
         \midrule
         {Adult} & {0.82} & {0.82} & {0.80} & {0.82} & {0.81} & {\textbf{0.82}} \\
         {ai4i} & {0.86} & {0.85} & {0.87} & {0.85} & {\textbf{0.86}} & {0.81} \\
         {Hepmass} & {0.84} & {0.84} & {0.85} & {0.85} & {0.84} & {0.84} \\
         {Poqemon} & {0.92} & {0.79} & {0.92} & {0.79} & {\textbf{0.92}} & {\textbf{0.79}} \\
         {Susy} & {0.78} & {0.79} & {0.79} & {0.79} & {\textbf{0.78}} & {\textbf{0.79}} \\
         {Wine} & {0.99} & {0.99} & {0.99} & {0.99} & {\textbf{0.99}} & {\textbf{0.99}} \\
        \bottomrule
    \end{tabular}
    \caption{The AUC and F1 score comparison of VFU-KD (Ours) with the retrained-from-scratch (RfS) model and the R2S method for unlearning the least important feature.}
    \label{tab:VFU-KD_LI}
\end{table}

\begin{figure}[!h]
    \centering
    \subfloat[Adult]{
        \includegraphics[width=0.48\columnwidth]{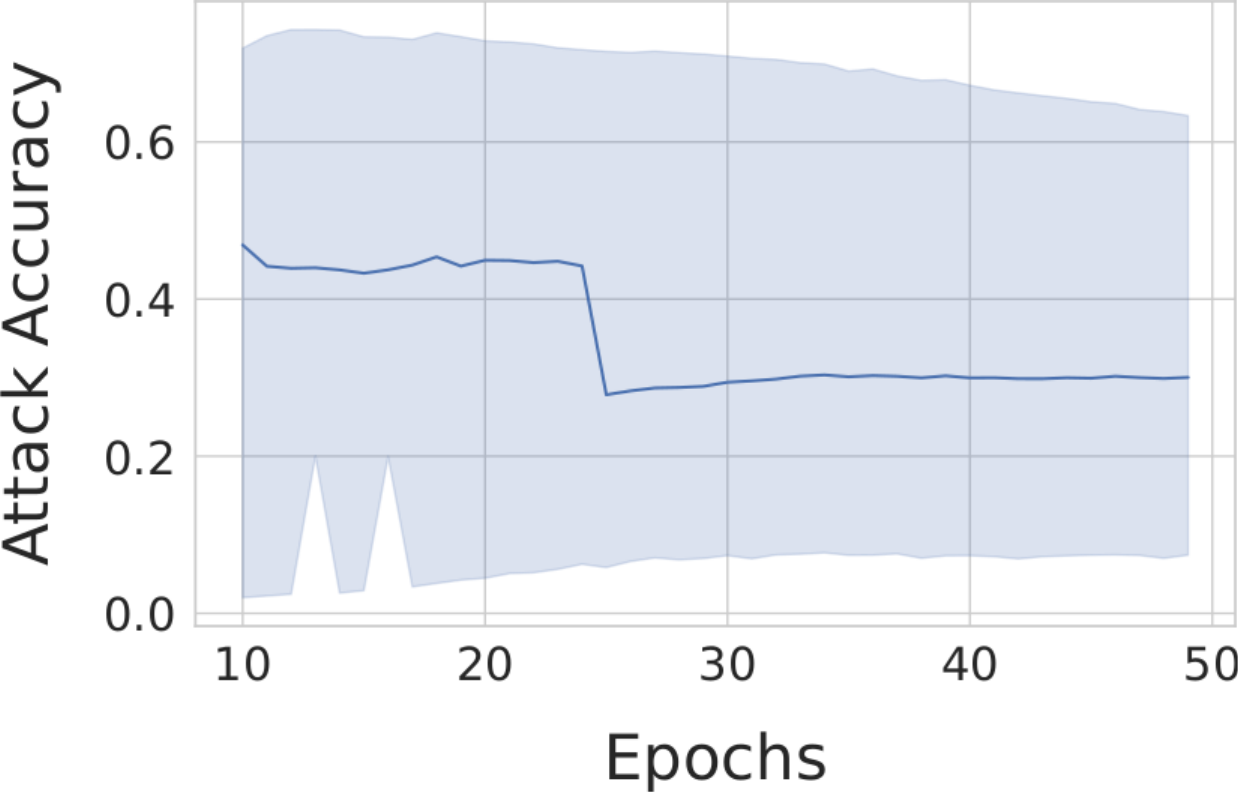}
        \label{adult_mia}
    }
    \hfill
    \subfloat[ai4i]{
        \includegraphics[width=0.48\columnwidth]{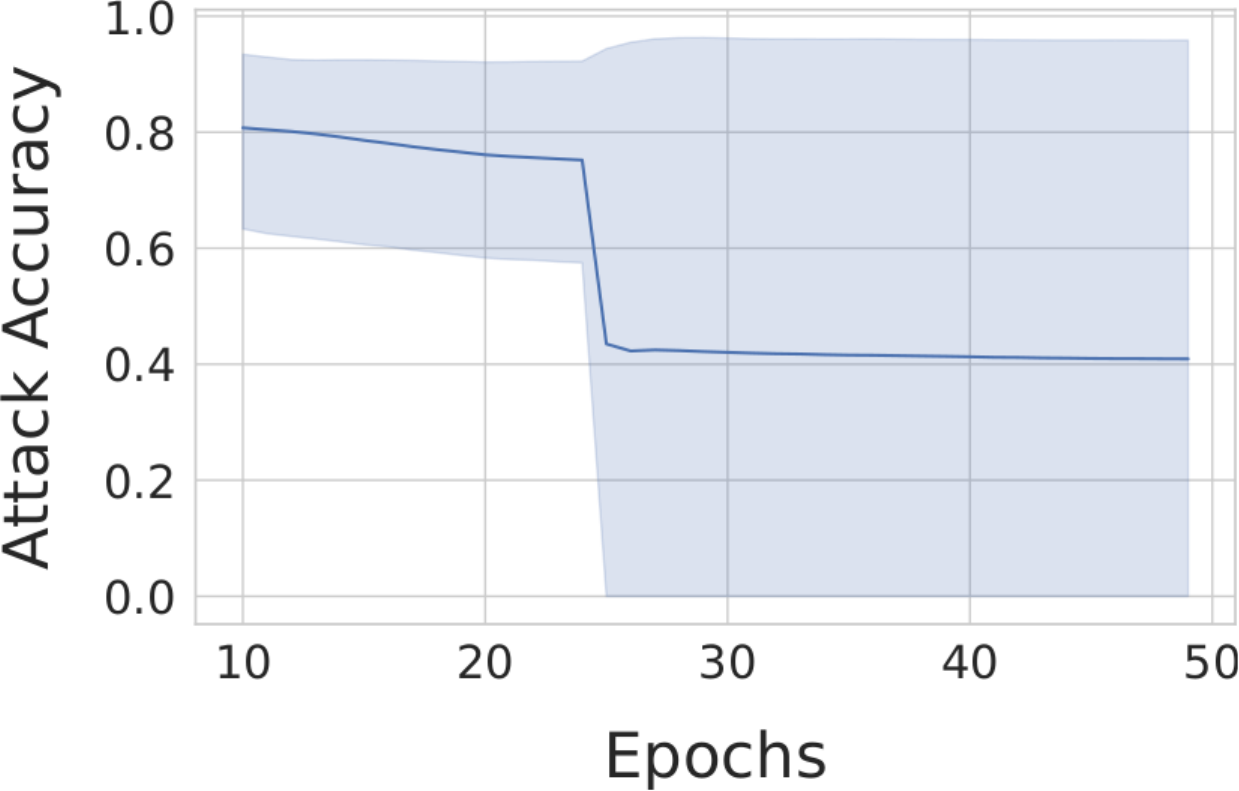}
        \label{ai4i_mia}
    }
    \hfill
    \subfloat[Hepmass]{
        \includegraphics[width=0.48\columnwidth]{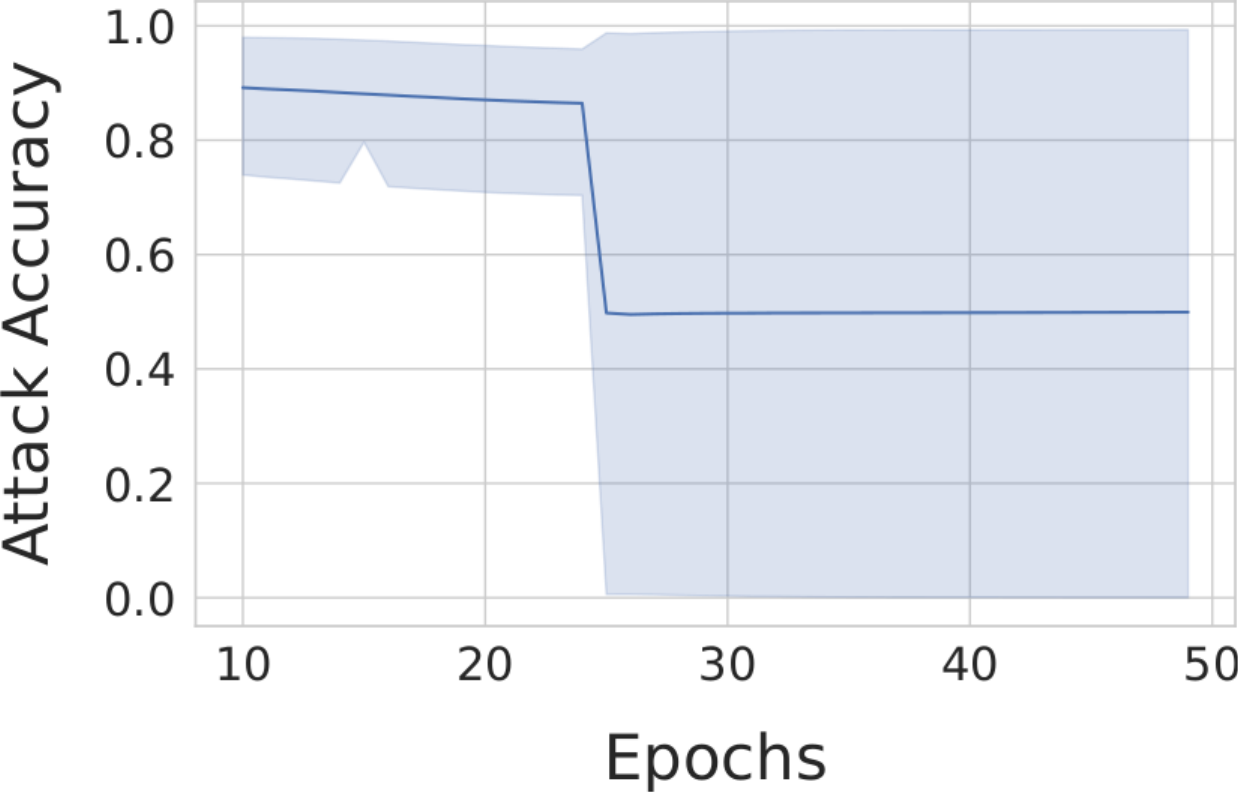}
        \label{hepmass_mia}
    }
    \hfill
    \subfloat[Poqemon]{
        \includegraphics[width=0.48\columnwidth]{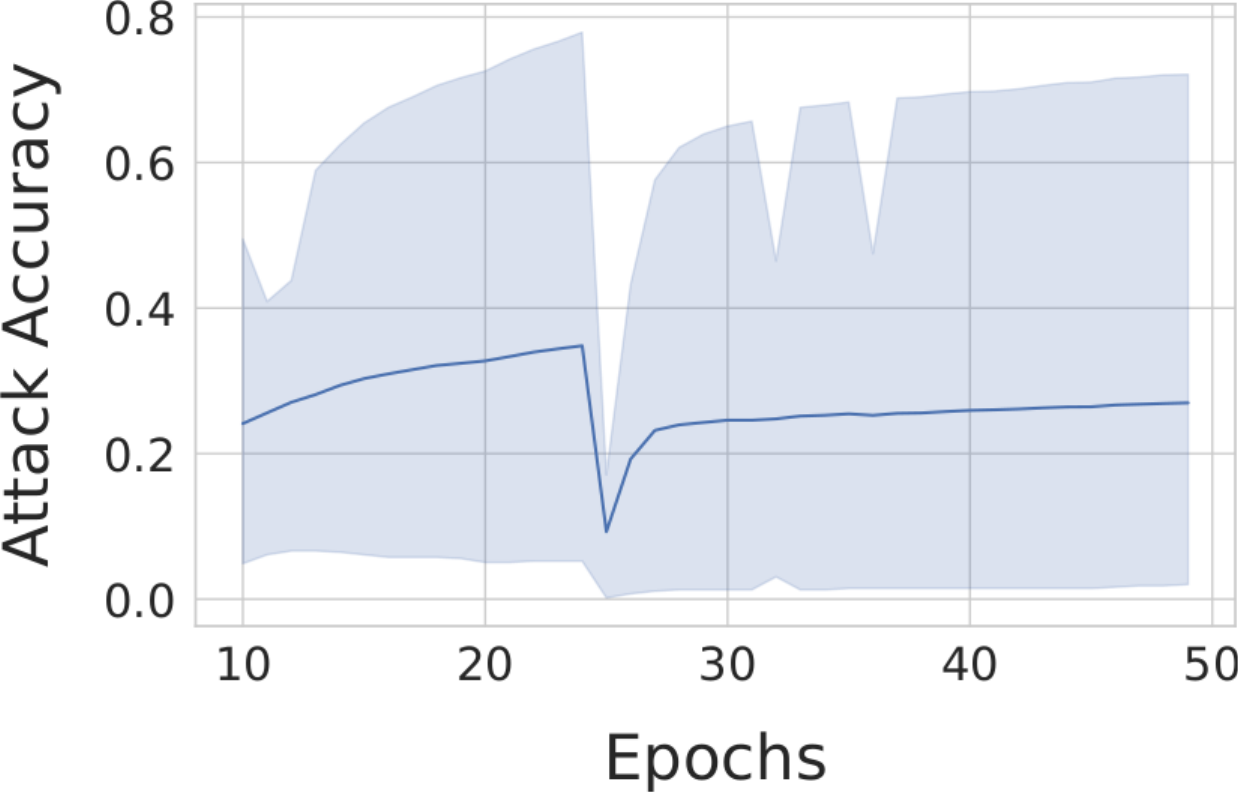}
        \label{poqemon_mia}
    }
    \hfill
    \subfloat[Susy]{
        \includegraphics[width=0.48\columnwidth]{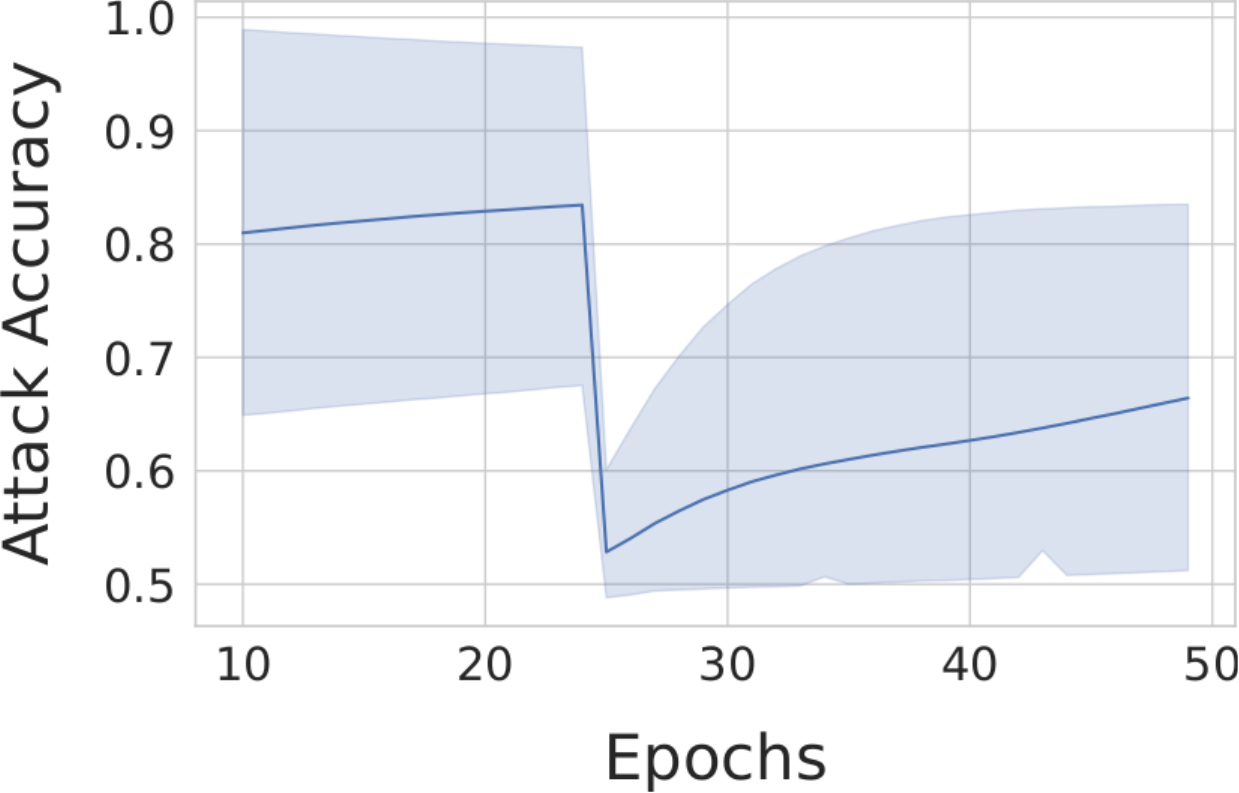}
        \label{susy_mia}
    }
    \hfill
    \subfloat[Wine]{
        \includegraphics[width=0.48\columnwidth]{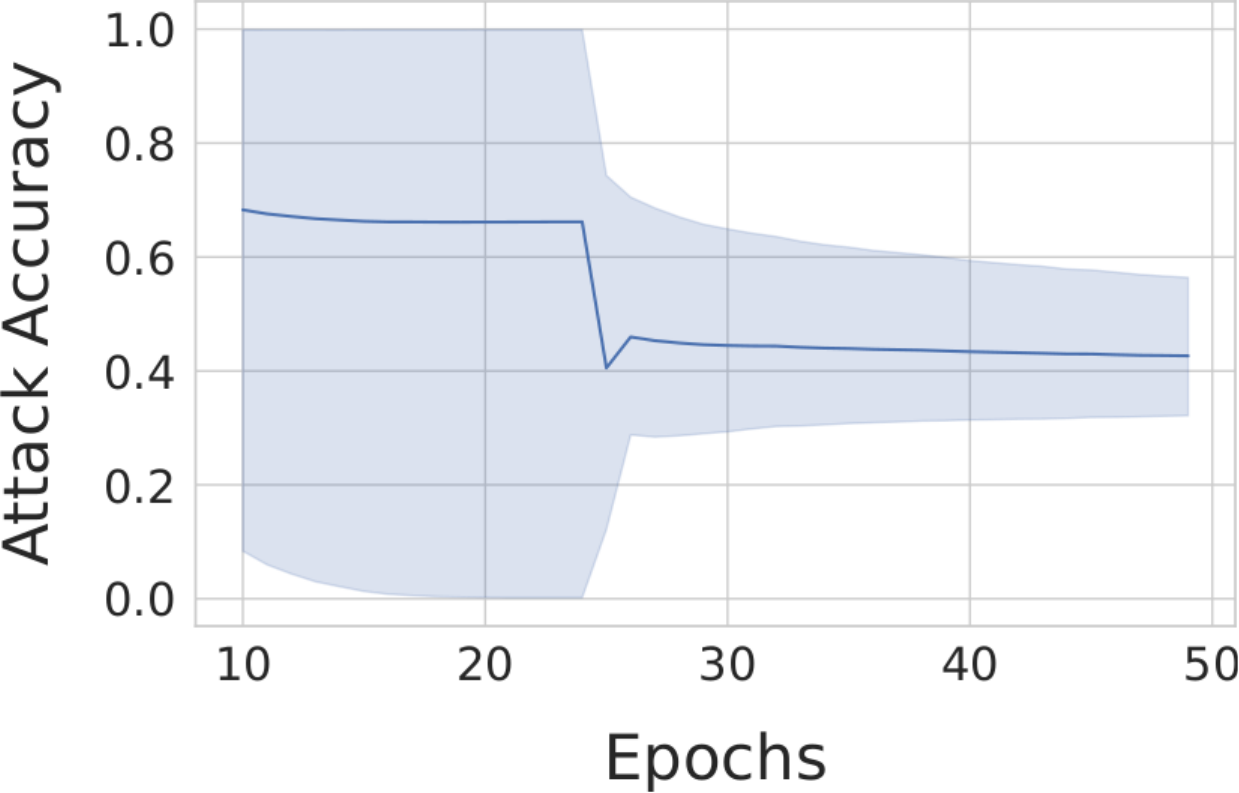}
        \label{wine_mia}
    }
    \caption{The MIA attack accuracy (y-axis) of VFU-KD.}
    \label{MIA_accuracy}
\end{figure}

Fig.~\ref{loss_values_scores_hessian} shows the training and test loss when the parameters are updated with $H^{-1}$ (refer eq. \ref{hessian vfl}). The results show that the model eventually has similar loss values to the benchmark model in most cases but has very sharp increase in the loss values after unlearning. From Fig.~\ref{loss_values_scores} and ~\ref{loss_values_scores_hessian}, we find that the results are better for unlearning with learning rates than with $\mathcal{H}^{-1}$. Further experiments with $\mathcal{H}^{-1}$ are available in supplementary material.

For comparative analysis on CIFAR10 and STL10 datasets, we argue that passive parties typically do not have incomplete images as shown in ~\cite{wang2024efficient}. We distribute the channels of image to each client i.e., $clientA$ has red channel of the image, $clientB$ has green and $clientC$ has blue channel of the image. Fig.~\ref{images_results} depicts the loss curves when unlearning occurs at the $25^{th}$ epoch. Small spikes in the loss curves, attributed to the distillation process, are observed but recover within a few epochs across all cases. Table \ref{tab:VFU-img_res} provides a comparison of AUC and F1 scores, demonstrating that our method achieves results equivalent to those obtained through retraining from scratch. From the results of Fig.~\ref{loss_values_scores}, and~\ref{images_results}, we can say that our approach, VFU-KD, has benchmark comparable losses and utility scores across 6 tabular and 2 image datasets.

\subsection{Feature unlearning}

Now that we have shown the feasibility and effectiveness of our approach specially at the later epochs. We fix the unlearning epoch at $25{th}$ for feature and sample unlearning. For feature unlearning, we have removed the most important and least important features from tabular datasets. The importance of the feature is computed with feature ablation\footnote{\href{https://captum.ai/api/feature_ablation.html}{https://captum.ai/api/feature\_ablation.html}} (feature importance for all the tabular dataset is available in supplementary material). Fig.~\ref{loss_values_mostImp_feat} shows the training and test loss for unlearning the most important feature and Fig.~\ref{loss_values_leastImp_feat} shows the training and test loss for unlearning the least important feature for all the datasets. Table \ref{tab:VFU-KD_MI} and Table \ref{tab:VFU-KD_LI} show the results for AUC score and F1 score. Here as well, in all the datasets, the loss values and utility scores of VFU-KD are benchmark comparable, even better in some cases. Hence, we can say that our approach, VFU-KD, can effectively unlearn passive party as well as feature(s) from passive party. Additional feature unlearning experiments such as unlearning multiple features and most-least important features from each client, are available in the supplementary material.

\begin{table}[]
    \centering
    \begin{tabular}{c|cc|cc|cc}
        \toprule
         \multirow{2}{*}{{
         Dataset}} & \multicolumn{2}{c|}{{RfS}} & \multicolumn{2}{c|}{{R2S}} & \multicolumn{2}{c}{{Ours}}\\
         & {AUC} & {F1} & {AUC} & {F1} & {AUC} & {F1} \\
         \midrule
         {Adult} & {0.82} & {0.81} & {0.81} & {0.81} & {\textbf{0.82}} & {\textbf{0.82}} \\
         {ai4i} & {0.91} & {0.93} & {0.91} & {0.93} & {\textbf{0.91}} & {\textbf{0.93}} \\
         {Hepmass} & {0.86} & {0.86} & {0.85} & {0.85} & {\textbf{0.86}} & {\textbf{0.86}} \\
         {Poqemon} & {0.92} & {0.74} & {0.92} & {0.72} & {\textbf{0.93}} & {\textbf{0.77}} \\
         {Susy} & {0.80} & {0.80} & {0.80} & {0.80} & {\textbf{0.80}} & {\textbf{0.80}} \\
         {Wine} & {0.99} & {0.99} & {0.99} & {0.99} & {\textbf{0.99}} & {\textbf{0.99}} \\
        \bottomrule
    \end{tabular}
    \caption{The AUC and F1 score comparison of VFU-GA (Ours) with the retrained-from-scratch (RfS) model and the R2S method for unlearning 5 batches.}
    \label{tab:VFU-GA_5batches}
\end{table}

\subsection{Sample unlearning}

\begin{figure}[!h]
    \centering
        \subfloat[Adult]{
        \includegraphics[width=0.48\columnwidth]{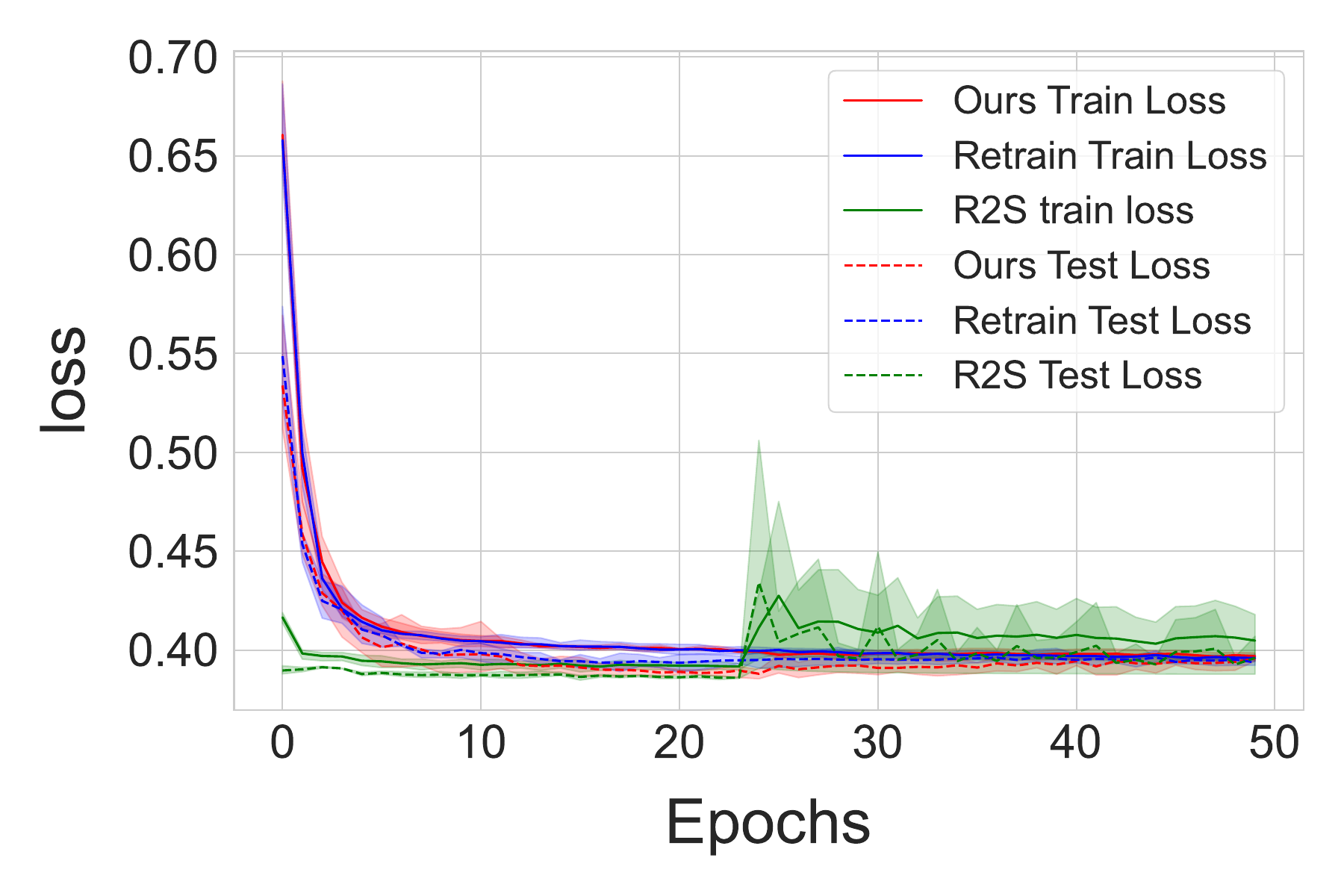}
        \label{adult_GA_loss}
    }
    \hfill
    \subfloat[ai4i]{
        \includegraphics[width=0.48\columnwidth]{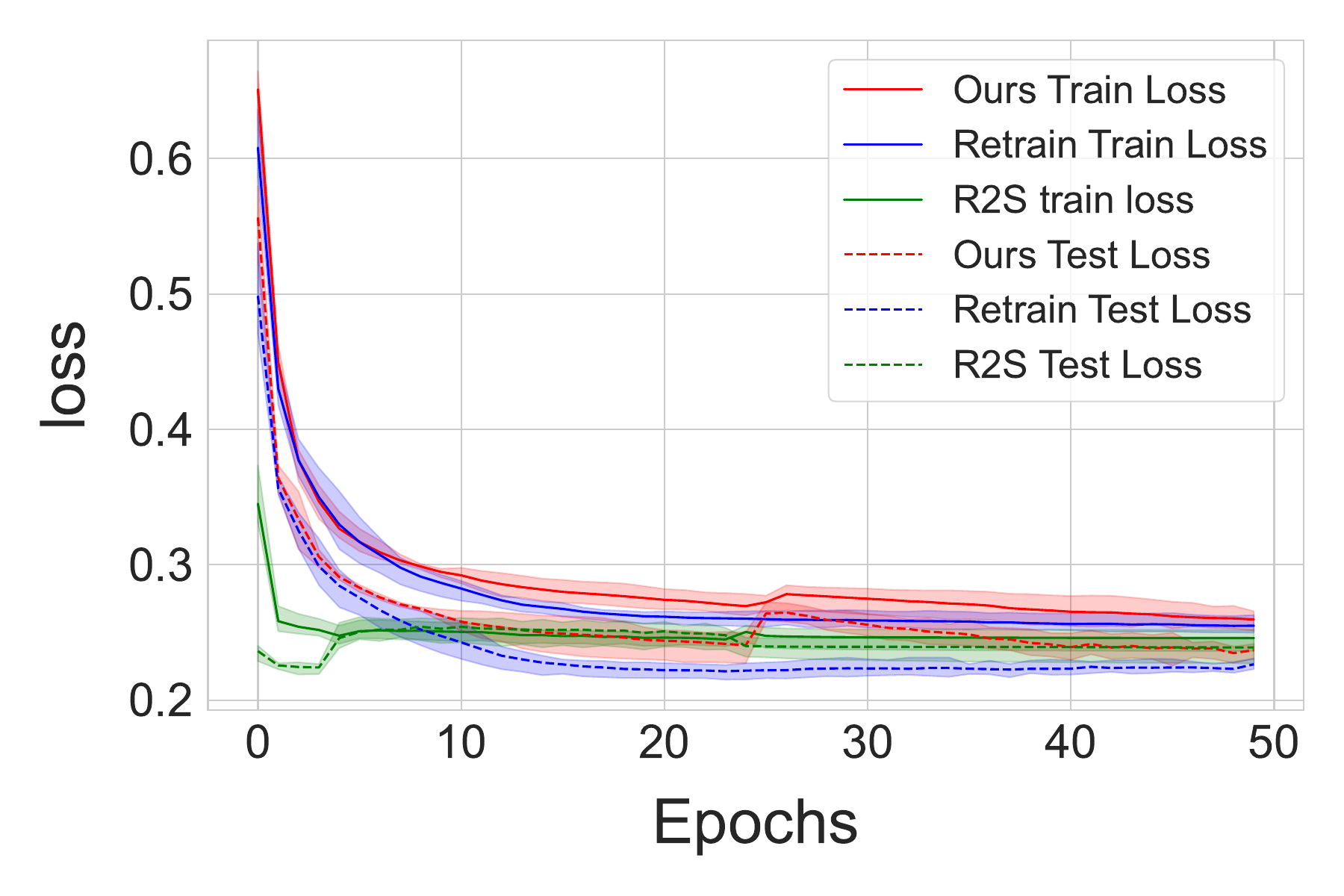}
        \label{ai4i_GA_loss}
    }
    \hfill
    \subfloat[Hepmass]{
        \includegraphics[width=0.48\columnwidth]{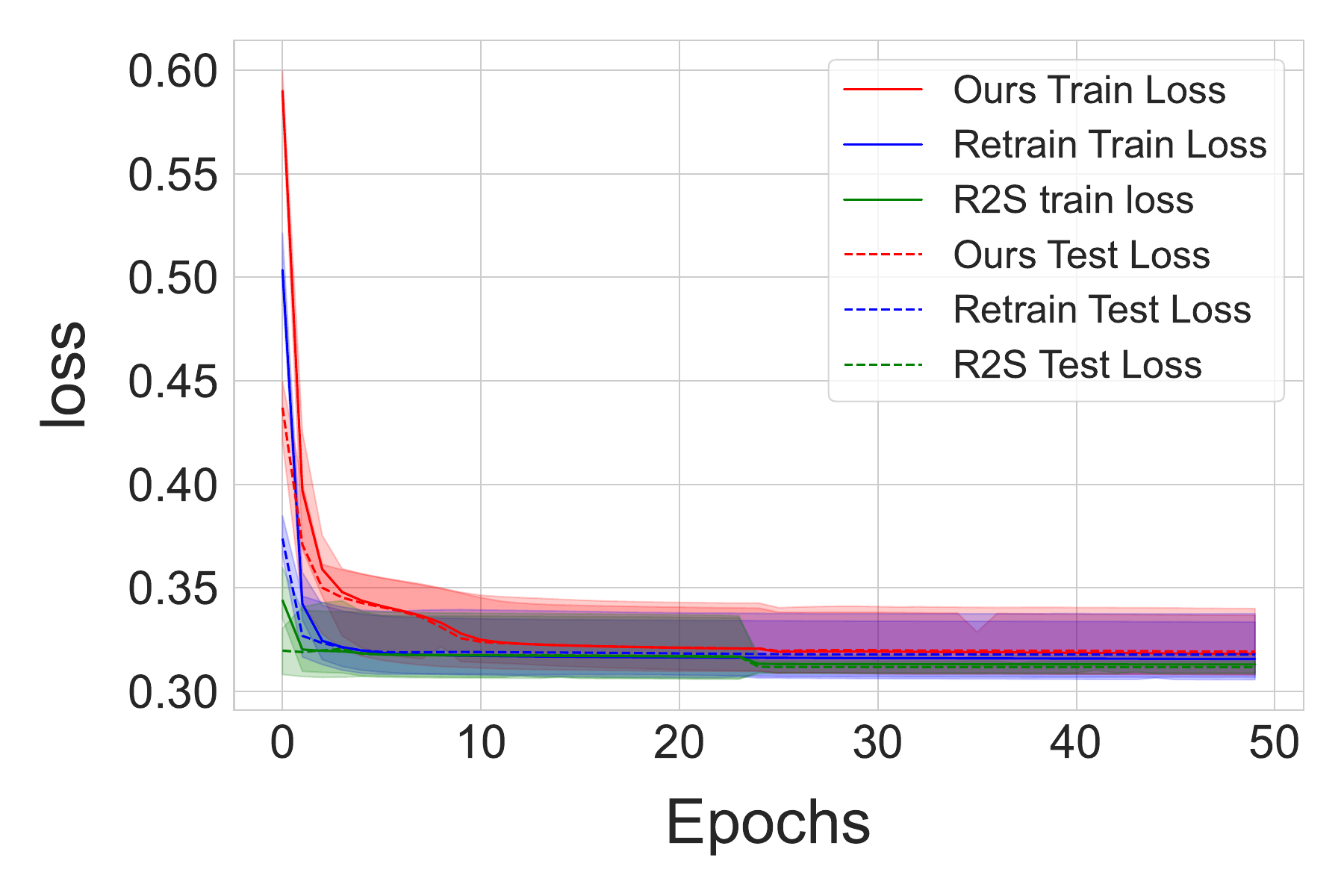}
        \label{hepmass_GA_loss}
    }
    \hfill
    \subfloat[Poqemon]{
        \includegraphics[width=0.48\columnwidth]{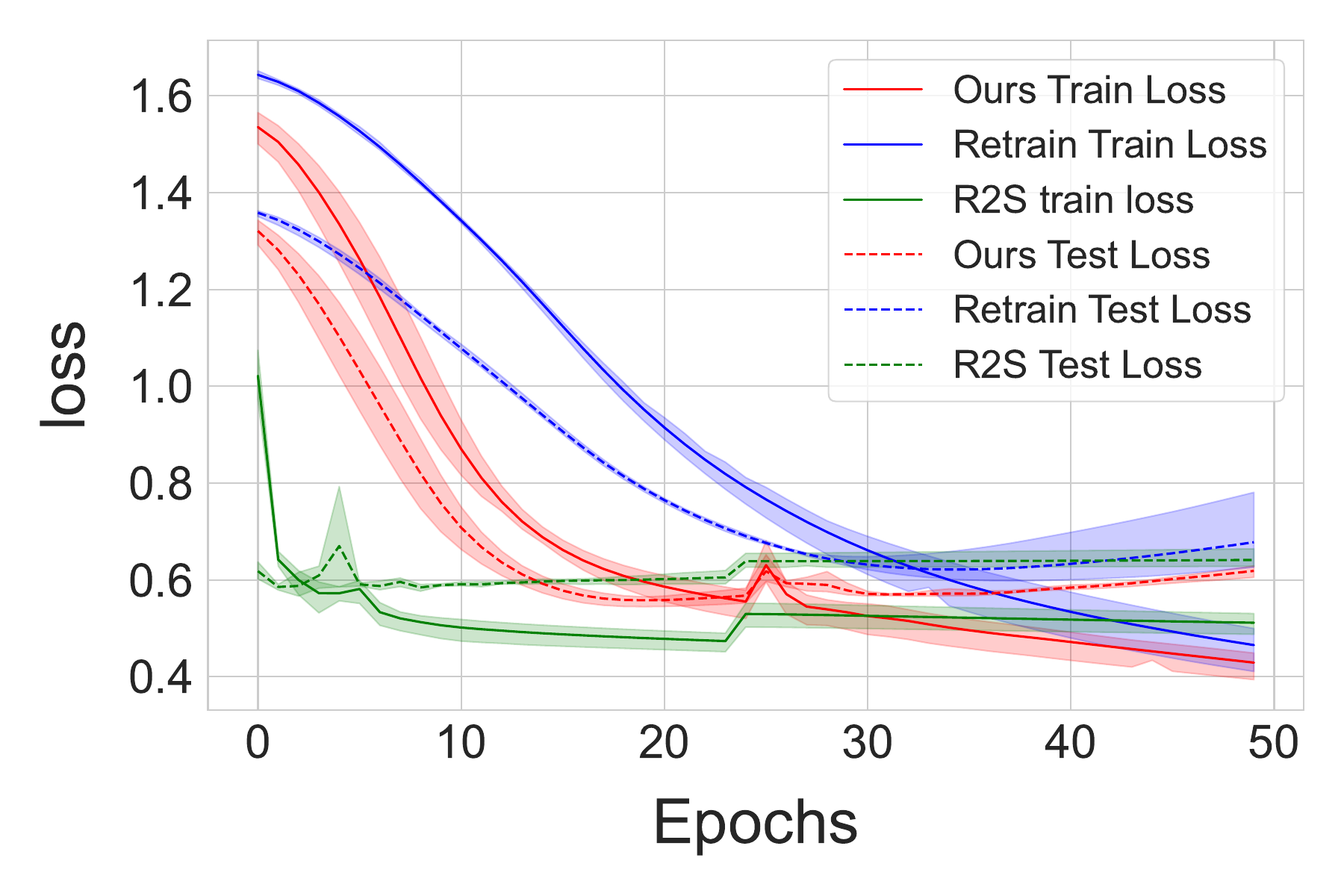}
        \label{poqemon_GA_loss}
    }
    \hfill
    \subfloat[Susy]{
        \includegraphics[width=0.48\columnwidth]{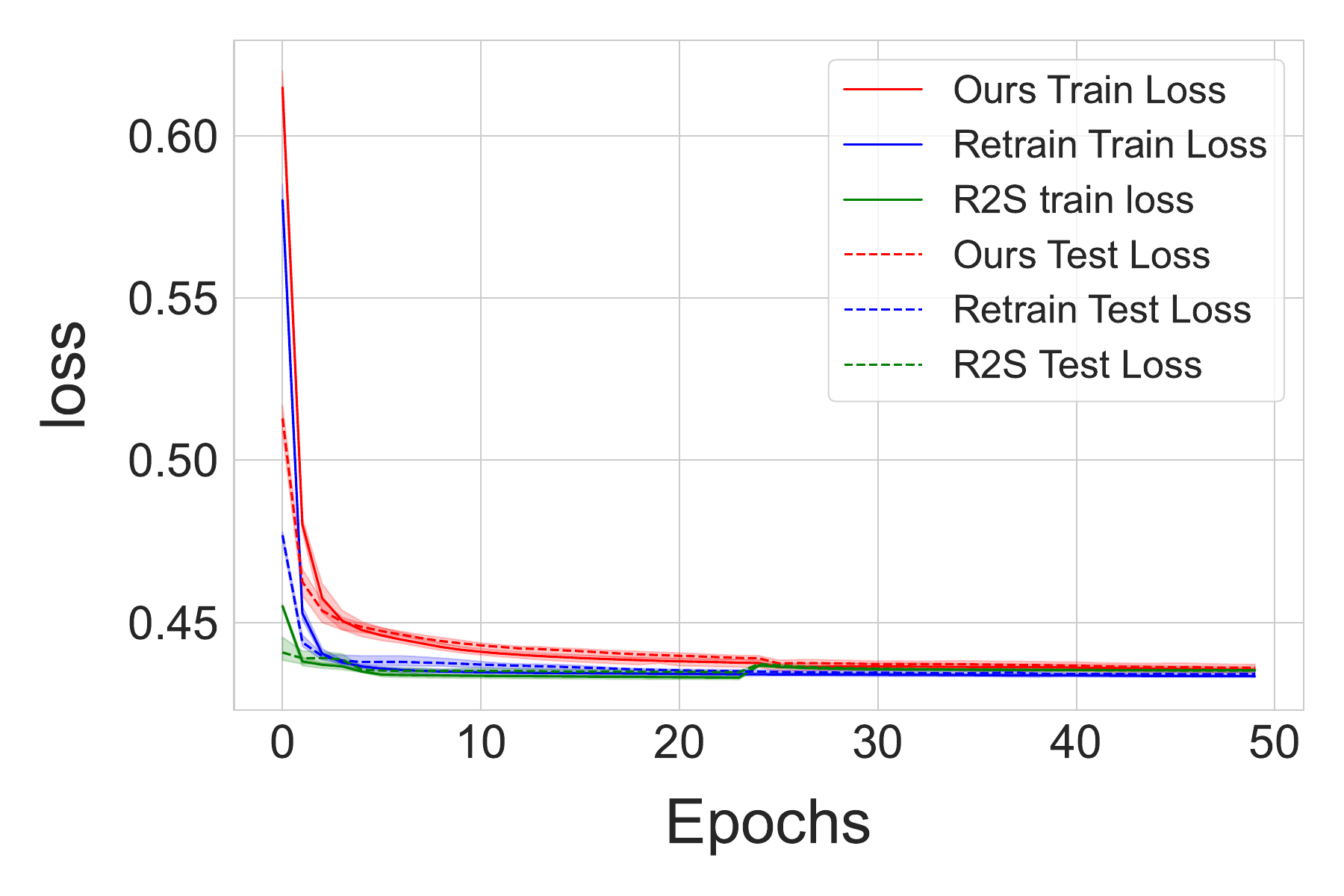}
        \label{susy_GA_loss}
    }
    \hfill
    \subfloat[Wine]{
        \includegraphics[width=0.48\columnwidth]{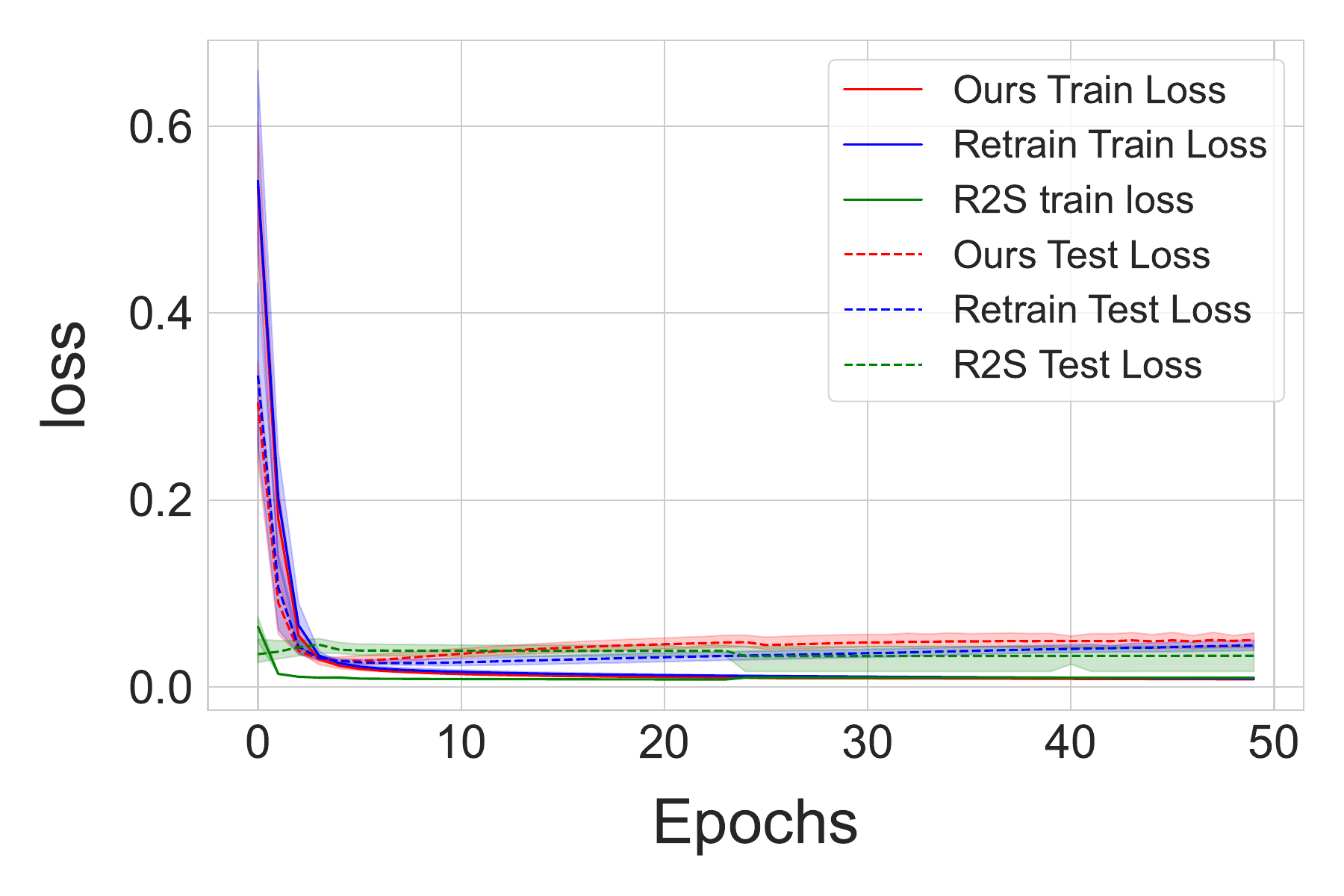}
        \label{wine_GA_loss}
    }
    \caption{The training and test loss of VFU-GA (5 batches) for least important feature, compared to the retrained model from scratch and R2S method.}
    \label{Sample_unlearning_loss_5}
\end{figure}

For sample unlearning, we removed 5 batches from the active model for each dataset. The number of gradient ascent steps was set to 5, chosen arbitrarily. This parameter can be adjusted and increased based on the unlearning requirements of the application. Fig.~\ref{Sample_unlearning_loss_5} shows   the comparison of training and test loss curves between our approach, retrained from scratch method and R2S method and Table \ref{tab:VFU-GA_5batches} shows the utility scores (F1 and AUC score) when unlearning happened at 25th epoch. The results show that VFU-GA has better utility score than retrained model. Notably, in the case of the Poqemon dataset, VFU-GA demonstrates significantly superior utility. This improvement can be attributed to the robustness introduced by gradient ascent, as models often become more robust following gradient ascent. This characteristic of gradient ascent has been leveraged in previous literature to enhance model robustness, as seen in studies such as ~\cite{sebbouh2022randomized} and ~\cite{yoon2023gradient}. Additional experimental results for unlearning a  batch are available in the supplementary material.

\begin{figure}[!h]
    \centering
    \subfloat[Adult]{
        \includegraphics[width=0.48\columnwidth]{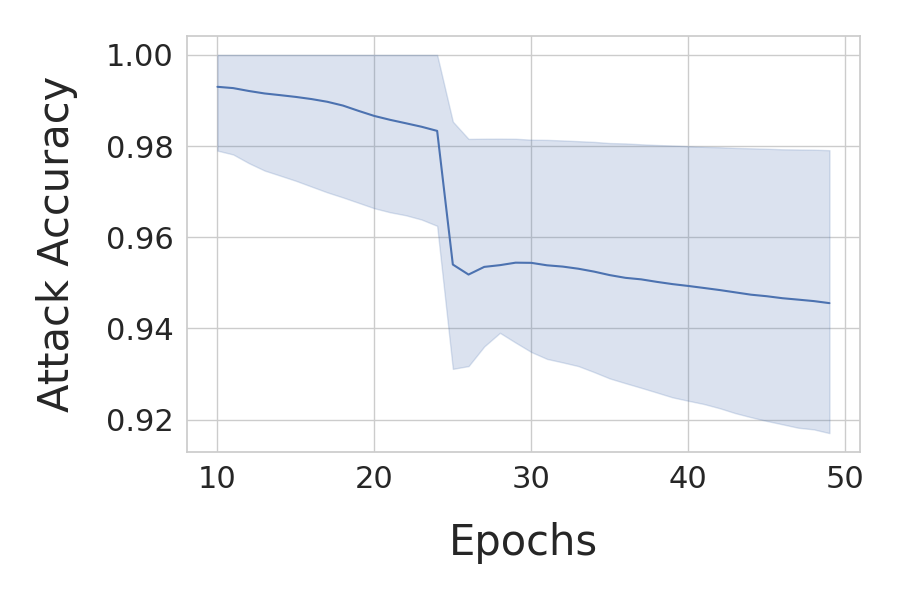}
        \label{adult_GA_MIA_5}
    }
    \hfill
    \subfloat[ai4i]{
        \includegraphics[width=0.48\columnwidth]{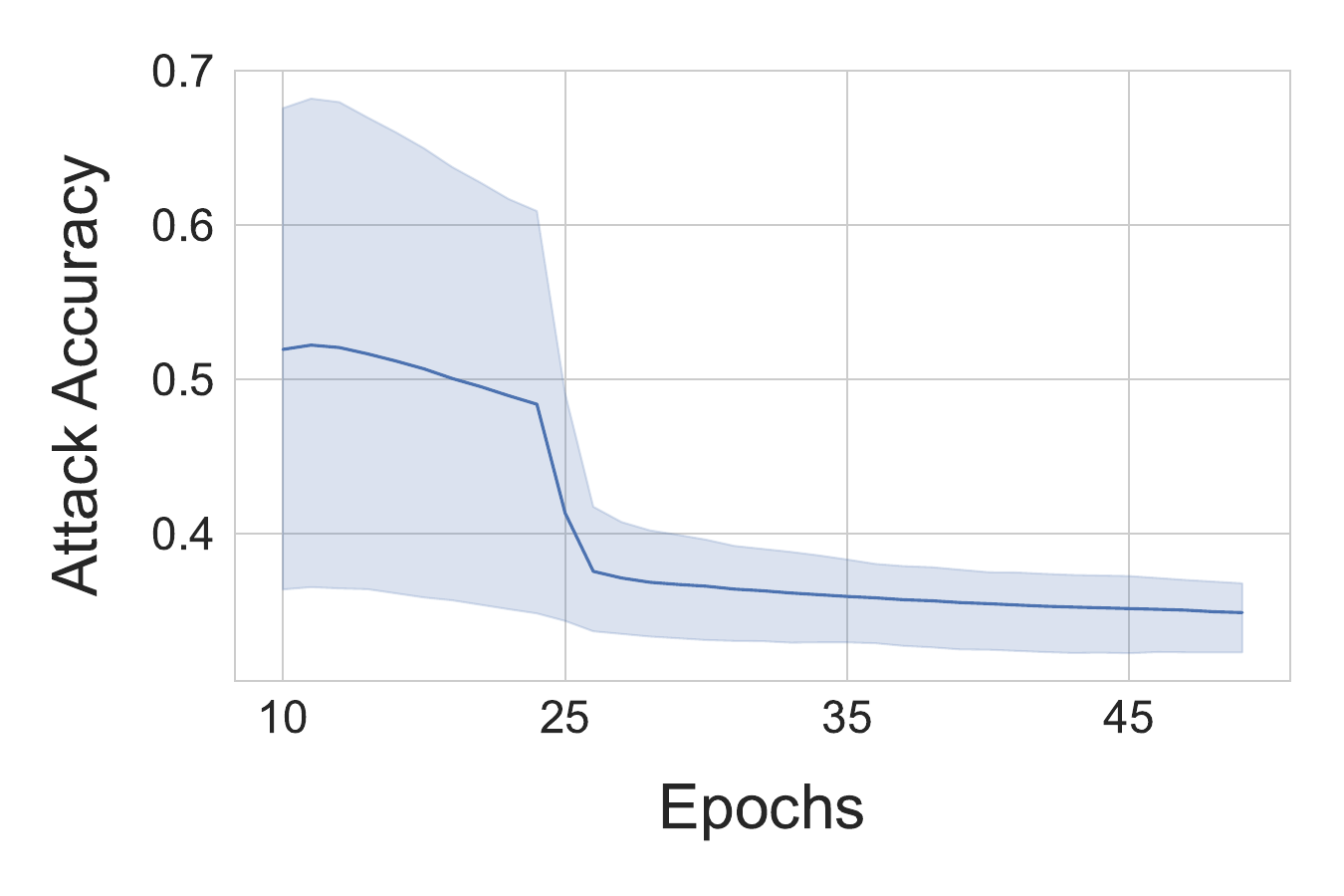}
        \label{ai4i_GA_MIA_5}
    }
    \hfill
    \subfloat[Hepmass]{
        \includegraphics[width=0.48\columnwidth]{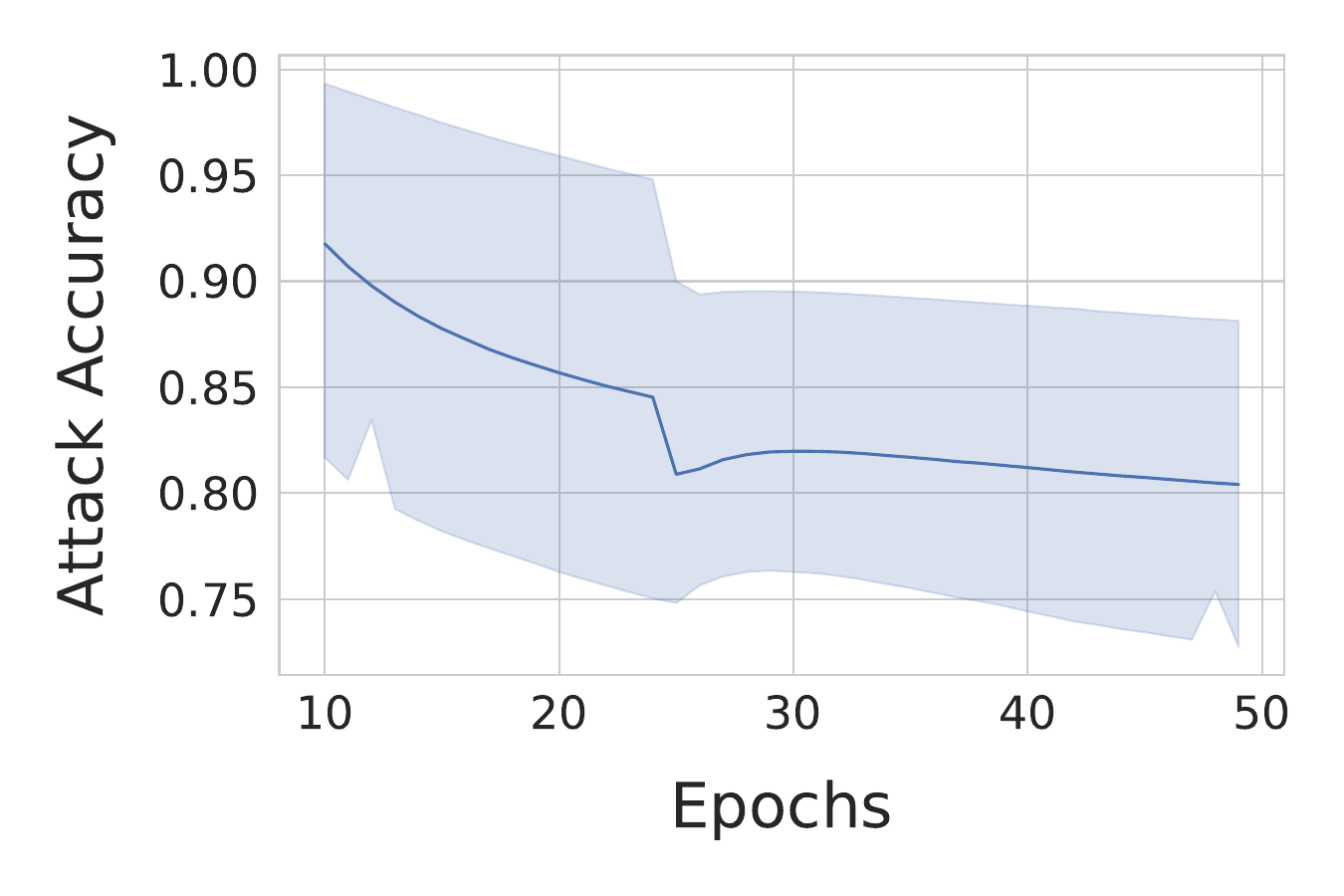}
        \label{hepmass_GA_MIA_5}
    }
    \hfill
    \subfloat[Poqemon]{
        \includegraphics[width=0.48\columnwidth]{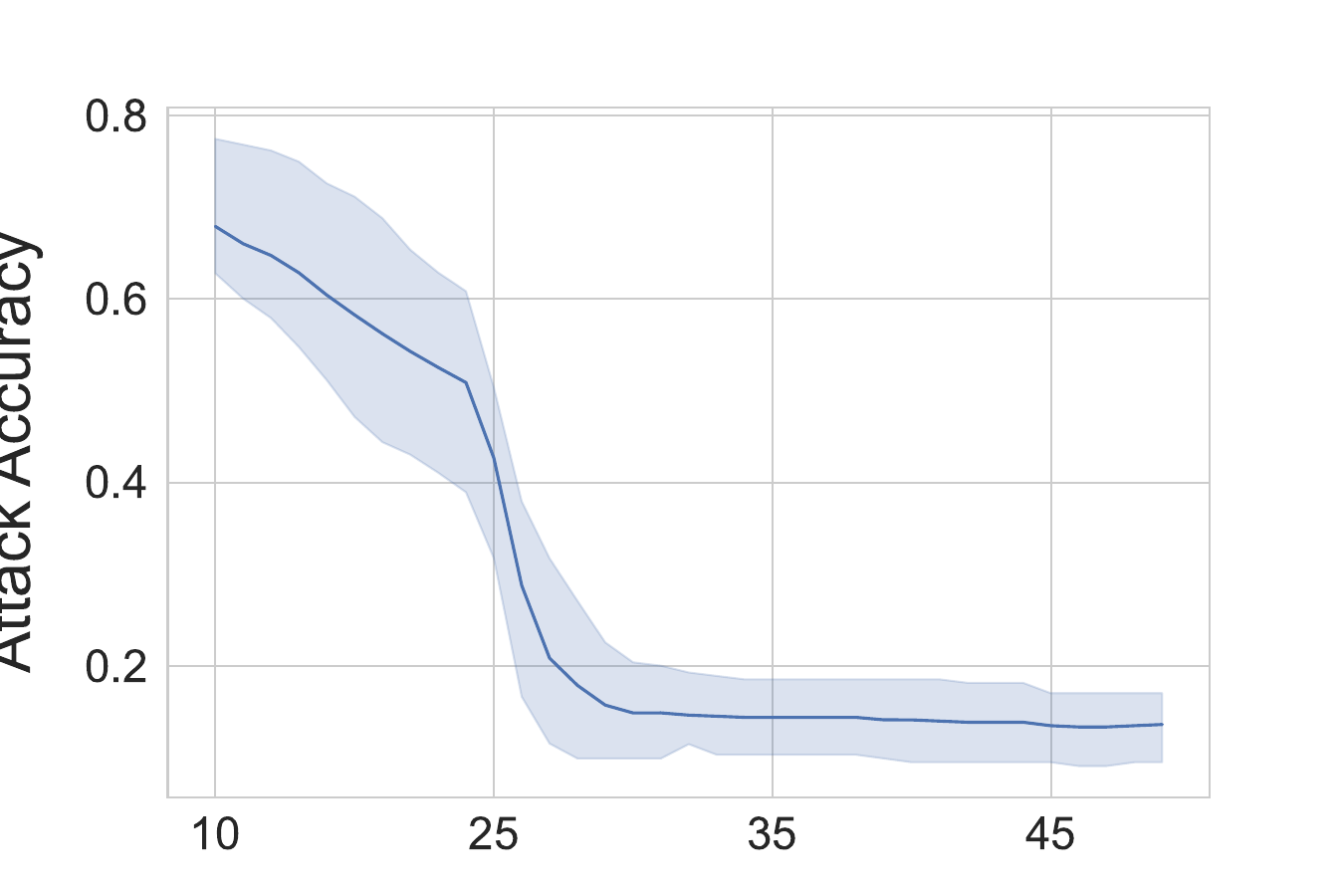}
        \label{poqemon_GA_MIA_5}
    }
    \hfill
    \subfloat[Susy]{
        \includegraphics[width=0.48\columnwidth]{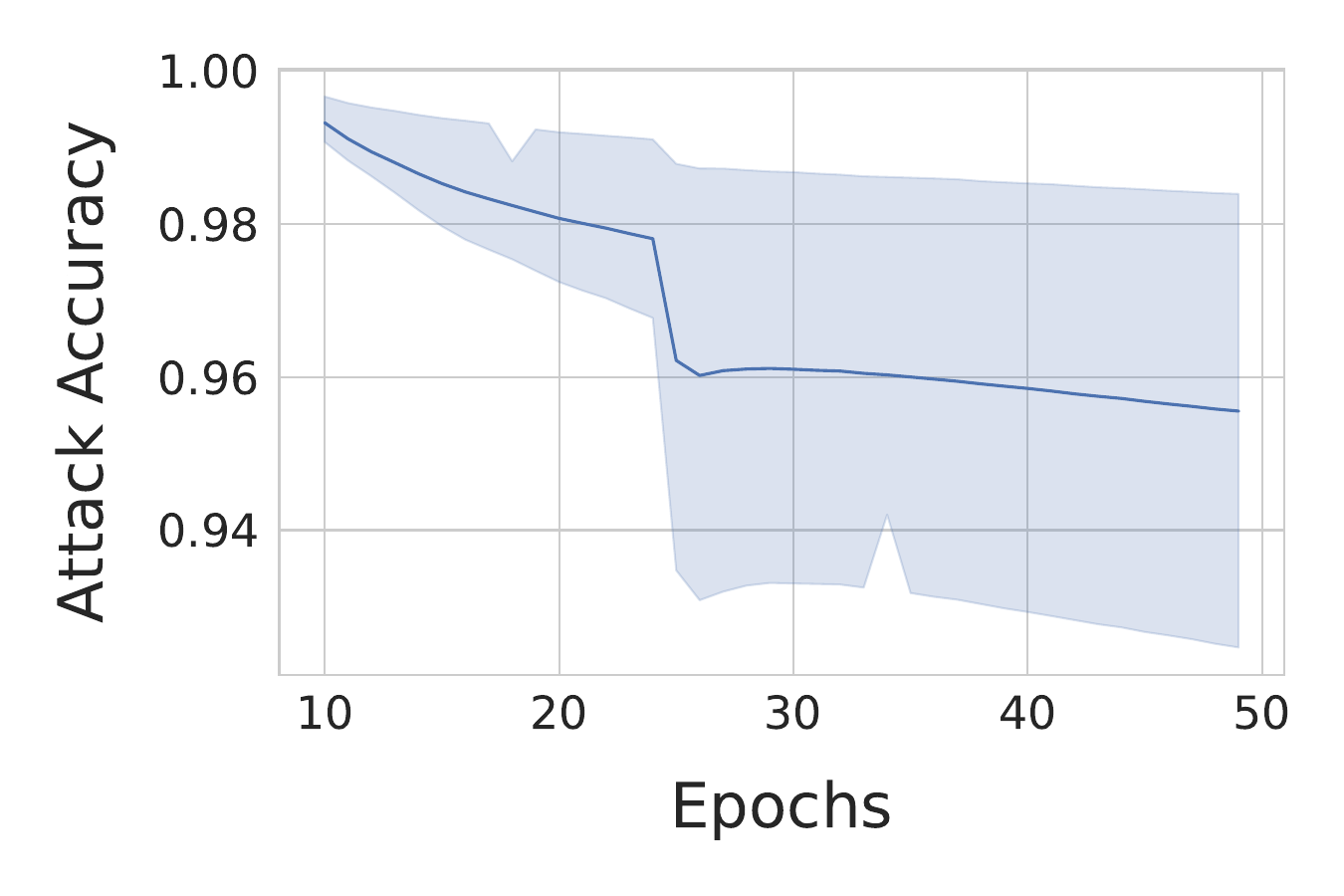}
        \label{susy_GA_MIA_5}
    }
    \hfill
    \subfloat[Wine]{
        \includegraphics[width=0.48\columnwidth]{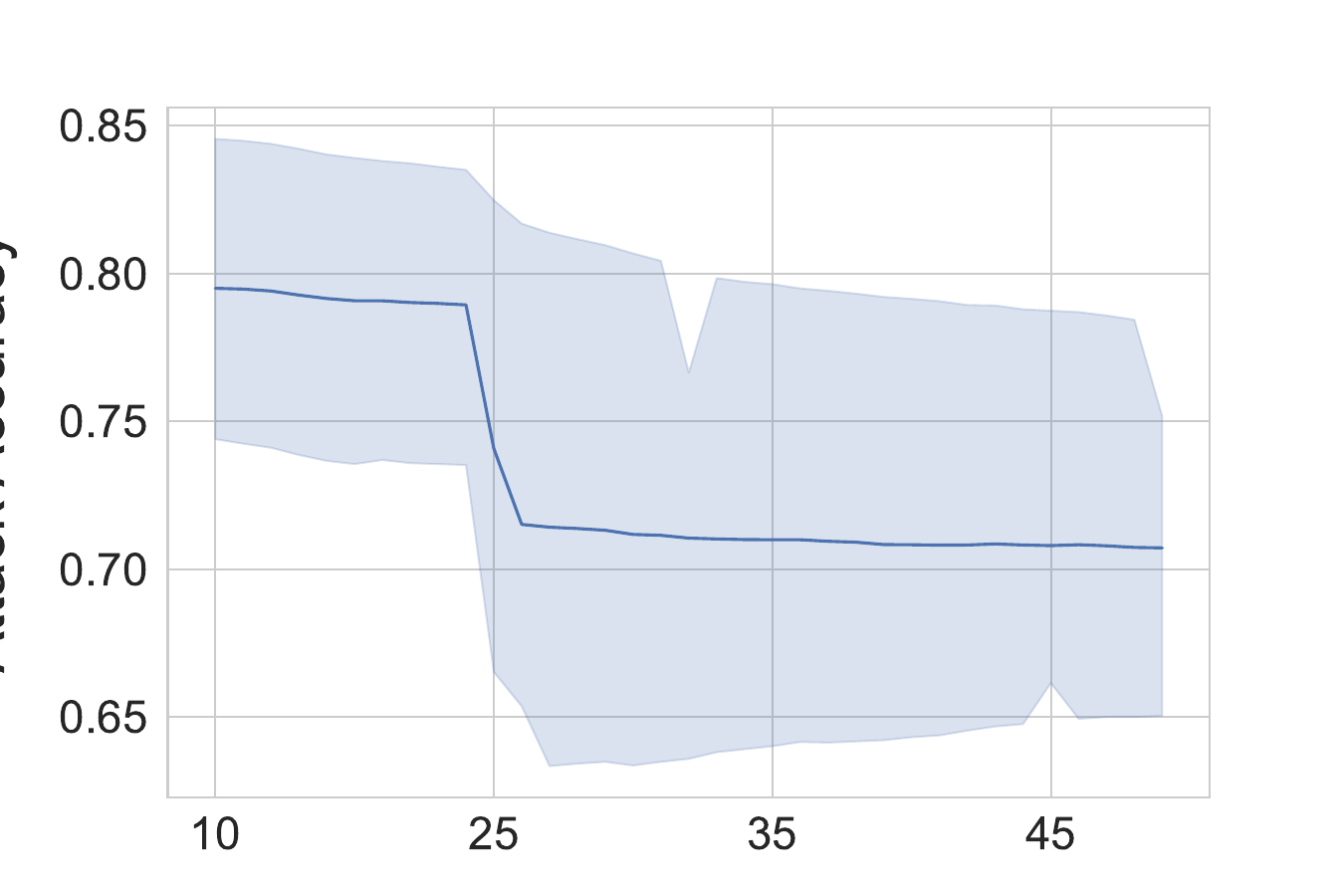}
        \label{wine_GA_MIA_5}
    }
    \caption{The MIA attack accuracy (y-axis) of VFU-GA.} 
    \label{Sample_unlearning_MIA} 
\end{figure}

\subsection{Auditing VFU-KD}

In this section, we evaluate the effectiveness of the unlearning process using a \ac{MIA}. As discussed in Section ~\ref{sec:MIA}, we train an MIA model with the output logits for 10 epochs both in the presence and absence of $clientA$. The MIA model consists of a single hidden layer comprising 32 neurons. The output layer of the MIA model is a binary classifier which predicts whether $clientA$ was present during training or not. 

Fig. ~\ref{MIA_accuracy} shows the accuracy of the \ac{MIA} on the tabular dataset starting from epoch 10 onwards. The results clearly demonstrate a significant drop in MIA accuracy at the 25th epoch, thereby indicating the effect of the unlearning process. Similar results were observed in Fig. \ref{Sample_unlearning_MIA} on all the datasets for \ac{MIA} attack accuracy. However, the drop in accuracy can vary with the impact of samples unlearned. 

\subsection{Limitations}

Based on our analysis and the results obtained, we highlight the following limitations of our approach.
\begin{enumerate}
    \item \textit{Attack Vulnerability}: Dishonest or honest-but-curious passive parties could potentially exploit the spikes caused by distillation to perform membership inference attacks or gradient based attacks.
    \item \textit{Limited Heterogeneity}: Similar to many existing works in the VFL literature, our approach assumes limited data heterogeneity and that all passive parties are readily available for training, with no stragglers.
    \item \textit{Unlearning Auditing}: For auditing unlearning, we employ a relatively weak membership inference attack (MIA) model. Utilizing a stronger model \cite{hu2022membership}, could yield more insightful results.
    \item \textit{Additional Storage}: Active party stores the communicated embeddings. This might be problematic where active and passive parties are in area which governs different data regulatory laws.
\end{enumerate}

\section{Conclusion and Future work} \label{Conclusion}

In this paper, we introduced a framework for unlearning in vertical federated learning (VFL), focusing on passive party unlearning and feature unlearning using knowledge distillation, termed VFU-KD, and sample unlearning using gradient ascent, termed VFU-GA. \ac{VFL} is inherently communication intensive. Thus, an effective unlearning approach should aim to minimize the communication between the active and passive parties. In VFU-KD, the active party is responsible for passive party unlearning, while the respective passive party handles feature unlearning. VFU-KD leverages knowledge distillation for effective model compression and unlearning. On the other hand, since sample unlearning does not require model compression, gradient ascent provides a more computationally efficient option in VFU-GA.

Our approach does not require any communication between active party and passive party for unlearning. However, it requires that the active party stores the communicated embeddings. This is essential in order to not have any communication. We have also proposed a \ac{MIA} which can be used to audit unlearning in VFL. We have compared VFU-KD, VFU-GA with the gold standard unlearning model i.e., model retrained from scratch and R2S optimization based faster retraining, on 6 tabular datasets, and 2 image dataset. The results demonstrate that, with our approach both active and passive parties can perform unlearning without any significant utility loss.

In our experiments, we employed a simple binary classifier for the membership inference attack (MIA). However, leveraging a more advanced and robust MIA model, such as the one proposed in \cite{carlini2022membership}, could potentially yield more insightful and accurate results. Exploring this avenue remains a priority for future work. Additionally, investigating the relationship between the distillation-induced spikes and their susceptibility to membership inference attacks presents an intriguing research direction. Another interesting direction for future work is to develop strategies to reduce the storage overhead for the active party, further enhancing the efficiency and scalability of the approach.


\begin{acks}

This work was carried out during an internship at Ericsson Research and was partially supported by the Wallenberg AI, Autonomous Systems, and Software Program (WASP), funded by the Knut and Alice Wallenberg Foundation. The computational resources required for this research were provided by Ericsson and the Berzelius supercomputer provided by National Supercomputer Centre at Linköping University and the Knut and Alice Wallenberg foundation.

\end{acks}

\bibliographystyle{ACM-Reference-Format}
\bibliography{bibliography}


\begin{thebibliography}{41}


\ifx \showCODEN    \undefined \def \showCODEN     #1{\unskip}     \fi
\ifx \showDOI      \undefined \def \showDOI       #1{#1}\fi
\ifx \showISBNx    \undefined \def \showISBNx     #1{\unskip}     \fi
\ifx \showISBNxiii \undefined \def \showISBNxiii  #1{\unskip}     \fi
\ifx \showISSN     \undefined \def \showISSN      #1{\unskip}     \fi
\ifx \showLCCN     \undefined \def \showLCCN      #1{\unskip}     \fi
\ifx \shownote     \undefined \def \shownote      #1{#1}          \fi
\ifx \showarticletitle \undefined \def \showarticletitle #1{#1}   \fi
\ifx \showURL      \undefined \def \showURL       {\relax}        \fi
\providecommand\bibfield[2]{#2}
\providecommand\bibinfo[2]{#2}
\providecommand\natexlab[1]{#1}
\providecommand\showeprint[2][]{arXiv:#2}

\bibitem[Amour et~al\mbox{.}(2015)]%
        {amour2015building}
\bibfield{author}{\bibinfo{person}{Lamine Amour}, \bibinfo{person}{Souihi Sami}, \bibinfo{person}{Said Hoceini}, {and} \bibinfo{person}{Abdelhamid Mellouk}.} \bibinfo{year}{2015}\natexlab{}.
\newblock \showarticletitle{Building a large dataset for model-based QoE prediction in the mobile environment}. In \bibinfo{booktitle}{\emph{Proceedings of the 18th ACM International Conference on Modeling, Analysis and Simulation of Wireless and Mobile Systems}}. \bibinfo{pages}{313--317}.
\newblock


\bibitem[Bagdasaryan et~al\mbox{.}(2020)]%
        {bagdasaryan2020backdoor}
\bibfield{author}{\bibinfo{person}{Eugene Bagdasaryan}, \bibinfo{person}{Andreas Veit}, \bibinfo{person}{Yiqing Hua}, \bibinfo{person}{Deborah Estrin}, {and} \bibinfo{person}{Vitaly Shmatikov}.} \bibinfo{year}{2020}\natexlab{}.
\newblock \showarticletitle{How to backdoor federated learning}. In \bibinfo{booktitle}{\emph{International conference on artificial intelligence and statistics}}. PMLR, \bibinfo{pages}{2938--2948}.
\newblock


\bibitem[Bourtoule et~al\mbox{.}(2021)]%
        {bourtoule2021machine}
\bibfield{author}{\bibinfo{person}{Lucas Bourtoule}, \bibinfo{person}{Varun Chandrasekaran}, \bibinfo{person}{Christopher~A Choquette-Choo}, \bibinfo{person}{Hengrui Jia}, \bibinfo{person}{Adelin Travers}, \bibinfo{person}{Baiwu Zhang}, \bibinfo{person}{David Lie}, {and} \bibinfo{person}{Nicolas Papernot}.} \bibinfo{year}{2021}\natexlab{}.
\newblock \showarticletitle{Machine unlearning}. In \bibinfo{booktitle}{\emph{2021 IEEE Symposium on Security and Privacy (SP)}}. IEEE, \bibinfo{pages}{141--159}.
\newblock


\bibitem[Carlini et~al\mbox{.}(2022)]%
        {carlini2022membership}
\bibfield{author}{\bibinfo{person}{Nicholas Carlini}, \bibinfo{person}{Steve Chien}, \bibinfo{person}{Milad Nasr}, \bibinfo{person}{Shuang Song}, \bibinfo{person}{Andreas Terzis}, {and} \bibinfo{person}{Florian Tramer}.} \bibinfo{year}{2022}\natexlab{}.
\newblock \showarticletitle{Membership inference attacks from first principles}. In \bibinfo{booktitle}{\emph{2022 IEEE Symposium on Security and Privacy (SP)}}. IEEE, \bibinfo{pages}{1897--1914}.
\newblock


\bibitem[Coates et~al\mbox{.}(2011)]%
        {coates2011analysis}
\bibfield{author}{\bibinfo{person}{Adam Coates}, \bibinfo{person}{Andrew Ng}, {and} \bibinfo{person}{Honglak Lee}.} \bibinfo{year}{2011}\natexlab{}.
\newblock \showarticletitle{An analysis of single-layer networks in unsupervised feature learning}. In \bibinfo{booktitle}{\emph{Proceedings of the fourteenth international conference on artificial intelligence and statistics}}. JMLR Workshop and Conference Proceedings, \bibinfo{pages}{215--223}.
\newblock


\bibitem[Deng et~al\mbox{.}(2023)]%
        {deng2023vertical}
\bibfield{author}{\bibinfo{person}{Zihao Deng}, \bibinfo{person}{Zhaoyang Han}, \bibinfo{person}{Chuan Ma}, \bibinfo{person}{Ming Ding}, \bibinfo{person}{Long Yuan}, \bibinfo{person}{Chunpeng Ge}, {and} \bibinfo{person}{Zhe Liu}.} \bibinfo{year}{2023}\natexlab{}.
\newblock \showarticletitle{Vertical federated unlearning on the logistic regression model}.
\newblock \bibinfo{journal}{\emph{Electronics}} \bibinfo{volume}{12}, \bibinfo{number}{14} (\bibinfo{year}{2023}), \bibinfo{pages}{3182}.
\newblock


\bibitem[Dong et~al\mbox{.}(2013)]%
        {dong2013private}
\bibfield{author}{\bibinfo{person}{Changyu Dong}, \bibinfo{person}{Liqun Chen}, {and} \bibinfo{person}{Zikai Wen}.} \bibinfo{year}{2013}\natexlab{}.
\newblock \showarticletitle{When private set intersection meets big data: an efficient and scalable protocol}. In \bibinfo{booktitle}{\emph{Proceedings of the 2013 ACM SIGSAC conference on Computer \& communications security}}. \bibinfo{pages}{789--800}.
\newblock


\bibitem[Gao et~al\mbox{.}(2024)]%
        {gao2024complementary}
\bibfield{author}{\bibinfo{person}{Dashan Gao}, \bibinfo{person}{Sheng Wan}, \bibinfo{person}{Lixin Fan}, \bibinfo{person}{Xin Yao}, {and} \bibinfo{person}{Qiang Yang}.} \bibinfo{year}{2024}\natexlab{}.
\newblock \showarticletitle{Complementary Knowledge Distillation for Robust and Privacy-Preserving Model Serving in Vertical Federated Learning}. In \bibinfo{booktitle}{\emph{Proceedings of the AAAI Conference on Artificial Intelligence}}, Vol.~\bibinfo{volume}{38}. \bibinfo{pages}{19832--19839}.
\newblock


\bibitem[Ginart et~al\mbox{.}(2019)]%
        {ginart2019making}
\bibfield{author}{\bibinfo{person}{Antonio Ginart}, \bibinfo{person}{Melody Guan}, \bibinfo{person}{Gregory Valiant}, {and} \bibinfo{person}{James~Y Zou}.} \bibinfo{year}{2019}\natexlab{}.
\newblock \showarticletitle{Making ai forget you: Data deletion in machine learning}.
\newblock \bibinfo{journal}{\emph{Advances in neural information processing systems}}  \bibinfo{volume}{32} (\bibinfo{year}{2019}).
\newblock


\bibitem[Gou et~al\mbox{.}(2021)]%
        {gou2021knowledge}
\bibfield{author}{\bibinfo{person}{Jianping Gou}, \bibinfo{person}{Baosheng Yu}, \bibinfo{person}{Stephen~J Maybank}, {and} \bibinfo{person}{Dacheng Tao}.} \bibinfo{year}{2021}\natexlab{}.
\newblock \showarticletitle{Knowledge distillation: A survey}.
\newblock \bibinfo{journal}{\emph{International Journal of Computer Vision}} \bibinfo{volume}{129}, \bibinfo{number}{6} (\bibinfo{year}{2021}), \bibinfo{pages}{1789--1819}.
\newblock


\bibitem[Graves et~al\mbox{.}(2021)]%
        {graves2021amnesiac}
\bibfield{author}{\bibinfo{person}{Laura Graves}, \bibinfo{person}{Vineel Nagisetty}, {and} \bibinfo{person}{Vijay Ganesh}.} \bibinfo{year}{2021}\natexlab{}.
\newblock \showarticletitle{Amnesiac machine learning}. In \bibinfo{booktitle}{\emph{Proceedings of the AAAI Conference on Artificial Intelligence}}, Vol.~\bibinfo{volume}{35}. \bibinfo{pages}{11516--11524}.
\newblock


\bibitem[Halimi et~al\mbox{.}(2022)]%
        {halimi2022federated}
\bibfield{author}{\bibinfo{person}{Anisa Halimi}, \bibinfo{person}{Swanand Kadhe}, \bibinfo{person}{Ambrish Rawat}, {and} \bibinfo{person}{Nathalie Baracaldo}.} \bibinfo{year}{2022}\natexlab{}.
\newblock \showarticletitle{Federated unlearning: How to efficiently erase a client in fl?}
\newblock \bibinfo{journal}{\emph{arXiv preprint arXiv:2207.05521}} (\bibinfo{year}{2022}).
\newblock


\bibitem[Hinton et~al\mbox{.}(2015)]%
        {hinton2015distilling}
\bibfield{author}{\bibinfo{person}{Geoffrey Hinton}, \bibinfo{person}{Oriol Vinyals}, {and} \bibinfo{person}{Jeff Dean}.} \bibinfo{year}{2015}\natexlab{}.
\newblock \showarticletitle{Distilling the knowledge in a neural network}.
\newblock \bibinfo{journal}{\emph{arXiv preprint arXiv:1503.02531}} (\bibinfo{year}{2015}).
\newblock


\bibitem[Hu et~al\mbox{.}(2022)]%
        {hu2022membership}
\bibfield{author}{\bibinfo{person}{Hongsheng Hu}, \bibinfo{person}{Zoran Salcic}, \bibinfo{person}{Lichao Sun}, \bibinfo{person}{Gillian Dobbie}, \bibinfo{person}{Philip~S Yu}, {and} \bibinfo{person}{Xuyun Zhang}.} \bibinfo{year}{2022}\natexlab{}.
\newblock \showarticletitle{Membership inference attacks on machine learning: A survey}.
\newblock \bibinfo{journal}{\emph{ACM Computing Surveys (CSUR)}} \bibinfo{volume}{54}, \bibinfo{number}{11s} (\bibinfo{year}{2022}), \bibinfo{pages}{1--37}.
\newblock


\bibitem[Jiang et~al\mbox{.}(2024)]%
        {jiang2024towards}
\bibfield{author}{\bibinfo{person}{Yu Jiang}, \bibinfo{person}{Jiyuan Shen}, \bibinfo{person}{Ziyao Liu}, \bibinfo{person}{Chee~Wei Tan}, {and} \bibinfo{person}{Kwok-Yan Lam}.} \bibinfo{year}{2024}\natexlab{}.
\newblock \showarticletitle{Towards efficient and certified recovery from poisoning attacks in federated learning}.
\newblock \bibinfo{journal}{\emph{arXiv preprint arXiv:2401.08216}} (\bibinfo{year}{2024}).
\newblock


\bibitem[Kelly et~al\mbox{.}(2023)]%
        {kelly2023uci}
\bibfield{author}{\bibinfo{person}{Markelle Kelly}, \bibinfo{person}{Rachel Longjohn}, {and} \bibinfo{person}{Kolby Nottingham}.} \bibinfo{year}{2023}\natexlab{}.
\newblock \showarticletitle{The UCI machine learning repository}.
\newblock \bibinfo{journal}{\emph{URL https://archive. ics. uci. edu}} (\bibinfo{year}{2023}).
\newblock


\bibitem[Krizhevsky(2009)]%
        {krizhevsky2009cifar-10}
\bibfield{author}{\bibinfo{person}{Alex Krizhevsky}.} \bibinfo{year}{2009}\natexlab{}.
\newblock \showarticletitle{Learning Multiple Layers of Features from Tiny Images}.
\newblock  (\bibinfo{year}{2009}).
\newblock
\urldef\tempurl%
\url{https://api.semanticscholar.org/CorpusID:18268744}
\showURL{%
\tempurl}


\bibitem[Li et~al\mbox{.}(2023)]%
        {li2023subspace}
\bibfield{author}{\bibinfo{person}{Guanghao Li}, \bibinfo{person}{Li Shen}, \bibinfo{person}{Yan Sun}, \bibinfo{person}{Yue Hu}, \bibinfo{person}{Han Hu}, {and} \bibinfo{person}{Dacheng Tao}.} \bibinfo{year}{2023}\natexlab{}.
\newblock \showarticletitle{Subspace based federated unlearning}.
\newblock \bibinfo{journal}{\emph{arXiv preprint arXiv:2302.12448}} (\bibinfo{year}{2023}).
\newblock


\bibitem[Li et~al\mbox{.}(2021)]%
        {li2021survey}
\bibfield{author}{\bibinfo{person}{Qinbin Li}, \bibinfo{person}{Zeyi Wen}, \bibinfo{person}{Zhaomin Wu}, \bibinfo{person}{Sixu Hu}, \bibinfo{person}{Naibo Wang}, \bibinfo{person}{Yuan Li}, \bibinfo{person}{Xu Liu}, {and} \bibinfo{person}{Bingsheng He}.} \bibinfo{year}{2021}\natexlab{}.
\newblock \showarticletitle{A survey on federated learning systems: Vision, hype and reality for data privacy and protection}.
\newblock \bibinfo{journal}{\emph{IEEE Transactions on Knowledge and Data Engineering}} \bibinfo{volume}{35}, \bibinfo{number}{4} (\bibinfo{year}{2021}), \bibinfo{pages}{3347--3366}.
\newblock


\bibitem[Liu et~al\mbox{.}(2021b)]%
        {liu2021federaser}
\bibfield{author}{\bibinfo{person}{Gaoyang Liu}, \bibinfo{person}{Xiaoqiang Ma}, \bibinfo{person}{Yang Yang}, \bibinfo{person}{Chen Wang}, {and} \bibinfo{person}{Jiangchuan Liu}.} \bibinfo{year}{2021}\natexlab{b}.
\newblock \showarticletitle{Federaser: Enabling efficient client-level data removal from federated learning models}. In \bibinfo{booktitle}{\emph{2021 IEEE/ACM 29th International Symposium on Quality of Service (IWQOS)}}. IEEE, \bibinfo{pages}{1--10}.
\newblock


\bibitem[Liu et~al\mbox{.}(2019)]%
        {liu2019variance}
\bibfield{author}{\bibinfo{person}{Liyuan Liu}, \bibinfo{person}{Haoming Jiang}, \bibinfo{person}{Pengcheng He}, \bibinfo{person}{Weizhu Chen}, \bibinfo{person}{Xiaodong Liu}, \bibinfo{person}{Jianfeng Gao}, {and} \bibinfo{person}{Jiawei Han}.} \bibinfo{year}{2019}\natexlab{}.
\newblock \showarticletitle{On the variance of the adaptive learning rate and beyond}.
\newblock \bibinfo{journal}{\emph{arXiv preprint arXiv:1908.03265}} (\bibinfo{year}{2019}).
\newblock


\bibitem[Liu et~al\mbox{.}(2021a)]%
        {liu2021fate}
\bibfield{author}{\bibinfo{person}{Yang Liu}, \bibinfo{person}{Tao Fan}, \bibinfo{person}{Tianjian Chen}, \bibinfo{person}{Qian Xu}, {and} \bibinfo{person}{Qiang Yang}.} \bibinfo{year}{2021}\natexlab{a}.
\newblock \showarticletitle{Fate: An industrial grade platform for collaborative learning with data protection}.
\newblock \bibinfo{journal}{\emph{Journal of Machine Learning Research}} \bibinfo{volume}{22}, \bibinfo{number}{226} (\bibinfo{year}{2021}), \bibinfo{pages}{1--6}.
\newblock


\bibitem[Liu et~al\mbox{.}(2024)]%
        {liu2024vertical}
\bibfield{author}{\bibinfo{person}{Yang Liu}, \bibinfo{person}{Yan Kang}, \bibinfo{person}{Tianyuan Zou}, \bibinfo{person}{Yanhong Pu}, \bibinfo{person}{Yuanqin He}, \bibinfo{person}{Xiaozhou Ye}, \bibinfo{person}{Ye Ouyang}, \bibinfo{person}{Ya-Qin Zhang}, {and} \bibinfo{person}{Qiang Yang}.} \bibinfo{year}{2024}\natexlab{}.
\newblock \showarticletitle{Vertical federated learning: Concepts, advances, and challenges}.
\newblock \bibinfo{journal}{\emph{IEEE Transactions on Knowledge and Data Engineering}} (\bibinfo{year}{2024}).
\newblock


\bibitem[Lu and Ding(2020)]%
        {lu2020multi}
\bibfield{author}{\bibinfo{person}{Linpeng Lu} {and} \bibinfo{person}{Ning Ding}.} \bibinfo{year}{2020}\natexlab{}.
\newblock \showarticletitle{Multi-party private set intersection in vertical federated learning}. In \bibinfo{booktitle}{\emph{2020 IEEE 19th International Conference on Trust, Security and Privacy in Computing and Communications (TrustCom)}}. IEEE, \bibinfo{pages}{707--714}.
\newblock


\bibitem[McMahan et~al\mbox{.}(2017)]%
        {mcmahan2017communication}
\bibfield{author}{\bibinfo{person}{Brendan McMahan}, \bibinfo{person}{Eider Moore}, \bibinfo{person}{Daniel Ramage}, \bibinfo{person}{Seth Hampson}, {and} \bibinfo{person}{Blaise~Aguera y Arcas}.} \bibinfo{year}{2017}\natexlab{}.
\newblock \showarticletitle{Communication-efficient learning of deep networks from decentralized data}. In \bibinfo{booktitle}{\emph{Artificial intelligence and statistics}}. PMLR, \bibinfo{pages}{1273--1282}.
\newblock


\bibitem[Nasr et~al\mbox{.}(2019)]%
        {nasr2019comprehensive}
\bibfield{author}{\bibinfo{person}{Milad Nasr}, \bibinfo{person}{Reza Shokri}, {and} \bibinfo{person}{Amir Houmansadr}.} \bibinfo{year}{2019}\natexlab{}.
\newblock \showarticletitle{Comprehensive privacy analysis of deep learning: Passive and active white-box inference attacks against centralized and federated learning}. In \bibinfo{booktitle}{\emph{2019 IEEE symposium on security and privacy (SP)}}. IEEE, \bibinfo{pages}{739--753}.
\newblock


\bibitem[Prayitno et~al\mbox{.}(2021)]%
        {prayitno2021systematic}
\bibfield{author}{\bibinfo{person}{Prayitno}, \bibinfo{person}{Chi-Ren Shyu}, \bibinfo{person}{Karisma~Trinanda Putra}, \bibinfo{person}{Hsing-Chung Chen}, \bibinfo{person}{Yuan-Yu Tsai}, \bibinfo{person}{KSM~Tozammel Hossain}, \bibinfo{person}{Wei Jiang}, {and} \bibinfo{person}{Zon-Yin Shae}.} \bibinfo{year}{2021}\natexlab{}.
\newblock \showarticletitle{A systematic review of federated learning in the healthcare area: From the perspective of data properties and applications}.
\newblock \bibinfo{journal}{\emph{Applied Sciences}} \bibinfo{volume}{11}, \bibinfo{number}{23} (\bibinfo{year}{2021}), \bibinfo{pages}{11191}.
\newblock


\bibitem[Sebbouh et~al\mbox{.}(2022)]%
        {sebbouh2022randomized}
\bibfield{author}{\bibinfo{person}{Othmane Sebbouh}, \bibinfo{person}{Marco Cuturi}, {and} \bibinfo{person}{Gabriel Peyr{\'e}}.} \bibinfo{year}{2022}\natexlab{}.
\newblock \showarticletitle{Randomized stochastic gradient descent ascent}. In \bibinfo{booktitle}{\emph{International Conference on Artificial Intelligence and Statistics}}. PMLR, \bibinfo{pages}{2941--2969}.
\newblock


\bibitem[Tarun et~al\mbox{.}(2023)]%
        {tarun2023fast}
\bibfield{author}{\bibinfo{person}{Ayush~K Tarun}, \bibinfo{person}{Vikram~S Chundawat}, \bibinfo{person}{Murari Mandal}, {and} \bibinfo{person}{Mohan Kankanhalli}.} \bibinfo{year}{2023}\natexlab{}.
\newblock \showarticletitle{Fast yet effective machine unlearning}.
\newblock \bibinfo{journal}{\emph{IEEE Transactions on Neural Networks and Learning Systems}} (\bibinfo{year}{2023}).
\newblock


\bibitem[Thudi et~al\mbox{.}(2022)]%
        {thudi2022unrolling}
\bibfield{author}{\bibinfo{person}{Anvith Thudi}, \bibinfo{person}{Gabriel Deza}, \bibinfo{person}{Varun Chandrasekaran}, {and} \bibinfo{person}{Nicolas Papernot}.} \bibinfo{year}{2022}\natexlab{}.
\newblock \showarticletitle{Unrolling sgd: Understanding factors influencing machine unlearning}. In \bibinfo{booktitle}{\emph{2022 IEEE 7th European Symposium on Security and Privacy (EuroS\&P)}}. IEEE, \bibinfo{pages}{303--319}.
\newblock


\bibitem[Wang et~al\mbox{.}(2023)]%
        {wang2023federated}
\bibfield{author}{\bibinfo{person}{Fei Wang}, \bibinfo{person}{Baochun Li}, {and} \bibinfo{person}{Bo Li}.} \bibinfo{year}{2023}\natexlab{}.
\newblock \showarticletitle{Federated unlearning and its privacy threats}.
\newblock \bibinfo{journal}{\emph{IEEE Network}} (\bibinfo{year}{2023}), \bibinfo{pages}{463--480}.
\newblock
\urldef\tempurl%
\url{https://doi.org/10.1109/MNET.004.2300056}
\showDOI{\tempurl}


\bibitem[Wang et~al\mbox{.}(2024)]%
        {wang2024efficient}
\bibfield{author}{\bibinfo{person}{Zichen Wang}, \bibinfo{person}{Xiangshan Gao}, \bibinfo{person}{Cong Wang}, \bibinfo{person}{Peng Cheng}, {and} \bibinfo{person}{Jiming Chen}.} \bibinfo{year}{2024}\natexlab{}.
\newblock \showarticletitle{Efficient Vertical Federated Unlearning via Fast Retraining}.
\newblock \bibinfo{journal}{\emph{ACM Transactions on Internet Technology}} \bibinfo{volume}{24}, \bibinfo{number}{2} (\bibinfo{year}{2024}), \bibinfo{pages}{1--22}.
\newblock


\bibitem[Warnecke et~al\mbox{.}(2021)]%
        {warnecke2021machine}
\bibfield{author}{\bibinfo{person}{Alexander Warnecke}, \bibinfo{person}{Lukas Pirch}, \bibinfo{person}{Christian Wressnegger}, {and} \bibinfo{person}{Konrad Rieck}.} \bibinfo{year}{2021}\natexlab{}.
\newblock \showarticletitle{Machine unlearning of features and labels}.
\newblock \bibinfo{journal}{\emph{arXiv preprint arXiv:2108.11577}} (\bibinfo{year}{2021}).
\newblock


\bibitem[Wei et~al\mbox{.}(2022)]%
        {wei2022vertical}
\bibfield{author}{\bibinfo{person}{Kang Wei}, \bibinfo{person}{Jun Li}, \bibinfo{person}{Chuan Ma}, \bibinfo{person}{Ming Ding}, \bibinfo{person}{Sha Wei}, \bibinfo{person}{Fan Wu}, \bibinfo{person}{Guihai Chen}, {and} \bibinfo{person}{Thilina Ranbaduge}.} \bibinfo{year}{2022}\natexlab{}.
\newblock \showarticletitle{Vertical federated learning: Challenges, methodologies and experiments}.
\newblock \bibinfo{journal}{\emph{arXiv preprint arXiv:2202.04309}} (\bibinfo{year}{2022}).
\newblock


\bibitem[Wu et~al\mbox{.}(2022)]%
        {wu2022federated}
\bibfield{author}{\bibinfo{person}{Chen Wu}, \bibinfo{person}{Sencun Zhu}, {and} \bibinfo{person}{Prasenjit Mitra}.} \bibinfo{year}{2022}\natexlab{}.
\newblock \showarticletitle{Federated unlearning with knowledge distillation}.
\newblock \bibinfo{journal}{\emph{arXiv preprint arXiv:2201.09441}} (\bibinfo{year}{2022}).
\newblock


\bibitem[Wu et~al\mbox{.}(2024)]%
        {wu2024federated}
\bibfield{author}{\bibinfo{person}{Zhaomin Wu}, \bibinfo{person}{Junyi Hou}, \bibinfo{person}{Yiqun Diao}, {and} \bibinfo{person}{Bingsheng He}.} \bibinfo{year}{2024}\natexlab{}.
\newblock \showarticletitle{Federated Transformer: Multi-Party Vertical Federated Learning on Practical Fuzzily Linked Data}.
\newblock \bibinfo{journal}{\emph{arXiv preprint arXiv:2410.17986}} (\bibinfo{year}{2024}).
\newblock


\bibitem[Yoon et~al\mbox{.}(2023)]%
        {yoon2023gradient}
\bibfield{author}{\bibinfo{person}{Dongkeun Yoon}, \bibinfo{person}{Joel Jang}, \bibinfo{person}{Sungdong Kim}, {and} \bibinfo{person}{Minjoon Seo}.} \bibinfo{year}{2023}\natexlab{}.
\newblock \showarticletitle{Gradient Ascent Post-training Enhances Language Model Generalization}. In \bibinfo{booktitle}{\emph{The 61st Annual Meeting Of The Association For Computational Linguistics}}.
\newblock


\bibitem[Yu et~al\mbox{.}(2024)]%
        {yu2024communication}
\bibfield{author}{\bibinfo{person}{Chong Yu}, \bibinfo{person}{Shuaiqi Shen}, \bibinfo{person}{Shiqiang Wang}, \bibinfo{person}{Kuan Zhang}, {and} \bibinfo{person}{Hai Zhao}.} \bibinfo{year}{2024}\natexlab{}.
\newblock \showarticletitle{Communication-Efficient Hybrid Federated Learning for E-Health With Horizontal and Vertical Data Partitioning}.
\newblock \bibinfo{journal}{\emph{IEEE Transactions on Neural Networks and Learning Systems}} (\bibinfo{year}{2024}).
\newblock


\bibitem[Zhang et~al\mbox{.}(2023)]%
        {zhang2023review}
\bibfield{author}{\bibinfo{person}{Haibo Zhang}, \bibinfo{person}{Toru Nakamura}, \bibinfo{person}{Takamasa Isohara}, {and} \bibinfo{person}{Kouichi Sakurai}.} \bibinfo{year}{2023}\natexlab{}.
\newblock \showarticletitle{A review on machine unlearning}.
\newblock \bibinfo{journal}{\emph{SN Computer Science}} \bibinfo{volume}{4}, \bibinfo{number}{4} (\bibinfo{year}{2023}), \bibinfo{pages}{337}.
\newblock


\bibitem[Zheng et~al\mbox{.}(2020)]%
        {zheng2020vertical}
\bibfield{author}{\bibinfo{person}{Fanglan Zheng}, \bibinfo{person}{Kun Li}, \bibinfo{person}{Jiang Tian}, \bibinfo{person}{Xiaojia Xiang}, {et~al\mbox{.}}} \bibinfo{year}{2020}\natexlab{}.
\newblock \showarticletitle{A vertical federated learning method for interpretable scorecard and its application in credit scoring}.
\newblock \bibinfo{journal}{\emph{arXiv preprint arXiv:2009.06218}} (\bibinfo{year}{2020}).
\newblock


\bibitem[Zhu et~al\mbox{.}(2023)]%
        {zhu2023heterogeneous}
\bibfield{author}{\bibinfo{person}{Xiangrong Zhu}, \bibinfo{person}{Guangyao Li}, {and} \bibinfo{person}{Wei Hu}.} \bibinfo{year}{2023}\natexlab{}.
\newblock \showarticletitle{Heterogeneous federated knowledge graph embedding learning and unlearning}. In \bibinfo{booktitle}{\emph{Proceedings of the ACM web conference 2023}}. \bibinfo{pages}{2444--2454}.
\newblock


\end{thebibliography}

\appendix

\section{Hessian utility score}

\begin{figure}[h]
    \centering
    \subfloat[Adult]{
        \includegraphics[width=0.48\columnwidth]{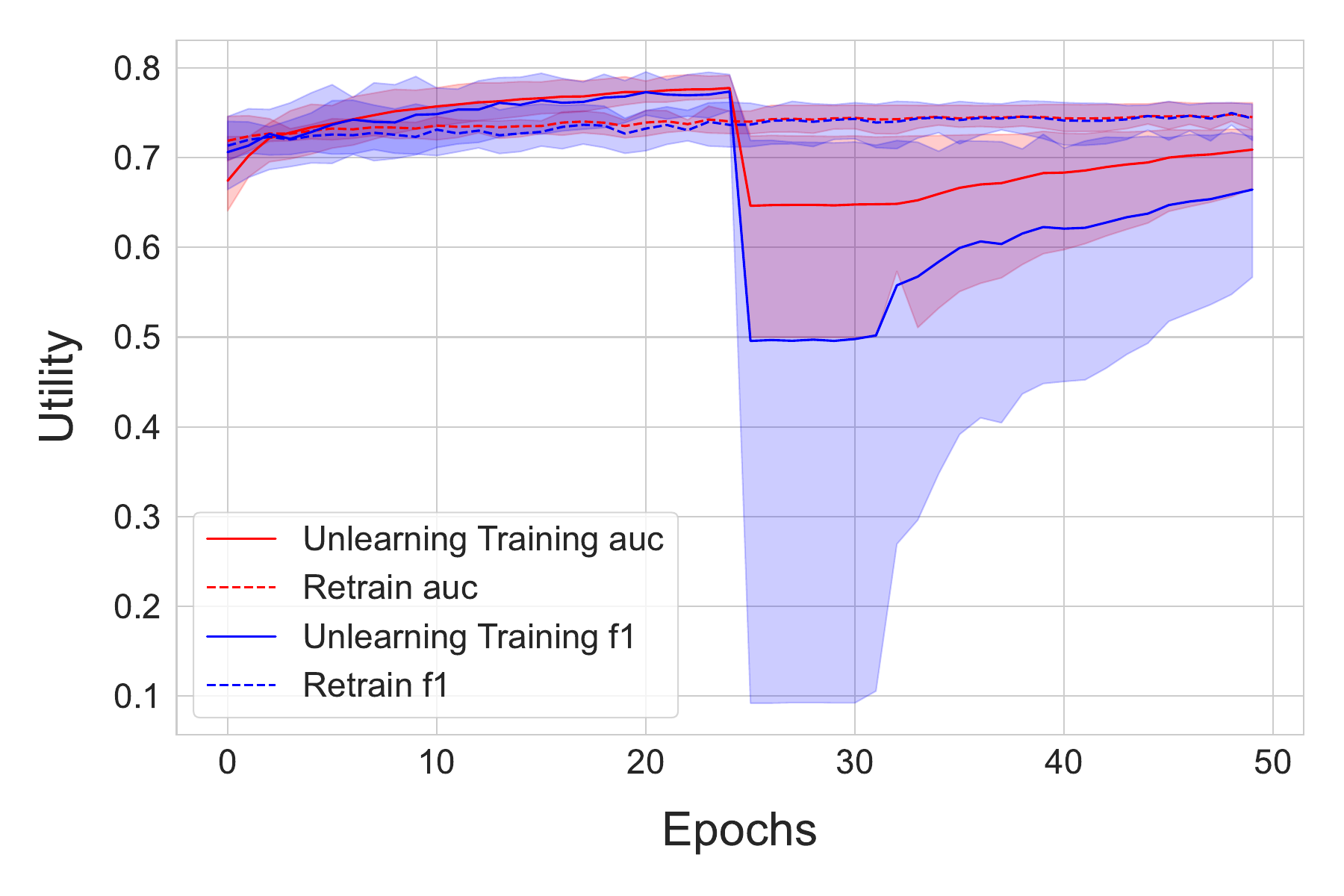}
        \label{adult_hess_utility}
    }
    \hfill
    \subfloat[ai4i]{
        \includegraphics[width=0.48\columnwidth]{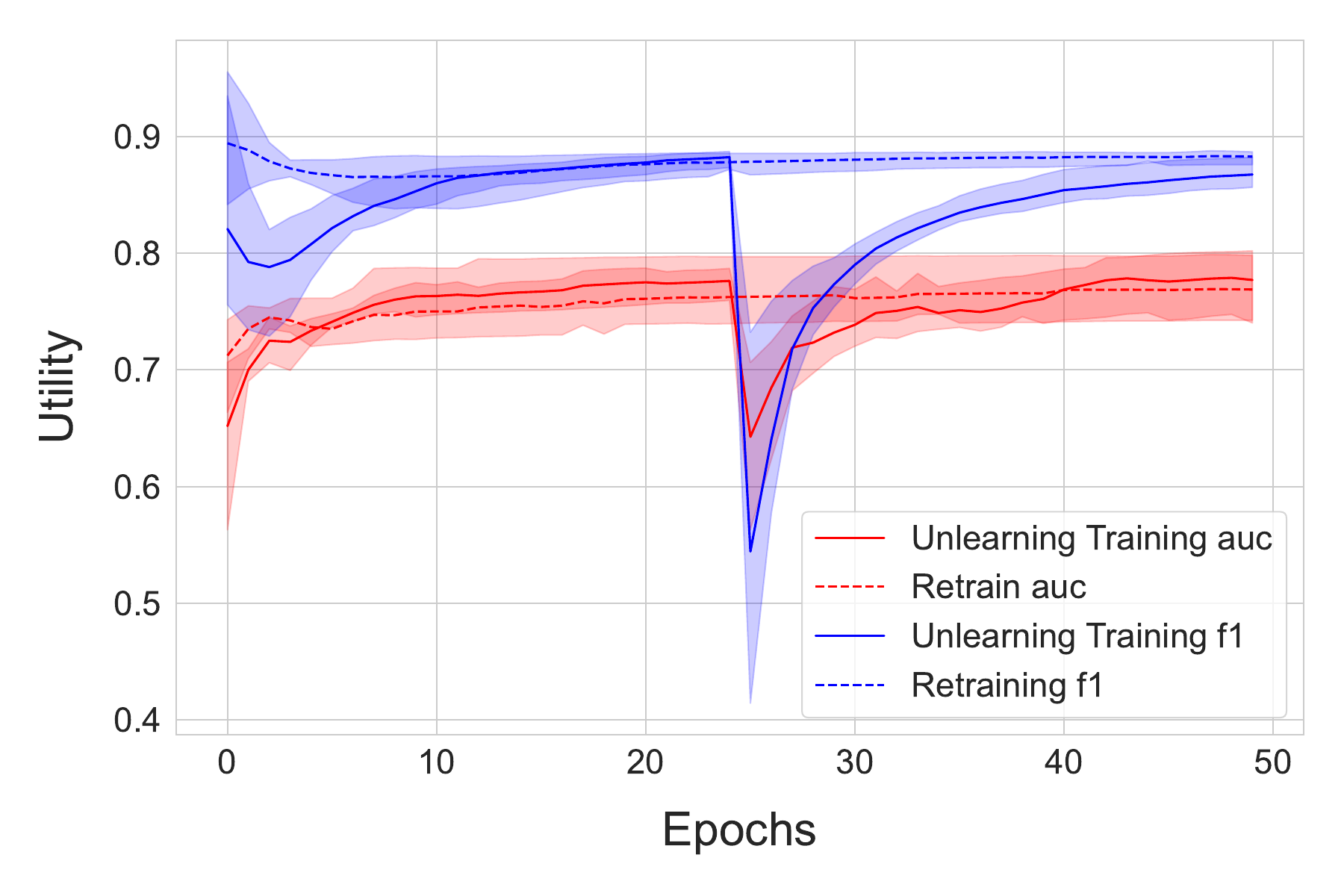}
        \label{ai4i_hess_utility}
    }
    \hfill
    \subfloat[Hepmass]{
        \includegraphics[width=0.48\columnwidth]{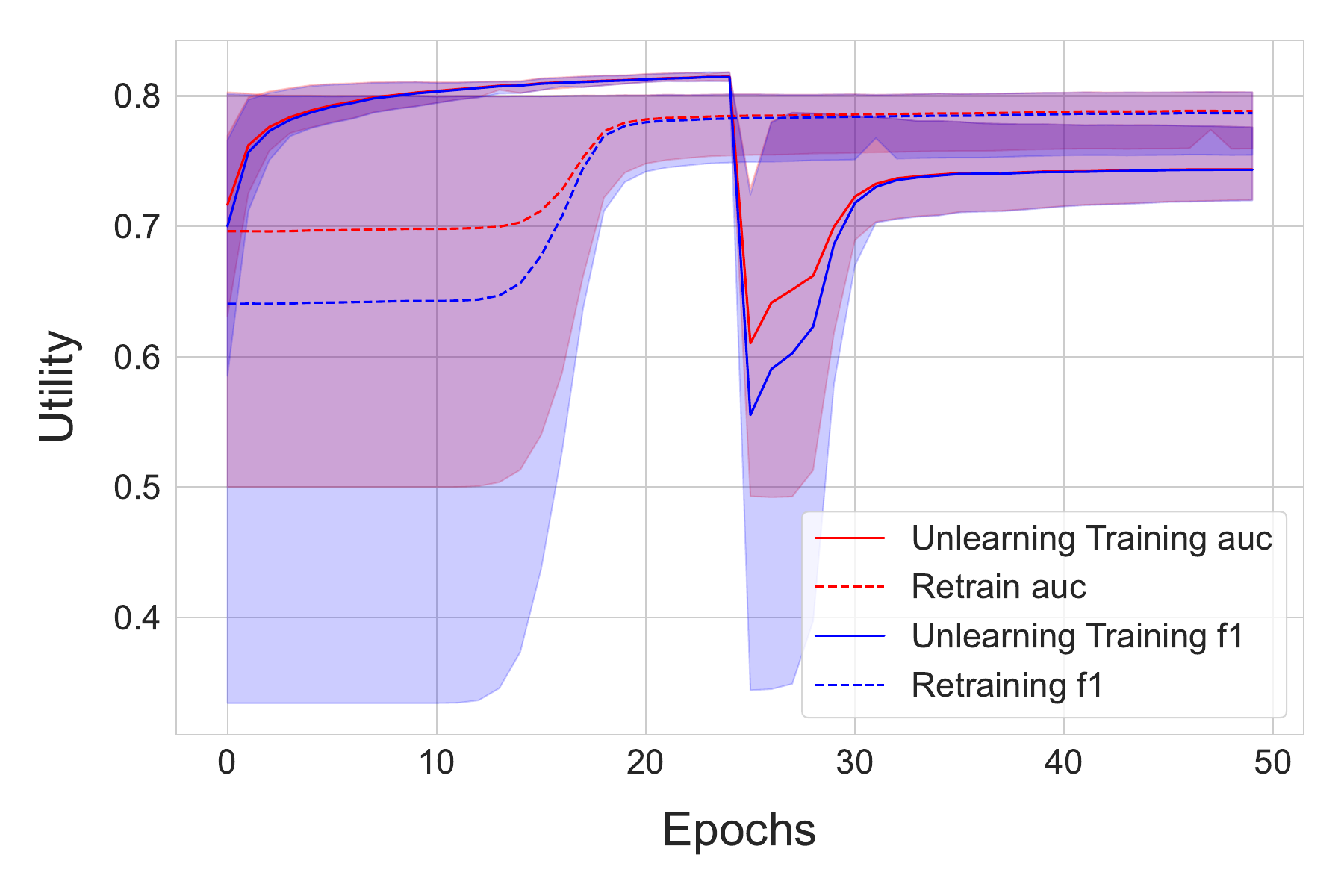}
        \label{hepmass_hess_utility}
    }
    \hfill
    \subfloat[Poqemon]{
        \includegraphics[width=0.48\columnwidth]{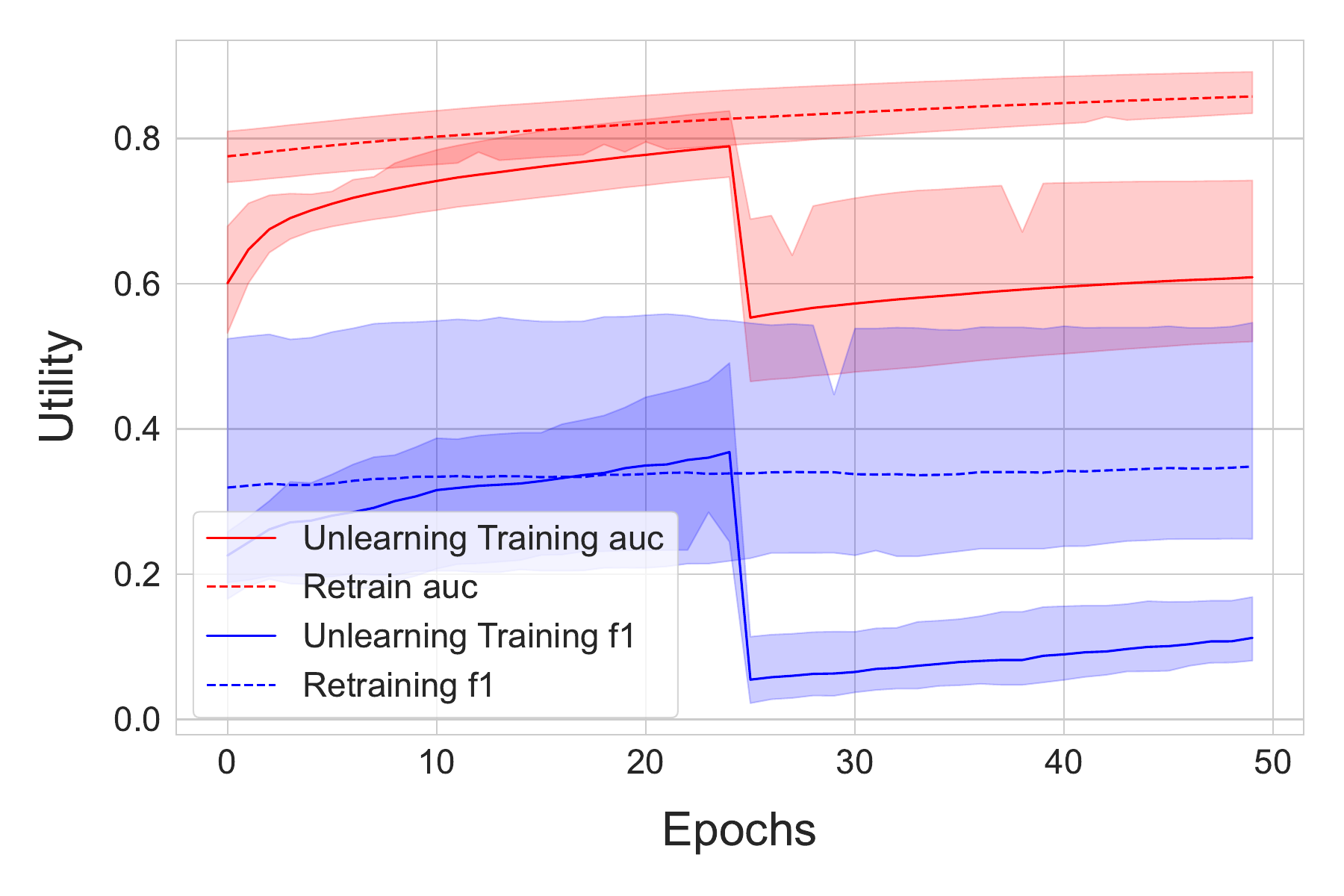}
        \label{poqemon_hess_utility}
    }
    \hfill
    \subfloat[Susy]{
        \includegraphics[width=0.48\columnwidth]{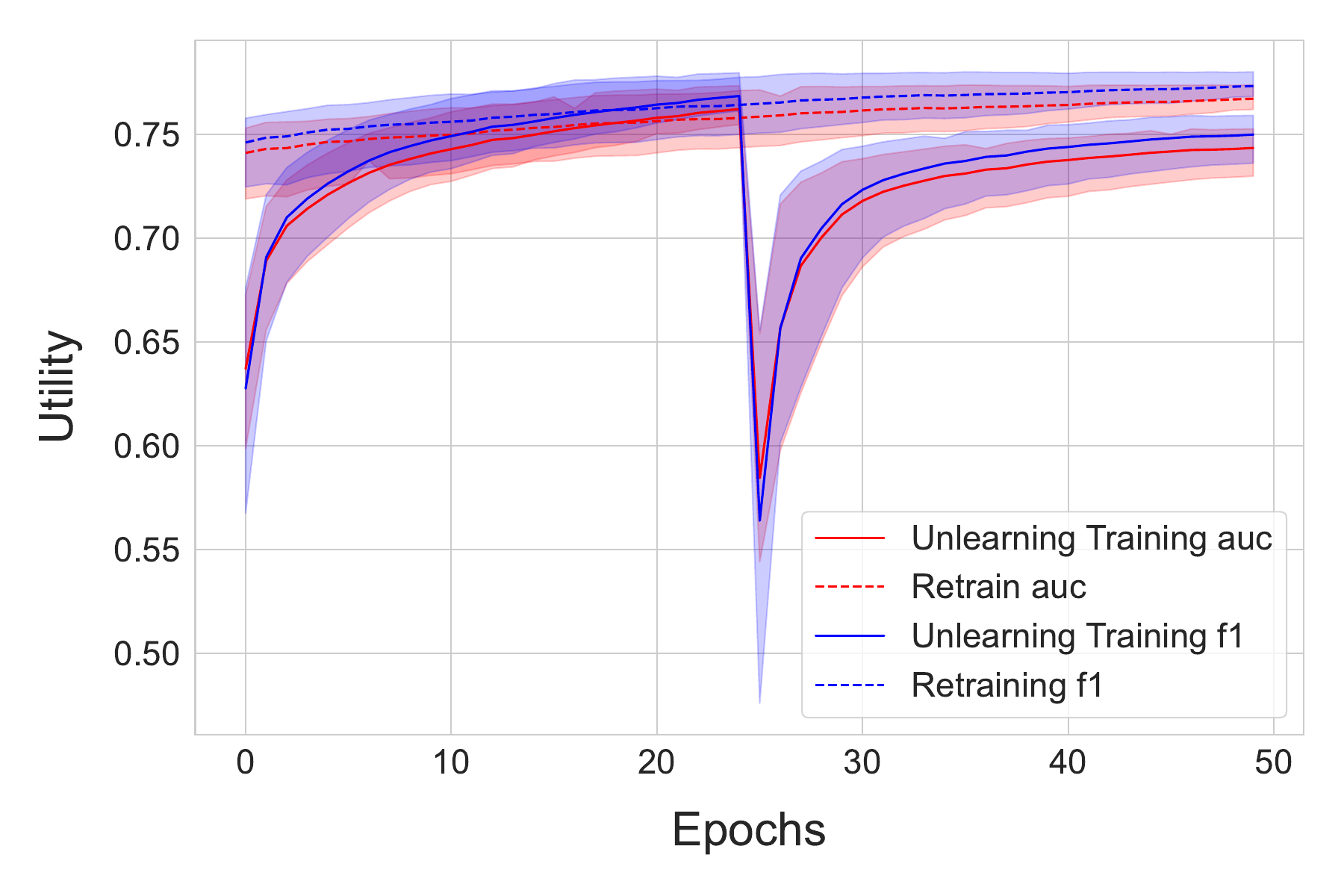}
        \label{susy_hess_utility}
    }
    \hfill
    \subfloat[Wine]{
        \includegraphics[width=0.48\columnwidth]{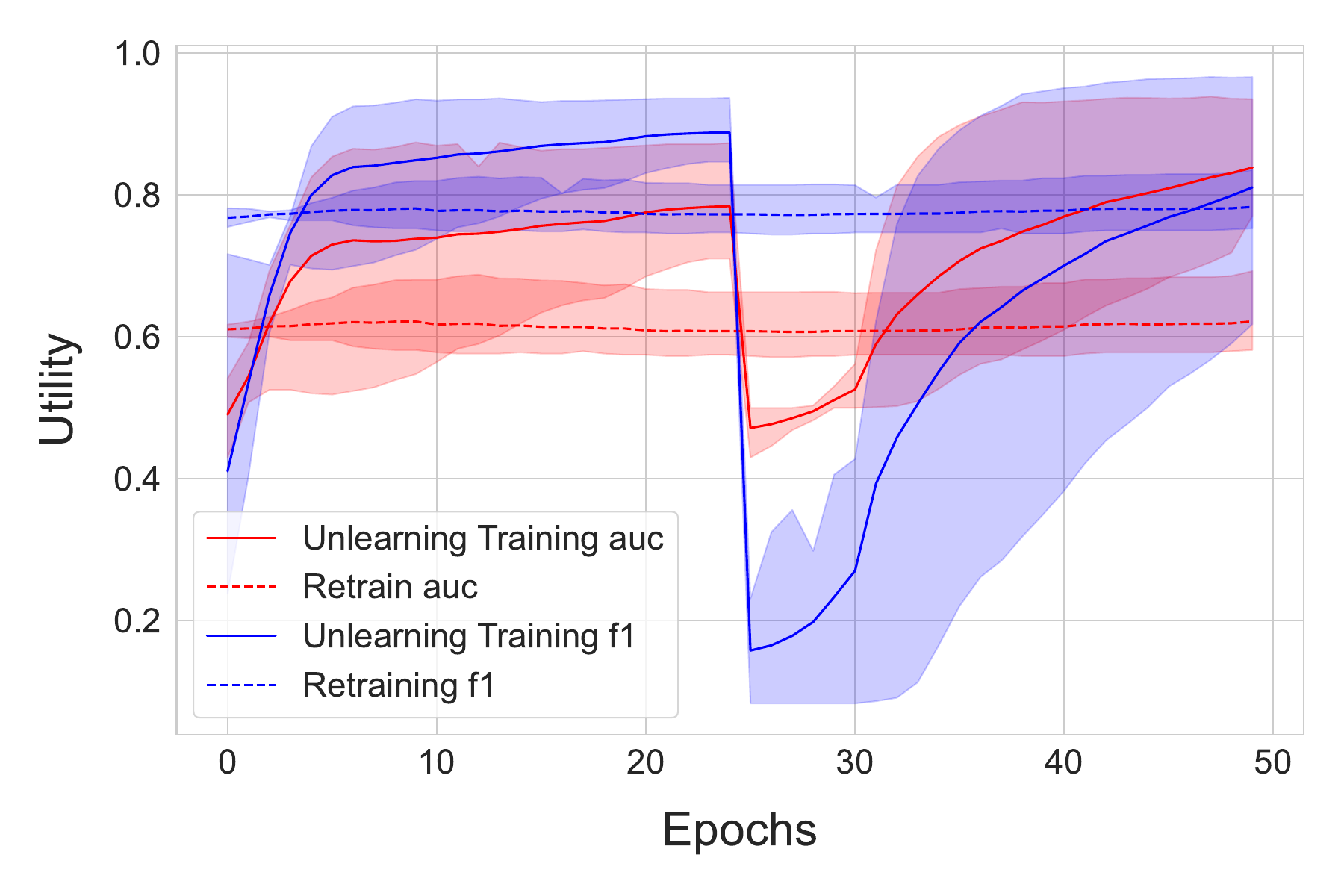}
        \label{wine_hess_utility}
    }
    \caption{The F1 and AUC scores of VFU-KD with $\mathcal{H}^{-1}$ compared to the retrained model from scratch.} 
    \label{utility_scores_hessian} 
\end{figure}

\begin{figure}
    \centering
    \subfloat[Two Most Important Features]{
        \includegraphics[width=0.48\columnwidth]{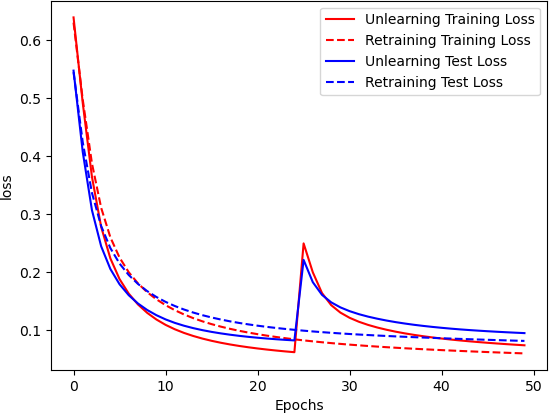}
        \label{wine_loss_2MI}
    }
    \hfill
    \subfloat[Two Least Important Features]{
        \includegraphics[width=0.48\columnwidth]{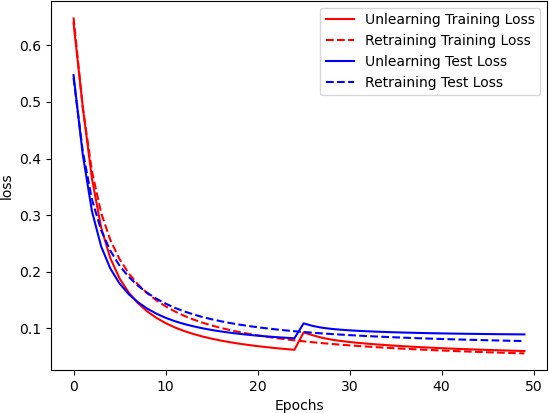}
        \label{wine_loss_2LI}
    }
    \hfill
    \subfloat[Most Important Feature from Each Client]{
        \includegraphics[width=0.48\columnwidth]{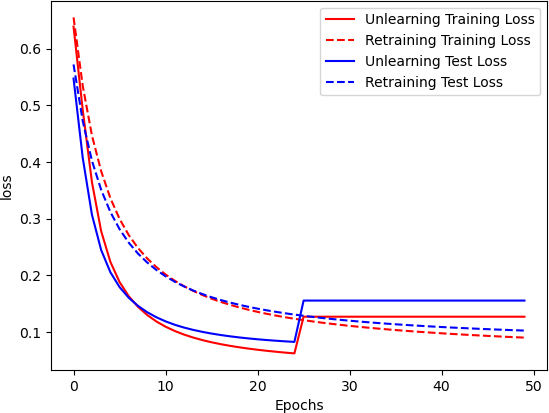}
        \label{wine_loss_MI_cl}
    }
    \hfill
    \subfloat[Least Important Feature from Each Client]{
        \includegraphics[width=0.48\columnwidth]{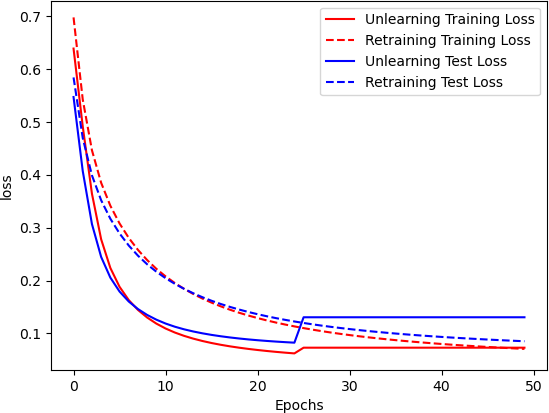}
        \label{wine_loss_LI_cl}
    }
    \caption{The training and test loss values of VFU-KD for the wine dataset.} 
    \label{loss_wineAdditional} 
\end{figure}

\begin{figure}
    \centering
    \subfloat[Two Most Important Features]{
        \includegraphics[width=0.48\columnwidth]{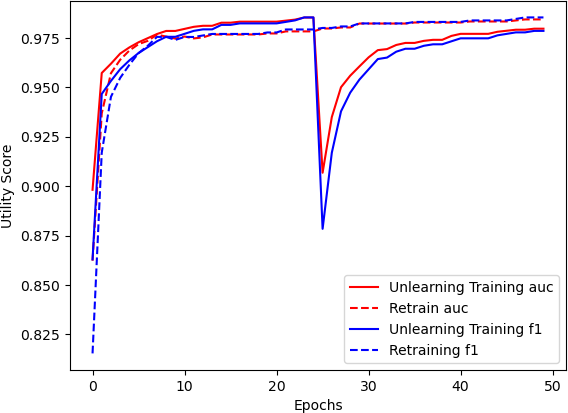}
        \label{wine_utility_2MI}
    }
    \hfill
    \subfloat[Two Least Important Features]{
        \includegraphics[width=0.48\columnwidth]{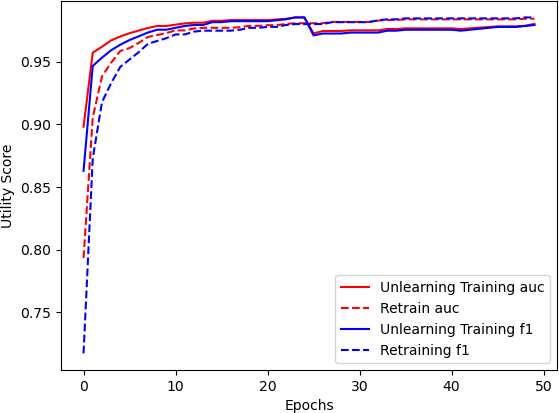}
        \label{wine_utility_2LI}
    }
    \hfill
    \subfloat[Most Important Feature from Each Client]{
        \includegraphics[width=0.48\columnwidth]{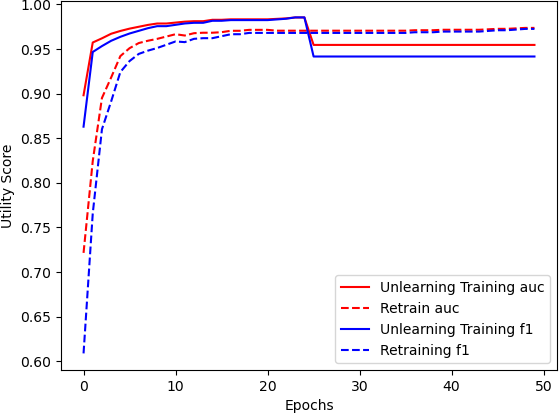}
        \label{wine_utility_MI_cl}
    }
    \hfill
    \subfloat[Least Important Feature from Each Client]{
        \includegraphics[width=0.48\columnwidth]{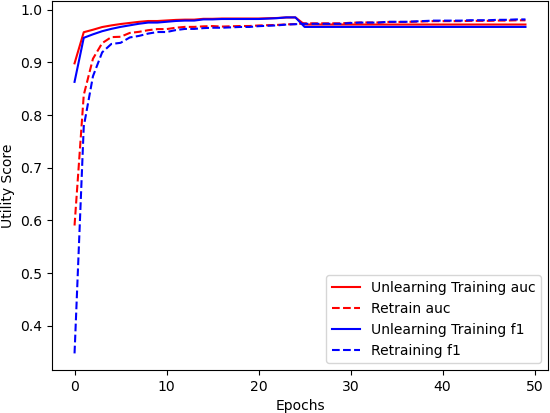}
        \label{wine_utility_LI_cl}
    }
    \caption{The F1 and AUC score of VFU-KD for the wine dataset.} 
    \label{utility_wineAdditional} 
\end{figure}

\begin{figure*}[h]
    \centering
    \subfloat[Adult]{
        \includegraphics[width=0.67\columnwidth]{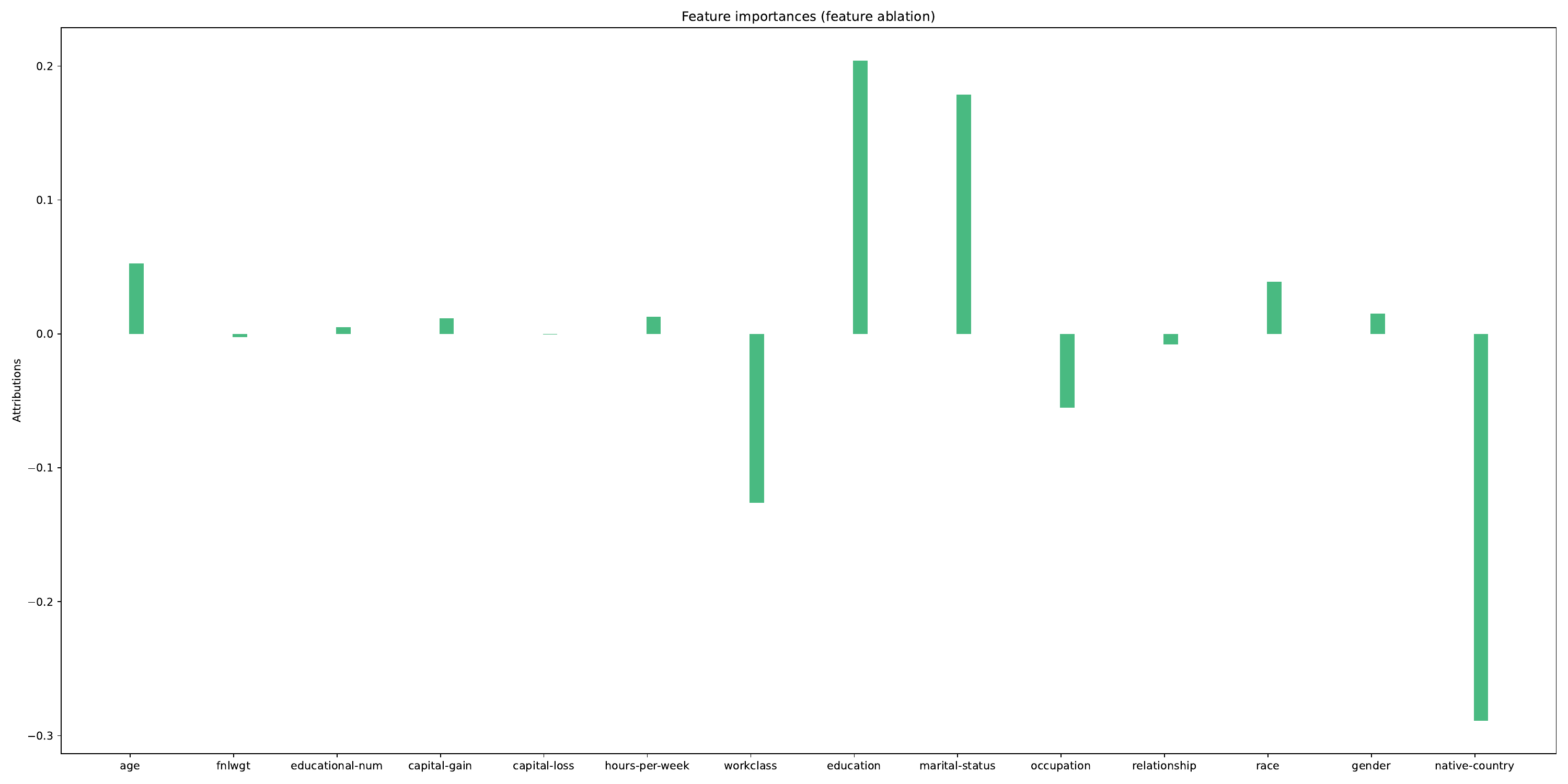}
        \label{adult_utility_feat}
    }
    \hfill
    \subfloat[ai4i]{
        \includegraphics[width=0.67\columnwidth]{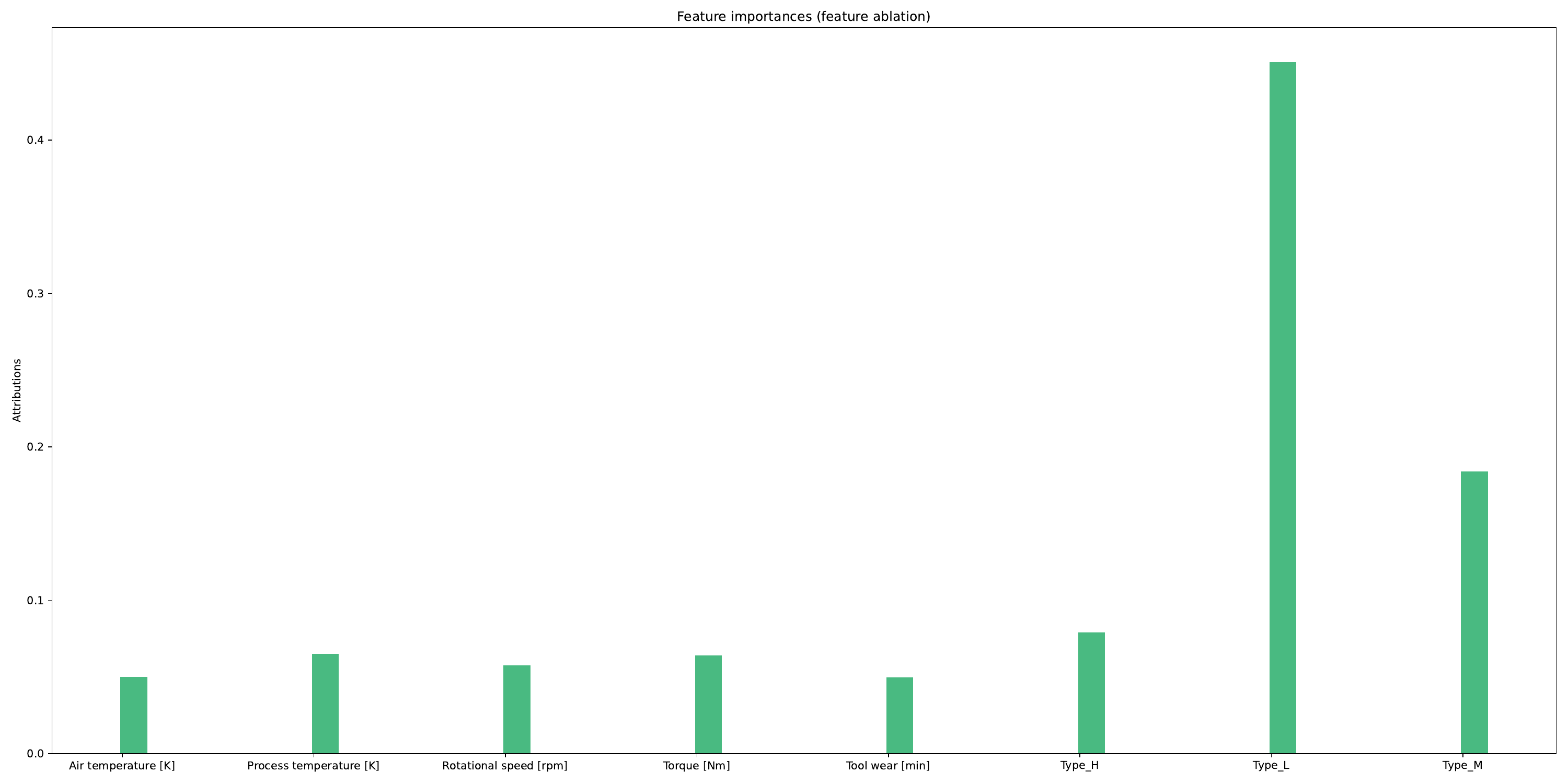}
        \label{ai4i_utility_feat}
    }
    \hfill
    \subfloat[Hepmass]{
        \includegraphics[width=0.67\columnwidth]{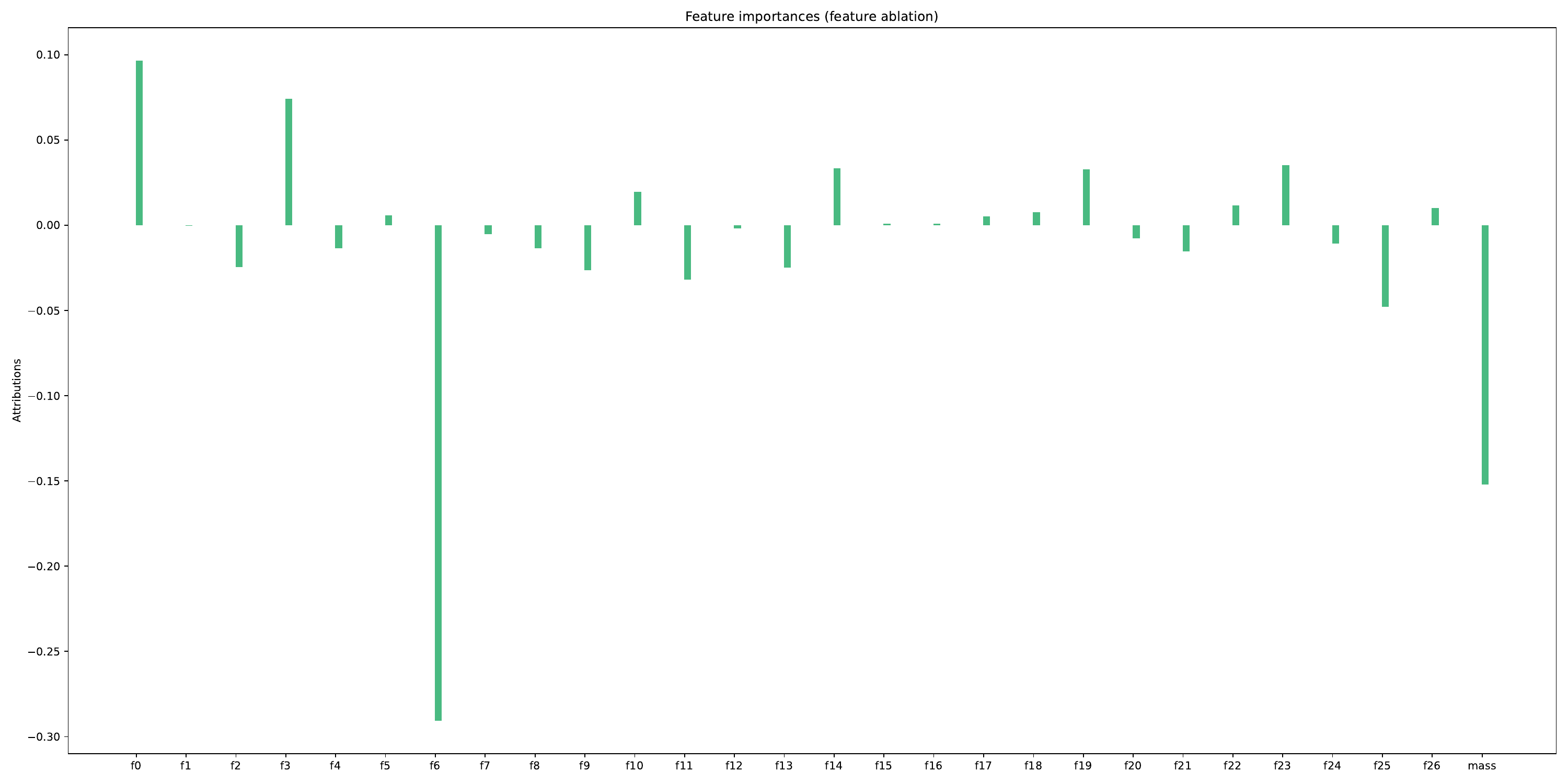}
        \label{hepmass_utility_feat}
    }
    \hfill
    \subfloat[Poqemon]{
        \includegraphics[width=0.67\columnwidth]{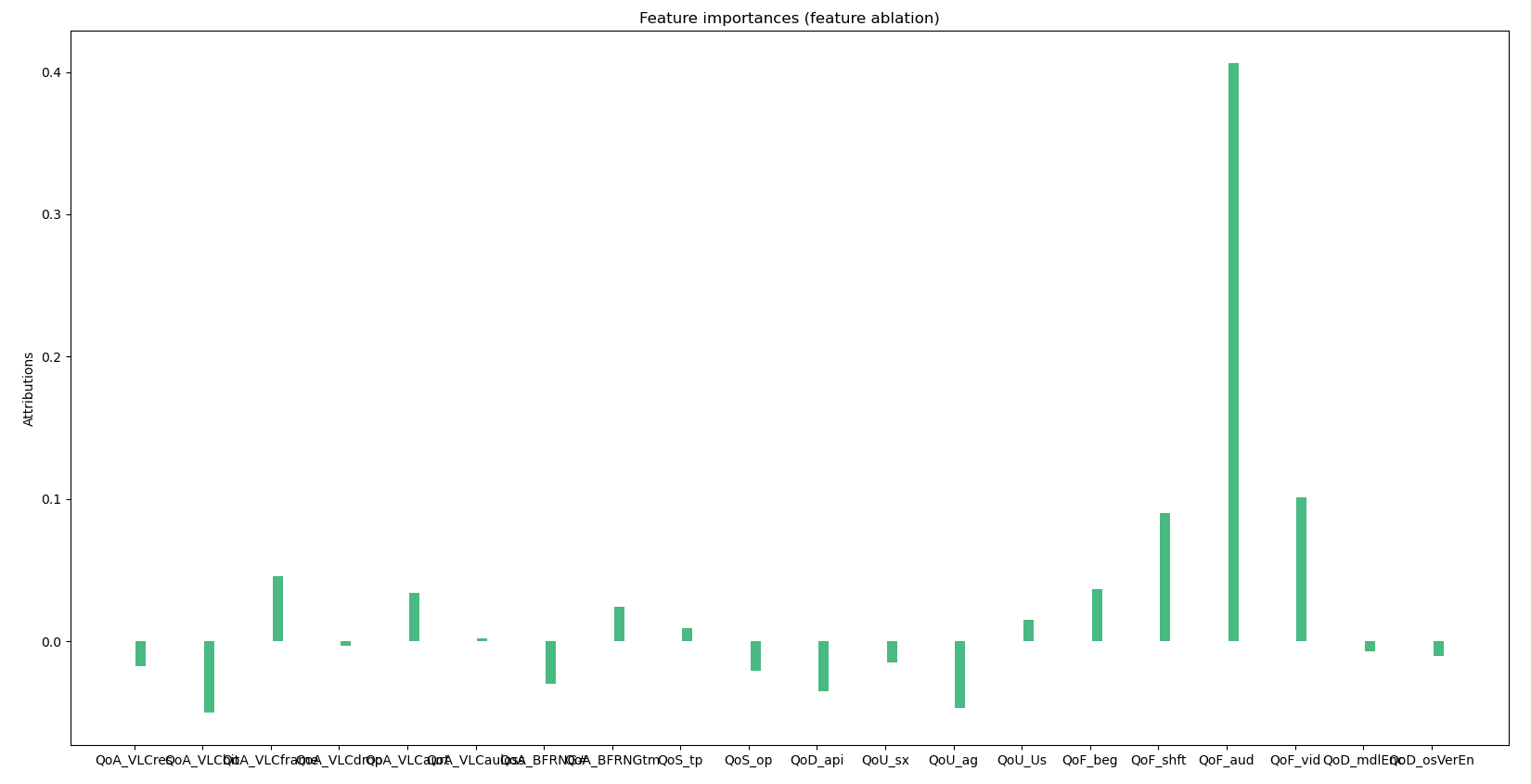}
        \label{poqemon_utility_feat}
    }
    \hfill
    \subfloat[Susy]{
        \includegraphics[width=0.67\columnwidth]{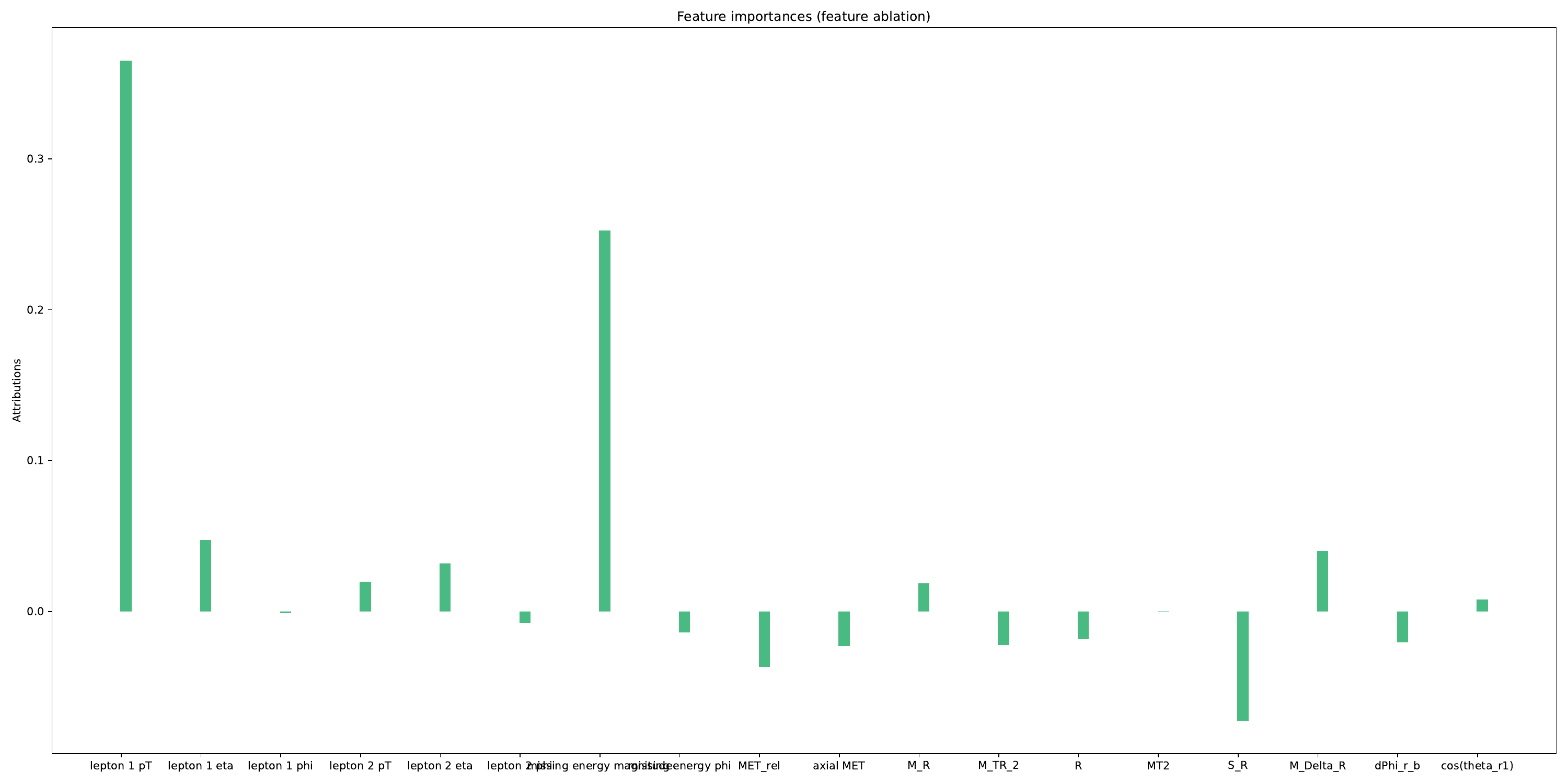}
        \label{susy_utility_feat}
    }
    \hfill
    \subfloat[Wine]{
        \includegraphics[width=0.67\columnwidth]{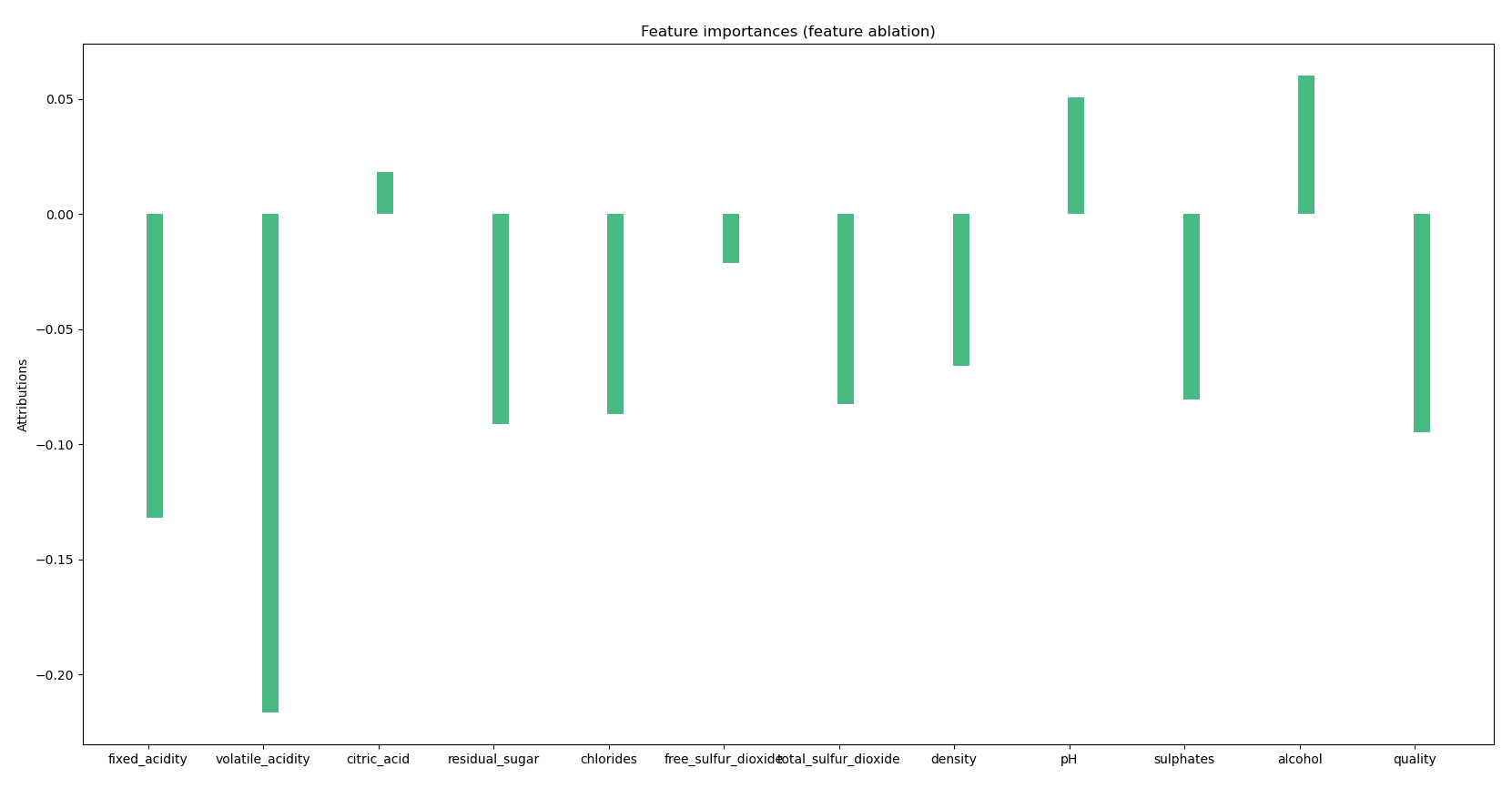}
        \label{wine_utility_feat}
    }
    \caption{The feature importance plot. Each bar, from left to right, represents the features in order from the first to the last column of the respective dataset.} 
    \label{feature_importance} 
\end{figure*}

Fig.~\ref{utility_scores_hessian} shows the F1 and AUC scores for client unlearning with $\mathcal{H}^{-1}$. Here as well, it is clear from the results that the utility score eventually converges to the benchmark comparable utility scores in all the cases except Hepmass and Poqemon dataset. We found that, we have better results in the presence of a learning rate than with $\mathcal{H}^{-1}$. 

\section{Additional feature importance results}

Fig.~\ref{feature_importance} highlights the importance of each feature using feature ablation. Based on this information, the results for most important and least important feature were given in the main paper. 

We have also experimented with removing multiple features from the same client, most important feature and least important feature from each clients for wine dataset. 

Fig.~\ref{loss_wineAdditional} shows the training and test loss when two most-least important features were unlearning, and most-least important feature from each client were unlearning. Similary, Fig.~\ref{utility_wineAdditional} shows the utility scores for the same. In all the cases, we see that, VFU-KD achieves benchmark comparable results, even better than benchmark in some cases. 

\section{Additional VFU-GA results}

\begin{figure}[h]
    \centering
    \subfloat[Adult]{
        \includegraphics[width=0.48\columnwidth]{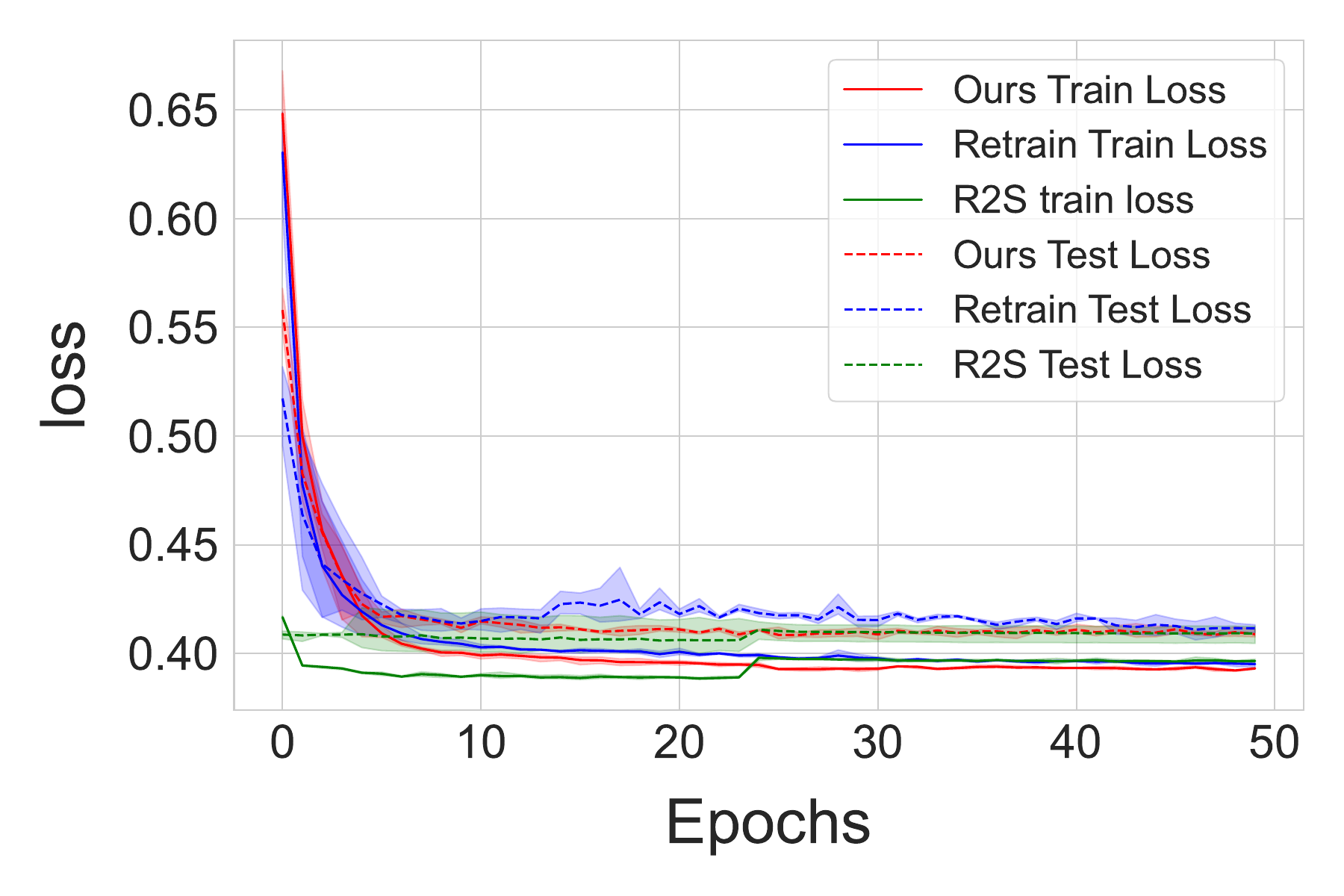}
        \label{adult_GA_loss_1}
    }
    \hfill
    \subfloat[ai4i]{
        \includegraphics[width=0.48\columnwidth]{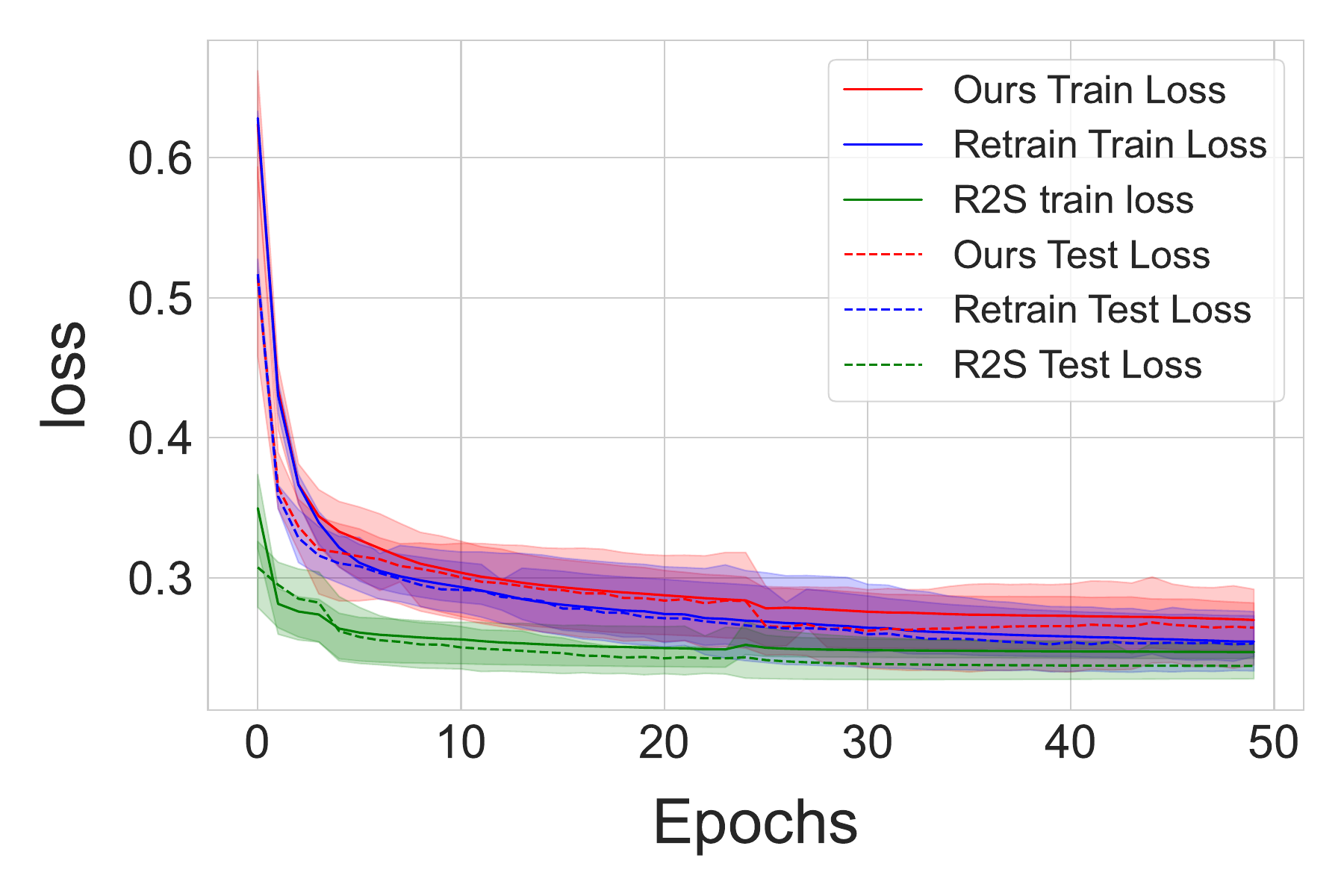}
        \label{ai4i_GA_loss_1}
    }
    \hfill
    \subfloat[Hepmass]{
        \includegraphics[width=0.48\columnwidth]{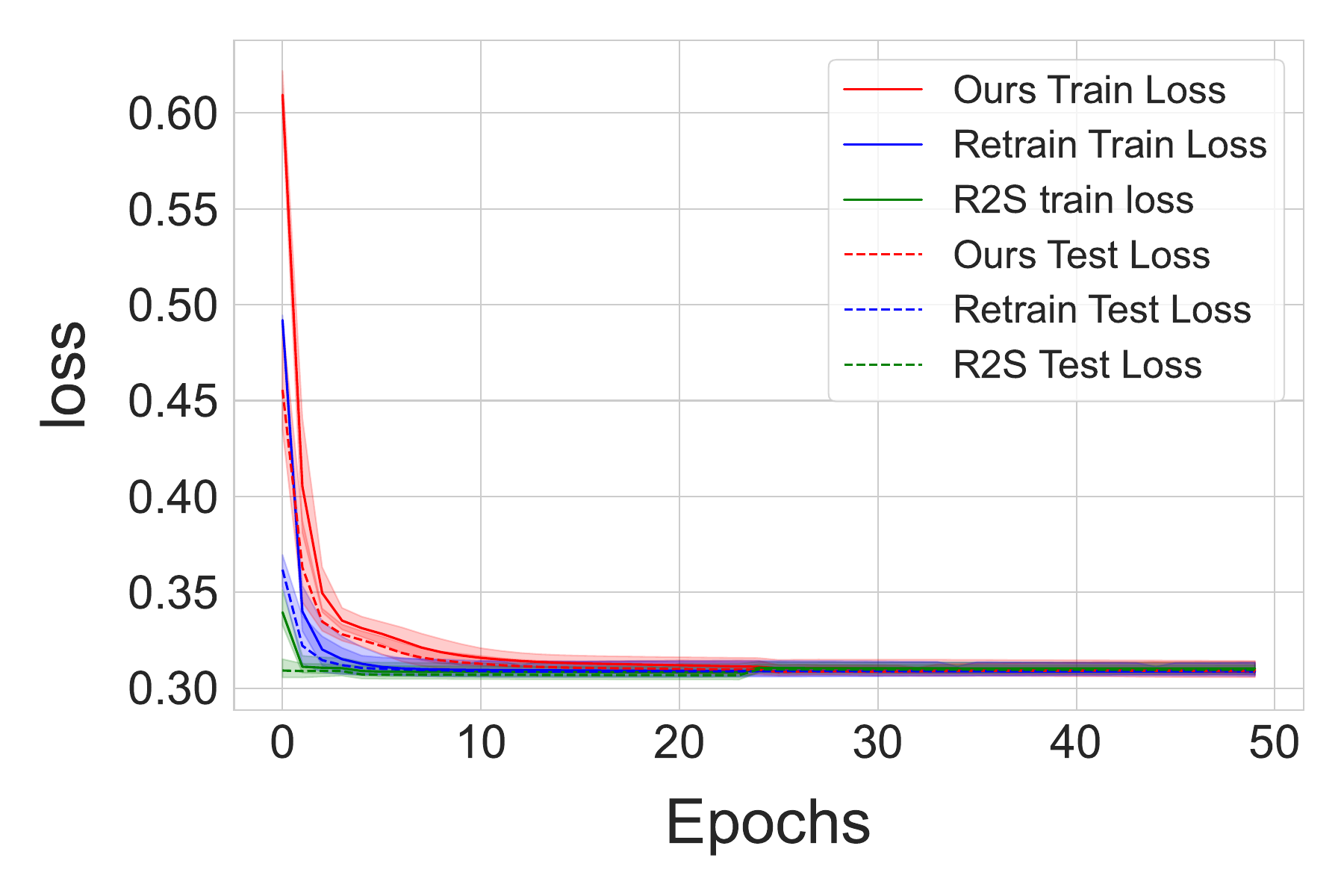}
        \label{hepmass_GA_loss_1}
    }
    \hfill
    \subfloat[Poqemon]{
        \includegraphics[width=0.48\columnwidth]{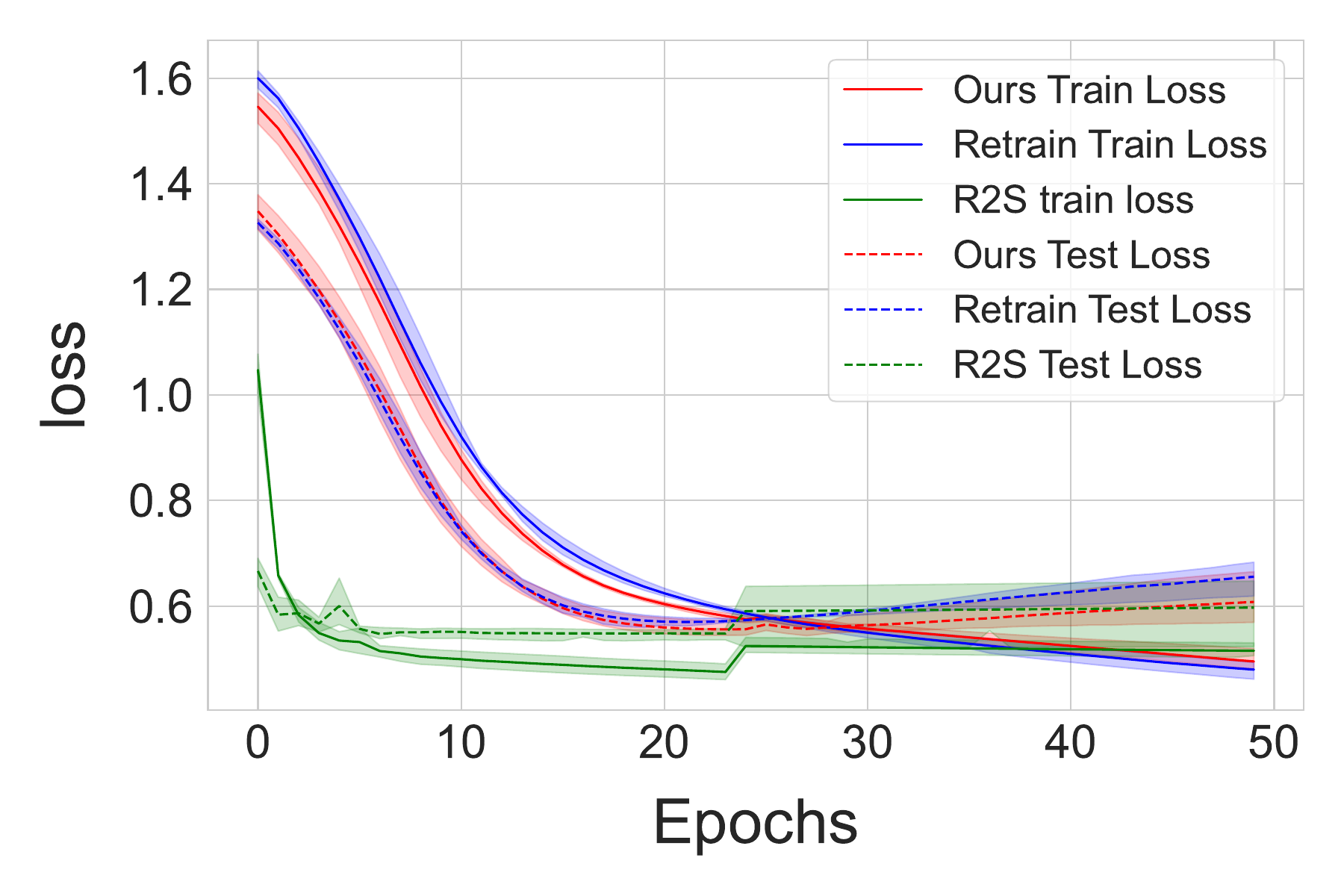}
        \label{poqemon_GA_loss_1}
    }
    \hfill
    \subfloat[Susy]{
        \includegraphics[width=0.48\columnwidth]{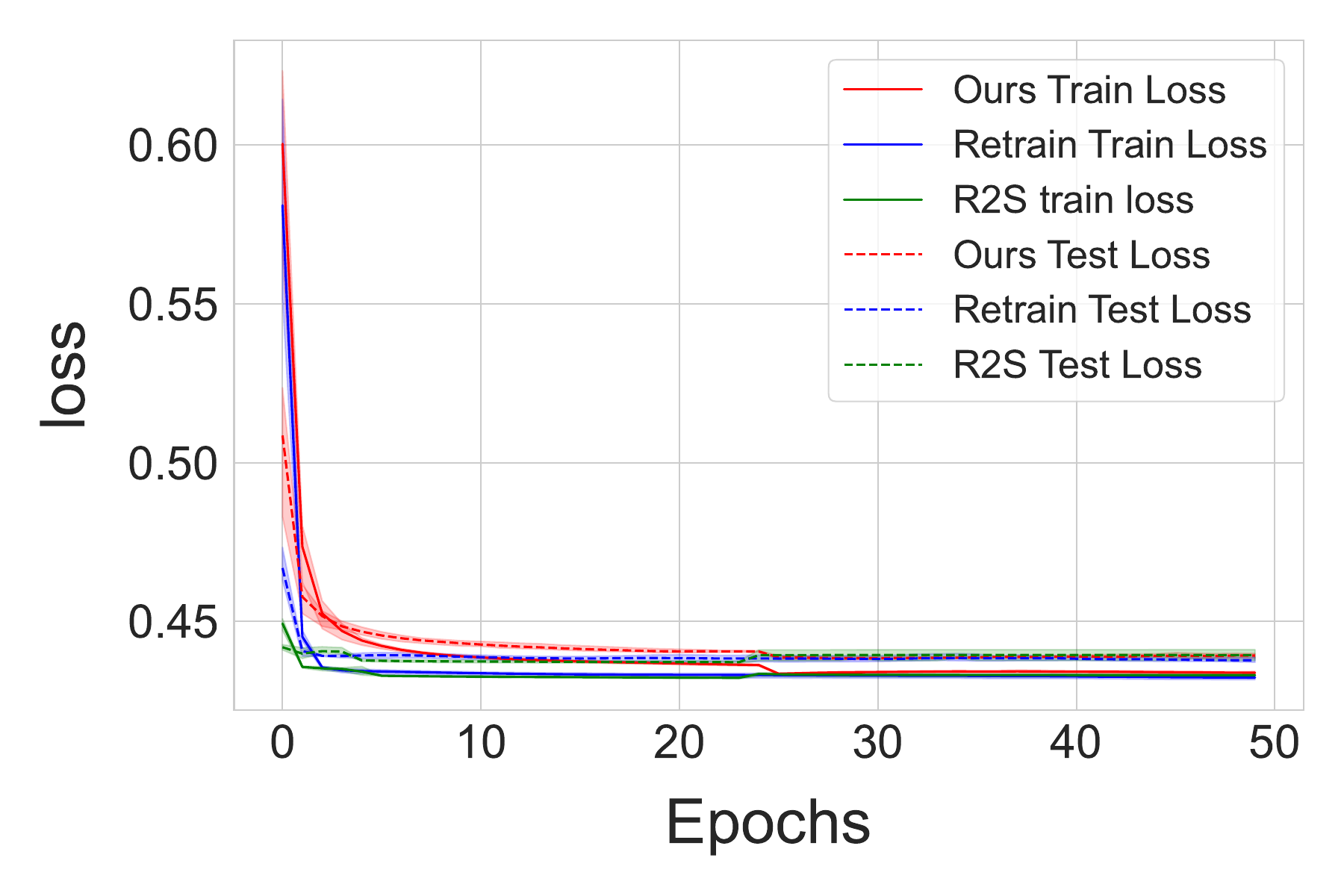}
        \label{susy_GA_loss_1}
    }
    \hfill
    \subfloat[Wine]{
        \includegraphics[width=0.48\columnwidth]{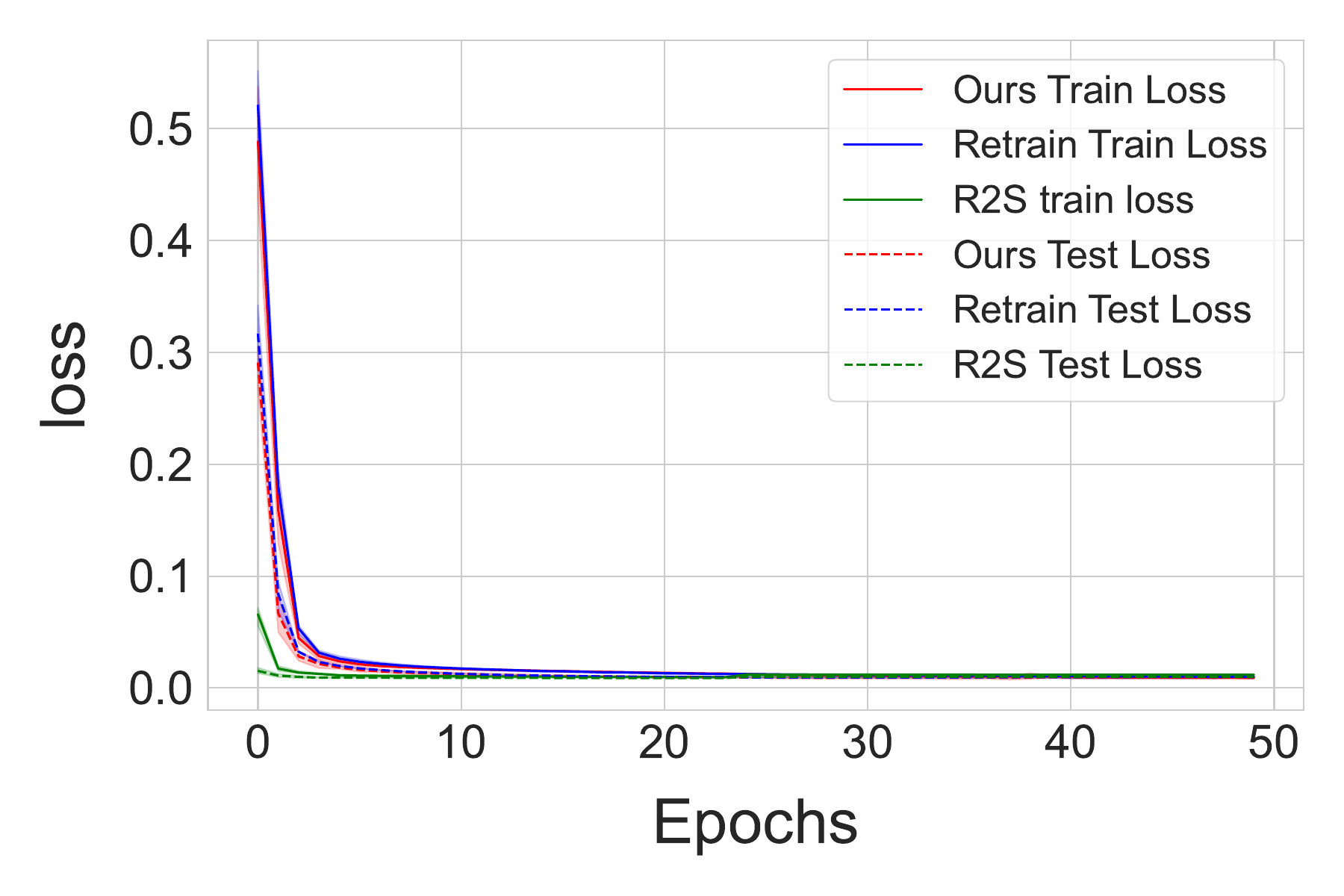}
        \label{wine_GA_loss_1}
    }    
    \caption{The training and test loss of VFU-GA (1 batch) for least important feature, compared to the retrained model from scratch and R2S method.}
    \label{Sample_unlearning_loss_1} 
\end{figure}

Fig.~\ref{Sample_unlearning_loss_5} shows the the comparison of training and test loss between VFU-GA and benchmark model when unlearning 5 batches.

\begin{figure}
    \centering
    \subfloat[Adult]{
        \includegraphics[width=0.48\columnwidth]{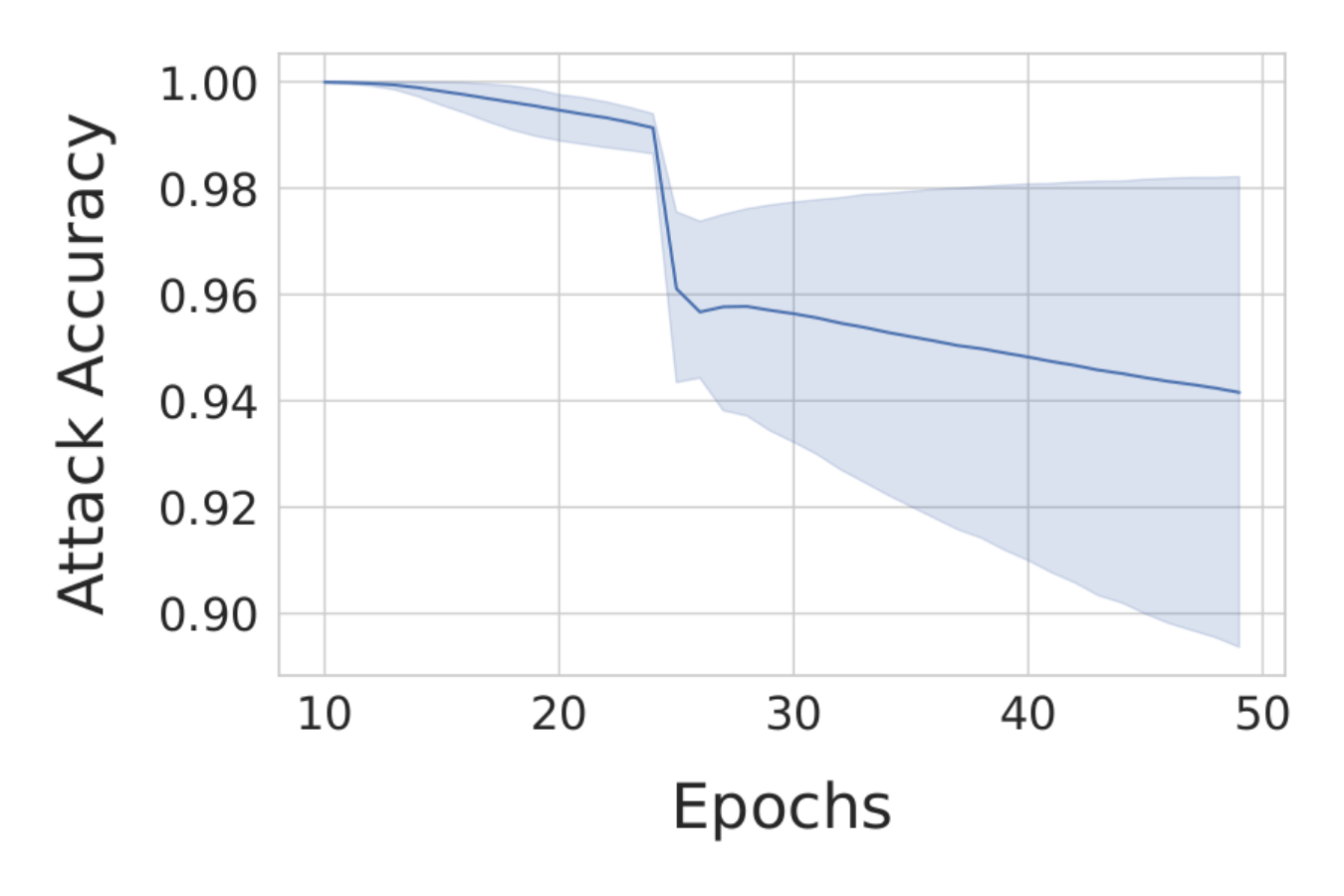}
        \label{adult_GA_MIA_1}
    }
    \hfill
    \subfloat[ai4i]{
        \includegraphics[width=0.48\columnwidth]{fig/ai4i_attack_Accuracy.pdf}
        \label{ai4i_GA_MIA_1}
    }
    \hfill
    \subfloat[Hepmass]{
        \includegraphics[width=0.48\columnwidth]{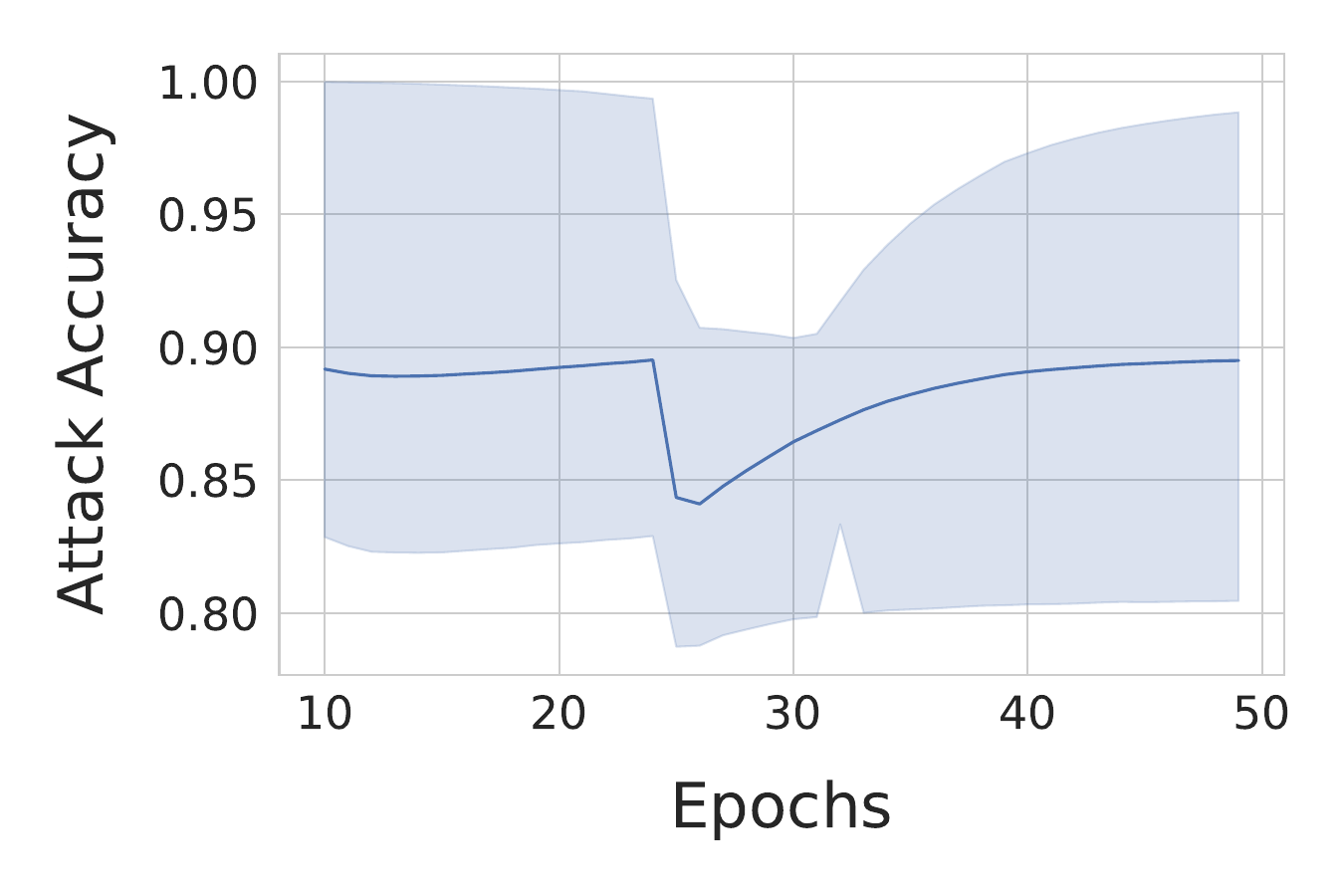}
        \label{hepmass_GA_MIA_1}
    }
    \hfill
    \subfloat[Poqemon]{
        \includegraphics[width=0.48\columnwidth]{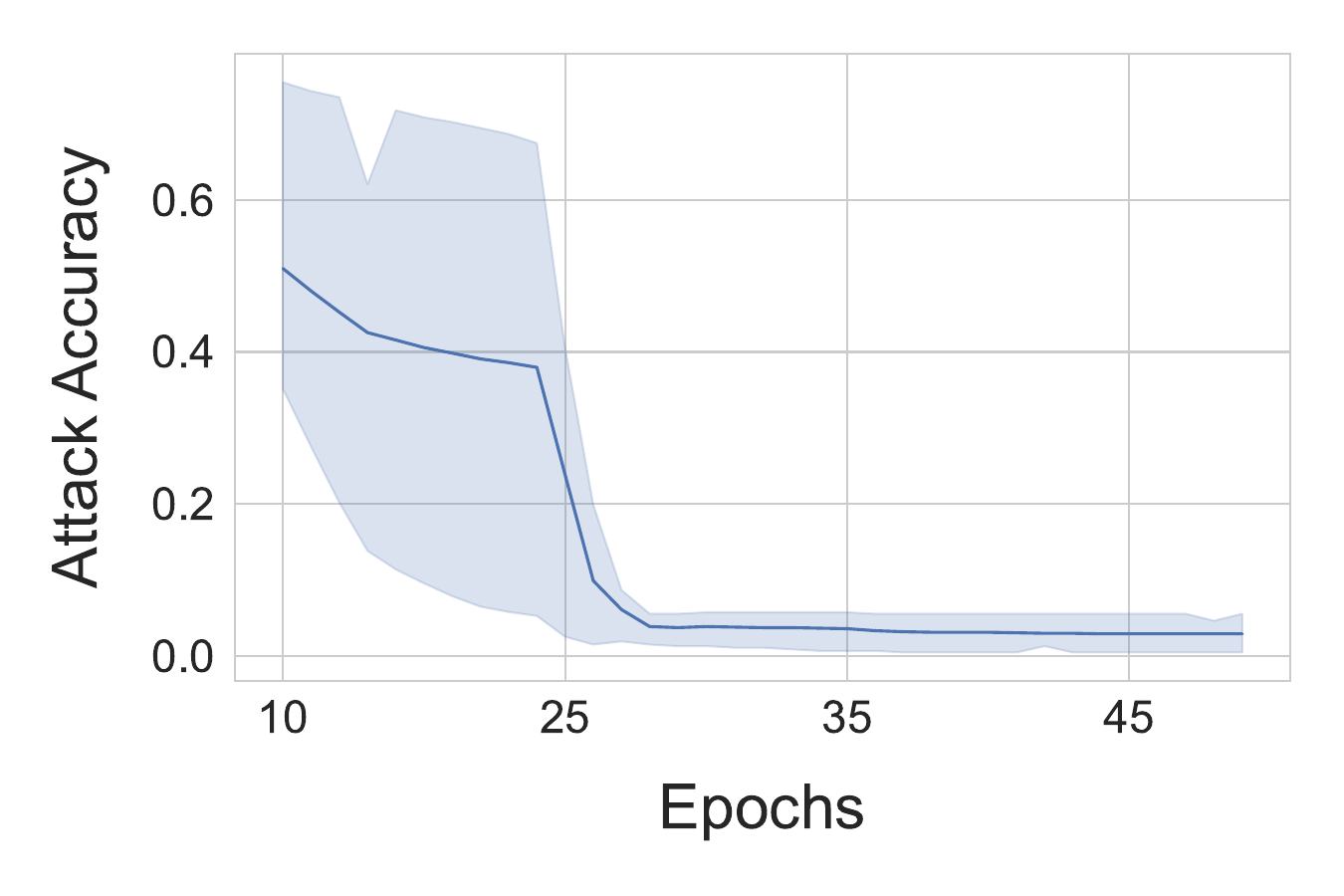}
        \label{poqemon_GA_MIA_1}
    }
    \hfill
    \subfloat[Susy]{
        \includegraphics[width=0.48\columnwidth]{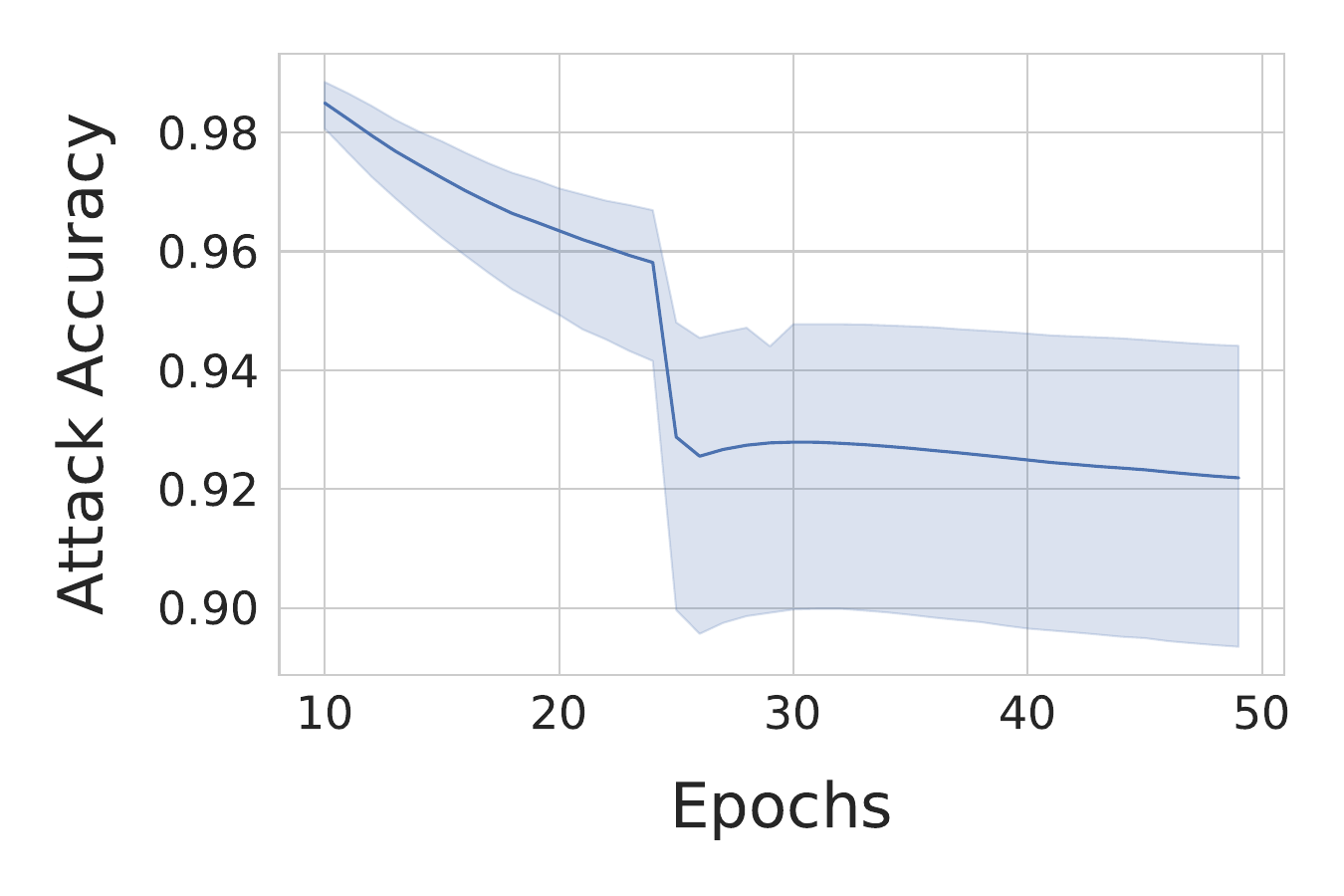}
        \label{susy_GA_MIA_1}
    }
    \hfill
    \subfloat[Wine]{
        \includegraphics[width=0.48\columnwidth]{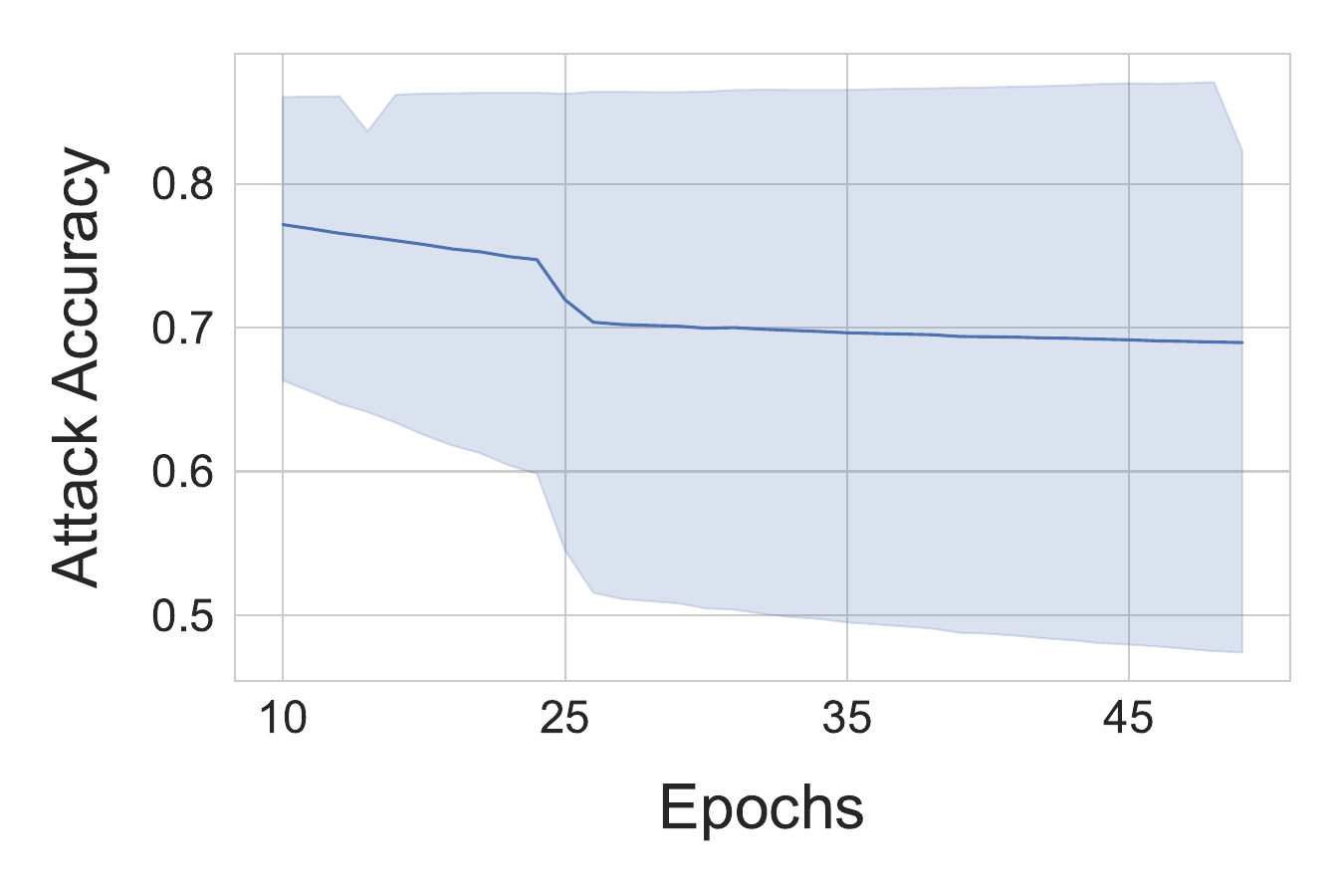}
        \label{wine_GA_MIA_1}
    }
    \caption{The MIA attack accuracy (y-axis) of VFU-GA (1 batch).}
    \label{Sample_unlearning_MIA_1} 
\end{figure}

Fig.~\ref{Sample_unlearning_loss_1} shows the comparison of training and test loss between VFU-GA and benchmark model. Here as well, the model performs better than the benchmark model after unlearning attributed to the robustness introduced while unlearning. Fig.~\ref{Sample_unlearning_MIA_1} shows the performance of MIA attack when unlearning 1 batch. We can see the evidence of unlearning with the drop in accuracy. 

\begin{table}[]
    \centering
    \begin{tabular}{c|cc|cc|cc}
        \toprule
         \multirow{2}{*}{{
         Dataset}} & \multicolumn{2}{c|}{{RfS}} & \multicolumn{2}{c|}{{R2S}} & \multicolumn{2}{c}{{Ours}}\\
         & {AUC} & {F1} & {AUC} & {F1} & {AUC} & {F1} \\
         \midrule
         {Adult} & {0.81} & {0.78} & {0.80} & {0.80} & {\textbf{0.81}} & {0.78} \\
         {ai4i} & {0.91} & {0.93} & {0.91} & {0.93} & {\textbf{0.91}} & {\textbf{0.93}} \\
         {Hepmass} & {0.86} & {0.86} & {0.86} & {0.86} & {\textbf{0.86}} & {\textbf{0.86}} \\
         {Poqemon} & {0.93} & {0.75} & {0.93} & {0.76} & {\textbf{0.93}} & {\textbf{0.76}} \\
         {Susy} & {0.80} & {0.80} & {0.80} & {0.80} & {\textbf{0.80}} & {\textbf{0.80}} \\
         {Wine} & {0.99} & {0.99} & {0.99} & {0.99} & {\textbf{0.99}} & {\textbf{0.99}} \\
        \bottomrule
    \end{tabular}
    \caption{The AUC and F1 score comparison of VFU-GA (Ours) with the retrained-from-scratch (RfS) model and the R2S method for unlearning 1 batches.}
    \label{tab:VFU-GA_1batches}
\end{table}

\end{document}